%% file: EPT_arXiv.tex
\documentclass{article} 
\usepackage{iclr2021_conference,times}

\input{math_commands.tex}

\def\real{\mathbb{R}}
\def\bX{\mathbf{X}}

\usepackage{hyperref}
\usepackage{url}
\usepackage{booktabs}       
\usepackage{amsfonts}       
\usepackage{nicefrac}       
\usepackage{microtype}      
\usepackage{subfigure}
\usepackage{dsfont}
\def\Id{\mathds{1}}
\RequirePackage{amsthm,amsmath,amsfonts,amssymb}
\RequirePackage{graphicx}
\usepackage{enumerate}
\numberwithin{equation}{section}

\usepackage{xcolor}
\usepackage[ruled]{algorithm2e}
\usepackage{algorithmic}
\definecolor{pinegreen}{rgb}{0.0, 0.47, 0.44}

\def\cT{\mathcal{T}}
\def\cL{\mathcal{L}}
\newtheorem{theorem}{Theorem}[section]
\newtheorem{lemma}{Lemma}[section]
\newtheorem{proposition}{Proposition}[section]

\newtheorem*{definition}{Definition}

\newtheorem{remark}{Remark}[section]

\usepackage{geometry}

\title{
Generative Learning With Euler Particle Transport}

\author{Yuan Gao\\
  School of Mathematics and Statistics\\
   Xi'an Jiaotong University,
  Xi'an, China \\
  \texttt{xjtuygao@gmail.com} \\
 \And
  Jian Huang\\
  Department of Statistics and Actuarial Science\\
  University of Iowa,
  Iowa City, Iowa, USA \\
  \texttt{jian-huang@uiowa.edu} \\
   \And
   Yuling Jiao \\
  School of Mathematics and Statistics\\
   Wuhan University,
   Wuhan, China  \\
  \texttt{yulingjiaomath@whu.edu.cn} \\
   \And
 Jin Liu \\
 Center of Quantitative Medicine\\
 Duke-NUS Medical School, Singapore,
 Singapore \\
 \texttt{jin.liu@duke-nus.edu.sg} \\
\And
Xiliang Lu \\
School of Mathematics and Statistics\\
   Wuhan University,
   Wuhan, China  \\
  \texttt{xllv.math@whu.edu.cn}\\
\And
Zhijian Yang \\
School of Mathematics and Statistics\\
   Wuhan University,
   Wuhan, China  \\
  \texttt{zjyang.math@whu.edu.cn}
 }

\iclrfinalcopy 
\begin{document}

\maketitle

\vspace{-0.8 cm}
\begin{abstract}
We propose an Euler particle transport  (EPT) approach for generative learning. The proposed approach is motivated by the problem of finding an optimal transport map from a reference distribution to a target distribution characterized by the Monge-Ampere equation. Interpreting the infinitesimal linearization of the Monge-Ampere equation from the perspective of gradient flows in measure spaces leads to a stochastic McKean-Vlasov equation. We use the forward Euler method to solve this equation. The resulting forward Euler map pushes forward a reference distribution to the target. This map is the composition of a sequence of simple residual maps, which are computationally stable and easy to train. The key task in training is the estimation of the density ratios or differences that determine the residual maps. We estimate the density ratios (differences) based on the Bregman divergence with a gradient penalty using deep density-ratio (difference) fitting. We show that the proposed density-ratio (difference) estimators do not suffer from the ``curse of dimensionality" if data is supported on a lower-dimensional manifold. Numerical experiments with multi-mode synthetic datasets and comparisons with the existing methods on real benchmark datasets support our theoretical results and demonstrate the effectiveness of the proposed method.\footnote{This is an updated version of the manuscript: Gao, Huang, Jiao and Liu (2020). Learning Implicit Generative Models with Theoretical Guarantees. arXiv preprint arXiv:2002.02862.  \url{https://arxiv.org/abs/2002.02862}}\footnote{This version: November 27, 2020}
\end{abstract}


\section{Introduction}
\label{intr}
The ability to efficiently sample from complex distributions plays a key role in a variety of prediction and inference tasks in machine learning and statistics \citep{sala2015}.
The long-standing methodology for learning an underlying distribution relies on an explicit statistical data model, which can be difficult to specify in many applications
such as image analysis, computer vision and natural language processing. In contrast, implicit generative models do not assume a specific form of the data distribution, but rather
learn a nonlinear map to transform a reference distribution to the target distribution.   This modeling approach has been shown to achieve impressive performance in many machine learning tasks \citep{reed16,zhu17}.
\emph{Generative adversarial networks} (GAN) \citep{goodfellow14}, \emph{variational auto-encoders} (VAE) \citep{kingma14} and  \emph{flow-based methods} \citep{rezende2015variational} are important representatives of implicit generative models.

GANs model the low-dimensional latent structure via deep nonlinear factors. They are trained by sequential differentiable surrogates of two-sample tests,  including the density-ratio test \citep{goodfellow14,nowozin16,mao17,mroueh17,tao18} and the density-difference test  \citep{li15,sutherland16,li17,arjovsky17,binkowski18}, among others.
VAE is a probabilistic deep latent
factor model trained with  variational inference and stochastic approximation.
Several authors have proposed improved versions of VAE
by enhancing  the  
representation power of the learned  latent codes  and reducing the blurriness of the generated images in vanilla VAE
 \citep{makhzani2016adversarial,higgins16,tolstikhin2018wasserstein,zhang2019wasserstein}.
Flow-based methods learn a diffeomorphism map between the reference distribution and the target  distribution by maximum likelihood using the change of variables formula.
Recent work on flow-based methods has been focused on developing training methods and designing neural network architectures to trade off between the efficiency of training and sampling and the representation power of the learned map
\citep{rezende2015variational,dinh2014nice,dinh2016density,
kingma2016improved,
papamakarios2017masked,kingma2018glow,grathwohl2019ffjord}.

We propose an Euler particle transport (EPT)
approach for learning a generative model by integrating ideas from optimal transport, numerical ODE,  density-ratio estimation and deep neural networks.
EPT is motivated by the problem of finding an optimal transport from a reference distribution to the target distribution
based on the quadratic Wasserstein distance.
Since it is challenging to solve the resulting  Monge-Amp\`{e}re equation that characterizes the optimal transport, we consider the continuity equation derived from the linearization of the Monge-Amp\`{e}re equation, which is a gradient flows converging to the target distribution. We solve the Mckean-Vlasov equation associated with the gradient flow using the forward Euler method.
The resulting EPT that pushes forward a reference distribution to the target is a composition of a sequence of simple residual maps, which are computationally stable and easy to train. The residual maps are completely determined by the density ratios between the distributions at the current iterations and the target distribution. We estimate density ratios based on the Bregman divergence with a gradient regularizer using deep density-ratio fitting.

We establish bounds on the approximation errors due to linearization of the Monge-Amp\`{e}re equation, Euler discretization of the Mckean-Vlasov equation, and deep density-ratio estimation. Our result on the error rate for the proposed density-ratio estimators improves the minimax rate of nonparametric estimation via exploring the low-dimensional structure of the data and circumvents the ``curse of dimensionality''.
Experimental results on multi-mode synthetic data and comparisons with state-of-the-art GANs on benchmark data support our theoretical findings and demonstrate that EPT is computationally more stable and easier to train than GANs. Using simple ReLU ResNets without batch normalization and spectral normalization, we obtained results that are better than or comparable with those using GANs trained with such tricks.

\section{Euler particle transport}
\label{method}

Let $X \in \real^m$ be a random vector with distribution $\nu$, and let $Z$ be a random vector with distribution $\mu$. We assume that $\mu$ has a known and simple form. Our goal is to construct a transformation $\cT$ such that $\cT_{\#}\mu=  \nu$, where $\cT_{\#}\mu$ denotes the push-forward distribution of $\mu$ by $\cT$, that is, the distribution of $\cT(Z)$. Then we can sample from $\nu$ by first generating a $Z \sim \mu$ and calculate $\cT(Z)$. In practice, $\nu$ is unknown and only a random sample $\{X_i\}_{i=1}^n$ i.i.d. $\nu$ is available. We must construct $\cT$ based on the sample.

There may exist multiple transports $\cT$ with $\cT_{\#}\mu=\nu$. The optimal transport is the one that minimizes
the quadratic Wasserstein distance between $\mu$ and $\nu$ defined by
\begin{align} \label{wd}
\mathcal{W}_{2}(\mu, \nu) = \{ \inf_{\gamma \in \Gamma(\mu, \nu)} \mathbb{E}_{ (Z, X) \sim \gamma} [ \Vert Z - X\Vert_2^2 ] \}^{\frac12},
\end{align}
where $\Gamma(\mu, \nu)$ denotes the set of couplings of $(\mu, \nu)$ \citep{villani2008optimal,ambrosio2008gradient}.
Suppose that $\mu$ and $ \nu$ have densities $q$ and $p$ with respect to the Lesbeque measure, respectively.
Then the optimal transport map $\mathcal{T}$ such that $\mathcal{T}_{\#} \mu = \nu$
is characterized by the Monge-Amp\`{e}re equation \citep{brenier1991polar,mccann1995existence,santambrogio2015optimal}.
Specifically, the minimization problem in (\ref{wd}) admits a unique solution $\gamma = (\Id,\mathcal{T})_{\#} \mu$ with $\mathcal{T} = \nabla \Psi,  \mu \text{-} a.e.,$  where $\Id$ is the identity map and $\nabla \Psi$ is the gradient of the potential function $\Psi: \real^m \to \real$. This function is convex and satisfies the Monge-Amp\`{e}re equation
\begin{equation}\label{mae}
\mathrm{det}(\nabla^2 \Psi(\vz))= \frac{q(\vz)}{p(\nabla \Psi(\vz))}, \vz \in \real^{m}.
\end{equation}
Therefore, to find the optimal transport $\cT$, it suffices to solve (\ref{mae}) for $\Psi$. However, it is challenging to solve this degenerate elliptic equation due to its highly nonlinear nature.

Below we first provide an overall description of  the proposed EPT method for constructing a transport map that pushes forward a reference $\mu$ to the target
$\nu$.  EPT is motivated by solving a linearized version of the Monge-Amp\`{e}re equation (\ref{mae}). It consists of the following steps: (a) linearizing (\ref{mae}) via residual maps,
(b) determining the velocity fields governing the stochastic McKean-Vlasov equation resulting from the linearization, (c) calculating the forward Euler particle transport map and, (d) training the EPT map by estimating the velocity fields from data. Since velocity fields are completely determined by density ratios, this step amounts to nonparametric density ratio estimation.
We also provide bounds on the errors due to linearization, discretization and estimation.  Mathematical details and proofs are given in Section \ref{theory} and the appendix.

\textbf{Linearization via residual map }
A basic approach to addressing the difficulty due to nonlinearity is linearization. We use a linearization method based on the residual map
\begin{equation}\label{rm}
   \mathcal{T}_{t,\Phi} = \nabla \Psi = \Id  + t \nabla \Phi_t, t \ge 0,
    \end{equation}
where $\Phi_t: \mathbb{R}^{m}  \rightarrow \mathbb{R}^1$
is a function to be chosen such that the law of $\mathcal{T}_{t,\Phi}(Z)$
is closer to $\nu$ than that of $Z$ \citep{villani2008optimal}.
We then iteratively improve the approximation by repeatedly applying the residual map to the current particles.
We give the specific form of $\Phi$ below, see
Theorem \ref{th1b} in Section \ref{theory} for details.

This linearization scheme leads to the stochastic process
$\mathbf{X}_t: \mathbb{R}^{m} \rightarrow \mathbb{R}^{m} $ satisfying the McKean-Vlasov equation
\begin{equation}\label{mve}
\frac{{\rm d}}{{\rm d} t} \mathbf{X}_t(\vx) = \vv_{t}(\mathbf{X}_t(\vx)),\ t\ge 0,  \ \mathrm{with} \ \  \mathbf{X}_0  \sim \mu,\  \mu\text{- a.e.} \ \vx \in \mathbb{R}^{m},
\end{equation}
where $\vv_t$ is the velocity vector field  of $\bX_t$.
In addition,
we have $\vv_t= \nabla \Phi_t.$  Thus $\vv_t$ also determines the residual map (\ref{rm}). The detailed derivations are given in Theorems \ref{th1b} and \ref{th2}.
in Section \ref{theory}.
Therefore,  the problem of estimating the residual maps (\ref{rm}) is equivalent to that of estimating the velocity fields $\vv_t$.

The movement of $\bX_t$ along $t$ is completely governed by $\vv_t$, given the initial value. We choose a $\vv_t$ to decrease the discrepancy between the distribution of $\bX_t$, say $\mu_t $,  at time $t$ and the target $\nu$ with respect to a properly chosen measure. An equivalent formulation of (\ref{mve}) is through the gradient flow $\{\mu_t\}_{t\ge 0}$ with $\{\vv_t\}_{t\ge 0}$ as its velocity fields, see Proposition \ref{th1} in Section \ref{theory}. Computationally it is more convenient to work with (\ref{mve}).

\textbf{Determining velocity field }
The basic intuition is that we should move in the direction
that decreases the differences between $\mu_t$ and the target $\nu$. We use an energy functional $\cL[\mu_t]$ to measure such differences. An important energy functional $\cL[\mu_t]$ is the $f$-divergence \citep{ali1966general},
\begin{equation}
\label{fdiv}
\cL[\mu_t] =
\mathbb{D}_f(\mu_t \Vert \nu) = \int_{\mathbb{R}^{m}} p(\vx) f\left(\frac{q_t(\vx)}{p(\vx)} \right) {\mathrm{d}} \vx,
\end{equation}
where $q_t$ is the density of $\mu_t$, $p$ is the density of $\nu$  and $f: \mathbb{R}^+ \rightarrow \mathbb{R} $ is assumed to be a twice-differentiable convex 
function with $f(1) = 0$.
We choose $\Phi_t$ such that $\cL[\mu_t]$
is minimized.  We show in Theorem \ref{th1b}
in Section \ref{theory} that
$\Phi_t(\vx)  = -f^{\prime} (r_t(\vx))$ and
$\vv_t(\vx)=\nabla \Phi_t(\vx)$. Therefore,
\[
\vv_t(\vx)=-f^{\prime\prime}(r_t(\vx))\nabla r_t(\vx),\
\text{ where } \  r_t(\vx) = \frac{q_t(\vx)}{p(\vx)},\ \vx \in \real^m.
\]
For example, if we use the $\chi^2$-divergence with $f(c)=(c-1)^2/2$, then $\vv_t(\vx)=\nabla r_t(\vx)$ is simply the gradient of the density ratio. Other types of velocity fields can be obtained by using different energy functionals such as the Lebesgue norm of the density difference, i.e., $\mathcal{L}[\mu_t] =  \int_{\mathbb{R}^{m}} |q_t(\vx)- p(\vx)|^2  {\mathrm{d}} \vx$, see Section \ref{theory} for details.

\textbf{The forward Euler method }
Numerically, we need to discretize the McKean-Vlasov equation (\ref{mve}). Let $s >0$ be a small step size.
We use the forward Euler method defined iteratively by:
\begin{align}
\mathcal{T}_{k} &= \Id + s \vv_{k},\label{eu1}\\
 \mathbf{X}_{k+1}& = \mathcal{T}_{k}(\mathbf{X}_k), \label{eu2}\\
\mu_{k+1} &=  (\mathcal{T}_{k})_{\#}  \mu_k, \label{eu3}
\end{align}
where $\mathbf{X}_0 \sim \mu$, $\mu_0 = \mu$, $\vv_k$ is the velocity field at the $k$th step, $k = 0,1,...,K$ for some large $K$.
The particle process $\{\bX_k\}_{k\ge 0}$ is a discretized version of the continuous process $\{\bX_t\}_{t\ge 0}$ in (\ref{mve}).
The final transport map is the composition of a sequence of simple residual maps $\cT_0, \cT_1\ldots, \cT_K$, i.e., $\cT=\cT_K\circ \cT_{K-1} \cdots \circ \cT_0.$
This updating scheme is based  on the forward Euler method for solving equation (\ref{mve}). This is the reason we refer to the proposed method as Euler particle transport (EPT).

\textbf{Training EPT }
When the target $\nu$ is unknown and only a random sample is available, it is natural  to
learn $\nu$  by  first estimating  the discrete velocity fields $\vv_k$  at the sample level and then plugging the estimator of $\vv_k$ in (\ref{eu1}).
For example, if we use the
$f$-divergence as the energy functional, estimating $\vv_k(\vx)=-f^{\prime\prime}(r_k(\vx))\nabla r_k(\vx)$ boils down to estimating
the density ratios $r_k(\vx)=q_k(\vx)/p(\vx)$ dynamically at each iteration $k.$ Nonparametric density-ratio estimation using Bregman divergences and gradient regularizer
are discussed in Section \ref{dr} below.
Let $\hat{\vv}_k$ be the estimated velocity fields at the $k$th iteration. The $k$th estimated residual map is
$\widehat{\cT}_k = \Id + s \hat{\vv}_k.$
Finally, the trained map is
\begin{equation}
\label{ETPa}
\widehat{\cT}=\widehat{\cT}_K\circ \widehat{\cT}_{K-1}\circ \cdots
\circ \widehat{\cT}_0.
\end{equation}

\textbf{Error analysis }
We establish the following bound on the approximation error due to the linearization of the Monge-Amp\`{e}re equation under appropriate conditions:
\begin{equation}
\label{erra}
\mathcal{W}_2 (\mu_t,\nu) = \mathcal{O}(e^{- \lambda t}),
\end{equation}
for some $\lambda > 0$, see Proposition \ref{th1} Section \ref{theory}.
Therefore, $\mu_t$ converges to $\nu$ exponentially fast as
$t \to \infty$.
{\color{black}
For an integer $K \ge 1$ and a small $s > 0$, let $\{\mu_t^{s}:  t\in [ks,(k+1)s), k=0, \ldots, K\}$ be a piecewise constant interpolation between $\mu_{ks}$ and  $\mu_{(k+1)s}, k=0, 1, \ldots, K.$
Under the assumption that the velocity fields $\vv_{t}$ are Lipschitz continuous with respect to $(\vx,\mu_t)$,
it is shown in Proposition \ref{prop2} in
Section \ref{theory} the discretization error of $\mu_{t}^{s}$ can be bounded in a finite time interval $[0,T)$ as follows:
\begin{equation}
\label{errb}
\sup_{t \in [0, T)} \mathcal{W}_2(\mu_t, \mu_t^s) = \mathcal{O}(s).
\end{equation}
The error bounds (\ref{erra}) and (\ref{errb})
imply that the distribution of the particles  $\mathbf{X}_k$  generated by the EPT map defined in (\ref{eu2}) with a small $s$ and a sufficiently large $k$ converges to the target $\nu$ at the rate of discretization size $s$.
}

When training the EPT map, we use the deep neural networks to estimate the density ratios (density differences) with samples. In Theorem \ref{th3}, we provide an estimation error bound that  improves the minimax rate of deep nonparametric estimation via exploring the low-dimensional structure of data and circumvents the ``curse of dimensionality.''  Thus this result is of independent interest in nonparametric estimation using deep neural networks.

\section{Gradient flows associated with EPT}
\label{theory}
For convenience, we first describe the notation used in the remaining sections.
Let $\mathcal{P}_2(\mathbb{R}^{m})$ denote the space of Borel probability measures on $\mathbb{R}^{m}$ with finite second moments, and let $\mathcal{P}_2^{a}(\mathbb{R}^{m})$ denote the subset of $\mathcal{P}_2(\mathbb{R}^{m})$ in which measures are absolutely continuous with respect to the Lebesgue measure (all distributions are assumed to satisfy this assumption hereinafter).
$\mathrm{Tan}_{\mu}\mathcal{P}_2(\mathbb{R}^{m})$ denotes the tangent space  to $ \mathcal{P}_2(\mathbb{R}^{m})$ at $\mu$. Let  $\mathrm{AC}_{\mathrm{loc}}(\mathbb{R}^+,\mathcal{P}_2(\mathbb{R}^{m})) = \{\mu_t : I \rightarrow \mathcal{P}_2(\mathbb{R}^{m})   \textrm{ is  absolutely continuous},$
$|\mu_t^{\prime}| \in L^{2}(I), I \subset \mathbb{R}^+\}$. $\mathrm{Lip}_{\mathrm{loc}}(\mathbb{R}^{m})$ denotes the set of functions that are Lipschitz continuous on any compact set of $\mathbb{R}^{m}$.
For any $\ell \in [1,\infty],$ we use  $L^{\ell}(\mu, \mathbb{R}^{m})$  ($L^{\ell}_{\mathrm{loc}}(\mu, \mathbb{R}^{m})$) to denote the
$L^{\ell}$ space of $\mu$-measurable functions on $\mathbb{R}^{m}$ (on any compact set of $\mathbb{R}^{m}$).
With $\Id$,  $\mathrm{det}$ and $\mathrm{tr}$, we refer to the identity map,  the determinant and the trace. We use $\nabla$, $\nabla^2$ and $\Delta$ to denote the gradient or Jacobian operator, the Hessian operator and the Laplace operator, respectively.

We are now ready to describe EPT in details. Specifically, we
describe the gradient flows associated with
EPT and the corresponding Mckean-Vlasov equation.
Let $X \sim q$, and let
$$\widetilde{X} = \mathcal{T}_{t,\Phi}(X)=
X+t\nabla\Phi(X), t \ge 0.$$
Here we let $\Phi$ be independent of $t$ for the moment.
Denote the distribution of $\widetilde{X}$ by $\widetilde{q}.$
With a small $t$, the map $\mathcal{T}_{t,\Phi} $ is invertible according to the implicit function theorem.
By the change of variables formula, we have
\begin{equation}\label{cv}
 \mathrm{det}(\nabla^2 \Psi)(\vx) =  |\mathrm{det}(\nabla \mathcal{T}_{t,\Phi})(\vx)| = \frac{q(\vx)}{\widetilde{q}(\tilde{\vx})},
 \end{equation}
where
\begin{equation}\label{xc}
\tilde{\vx} = \mathcal{T}_{t,\Phi}(\vx).
\end{equation}
Using the fact that the derivative
$$\left.\frac{\mathrm{d}}{\mathrm{d} t}\right|_{t=0} \mathrm{det}(\mathbf{A}+t \mathbf{B})=\mathrm{det}(\mathbf{A}) \mathrm{tr}\left(\mathbf{A}^{-1} \mathbf{B}\right)
\forall \mathbf{A}, \mathbf{B} \in \mathbb{R}^{m \times m},$$
provided that $\mathbf{A}$ is invertible, and applying the first order Taylor expansion to (\ref{cv}),  we have
\begin{align}
\log \widetilde{q}(\tilde{\vx}) - \log q(\vx)
= - t \Delta \Phi(\vx) + o(t). \label{dc}
\end{align}
Let $t\rightarrow 0$ in (\ref{xc}) and (\ref{dc}),  we obtain
 a random process $\{\vx_t\}$ and its law $q_t$ satisfying
\begin{align}
\frac{\mathrm{d} \vx_t}{\mathrm{d} t} &= \nabla \Phi(\vx_t), \ \  \mathrm{with} \ \  \vx_0  \sim q, \label{leq1}\\
\frac{\mathrm{d} \ln  q_t(\vx_t)}{\mathrm{d} t} &= - \Delta 
\Phi (\vx_t), \ \  \mathrm{with} \ \  q_0 = q.\label{leq2}
\end{align}
Equations (\ref{leq1}) and (\ref{leq2}) resulting from  linearization  of the Monge-Amp\`{e}re equation (\ref{mae})
can be interpreted as gradient flows in measure spaces \citep{ambrosio2008gradient}.
Thanks to this connection, we can resort to solving a continuity equation characterized by a  McKean-Vlasov equation, an ODE system that is easier to work with.

For $\mu \in \mathcal{P}_2^a (\mathbb{R}^{m})$ with density $q$, let
\begin{equation}
\label{energyFun}
\mathcal{L}[\mu] = \int_{\mathbb{R}^{m}} F(q(\vx)) {\textrm d} \vx: \mathcal{P}_2^a (\mathbb{R}^{m})  \rightarrow \mathbb{R}^{+} \cup \{0\}
\end{equation}
be an energy functional satisfying $\nu \in \arg\min \mathcal{L}[\cdot] ,$ where $F(\cdot): \mathbb{R}^{+} \rightarrow \mathbb{R}^{1}$ is a twice-differentiable convex function.
Among the widely used metrics on $ \mathcal{P}_2^a (\mathbb{R}^{m})$ in implicit generative learning, the following   two   are important examples of $\mathcal{L}[\cdot]:$
(1) $f$-divergence given in (\ref{fdiv}) \citep{ali1966general};
(2) Lebesgue norm of density difference:
\begin{equation}\label{lm}
\|\mu-\nu\|^2_{L^2(\mathbb{R}^{m})} =  \int_{\mathbb{R}^{m}} |q(\vx)- p(\vx)|^2  {\mathrm{d}} \vx.
\end{equation}

\begin{definition}
We call   $\{\mu_t\}_{t\in \mathbb{R}^+} \subset \mathrm{AC}_{\mathrm{loc}}(\mathbb{R}^+,\mathcal{P}_2(\mathbb{R}^{m}))$  a  gradient flow of the  functional $\mathcal{L}[\cdot]$,   if
 $\{\mu_t\}_{t\in \mathbb{R}^+} \subset  \mathcal{P}_2^a (\mathbb{R}^{m})$ $a.e.,   \ t \in \mathbb{R}^{+}$
and  the  velocity vector field $\vv_t \in \mathrm{Tan}_{\mu_t}\mathcal{P}_2 (\mathbb{R}^{m})$ satisfies
  $\vv_t \in -\partial \mathcal{L}[\mu_t] \quad a.e. \quad t \in \mathbb{R}^+,$
  where $\partial \mathcal{L}[\cdot]$ is the subdifferential of $\mathcal{L}[\cdot]$.
\end{definition}

The gradient flow $\{\mu_t\}_{t\in \mathbb{R}^+}$ of $\mathcal{L}[\cdot]$  enjoys the following nice properties.

\begin{proposition}\label{th1}
\begin{enumerate} 
 \item[(i)] The following continuity equation holds in the sense of distributions.
\begin{equation}\label{vfp}
\frac{\partial}{\partial t}\mu_t
 = -\nabla\cdot(\mu_t\vv_t)\ \ {\textrm in} \ \ \mathbb{R}^+ \times \mathbb{R}^{m} \ \ \mathrm{with} \ \ \mu_0 = \mu, t\ge 0.
\end{equation}

 \item[(ii)] Energy decay along the gradient flow: $\frac{{\textrm d} }{{\textrm d} t} \mathcal{L}[\mu_t] = - \|\vv_t\|^2_{L^2(\mu_t,\mathbb{R}^{m})} \quad a.e. \quad t \in \mathbb{R}^+.$
In addition,
\begin{equation}
\label{Wconverg}\mathcal{W}_2 (\mu_t,\nu) = \mathcal{O}(\exp^{- \lambda t}),
\end{equation}
if $\mathcal{L}[\mu]$ is $\lambda$-geodetically convex with $\lambda>0$
\footnote{
$\mathcal{L}$ is said to be
$\lambda$-geodetically convex if there exists a constant $\lambda > 0$ such that for every $\mu_1$, $\mu_2 \in \mathcal{P}_2^a (\mathbb{R}^{m}),$  there exists a constant
speed geodestic $\gamma: [0, 1] \to \mathcal{P}_2^a (\mathbb{R}^{m})$ such that $\gamma_0=\mu_1, \gamma_1=\mu_2$ and
\[
\mathcal{L}(\gamma_s) \le (1-s)\mathcal{L}(\mu_1)
+ s \mathcal{L}(\mu_2) - \frac{\lambda}{2} s(1-s)
d(\mu_1,\mu_2), \ \forall s \in [0,1],
\]
where  $d$ is a metric defined on $\mathcal{P}_2^a (\mathbb{R}^{m})$ such as the quadratic
Wasserstein distance.}.

\item[(iii)] Conversely, if $\{\mu_t\}_t$ is the solution of continuity equation  (\ref{vfp}) in (i) with $\vv_t(\vx)$  specified by (\ref{vr}) in (ii), then
$\{\mu_t\}_t$ is a gradient flow of $\mathcal{L}[\cdot]$.
\end{enumerate}
\end{proposition}

{\color{black}
\begin{remark}
In part (ii) of Proposition \ref{th1}, for general $f$-divergences,
we assume the functional $\cL$ to be $\lambda$-geodesically convex for the convergence of $\mu_t$ to the target $\nu$ in the quadratic Wasserstein distance. However,  for the KL divergence, the convergence can be guaranteed if $\nu$
satisfies the log-Sobolev inequality\citep{otto2000}.
In addition, the distributions that are strongly log-concave outside a bounded region, but not necessarily log-concave inside the region satisfy the log-Sobolev inequality, see, for example, \cite{holley1987}. Here the functional $\cL$ can even be nonconvex, an example includes the densities with double-well potential.
\end{remark}
}

\begin{remark}
Equation (8.48) in Proposition 8.4.6 of and Ambrosio et al. (2008)  shows the connection (locally) of the velocity $v_t$ of the gradient flow $\mu_t$ and the  optimal transport along $\mu_t$, i.e., let  $T_{\mu_t}^{\mu_{t+h}}$ be the optimal transport from $\mu_t$ to $\mu_{t+h}$ for a small $h>0$, then  $T_{\mu_t}^{\mu_{t+h}} = I + h v_t + o(h)$ in $L^p$.  So locally, $I+hv_t$ approximates the optimal transport map from $\mu_t$ to $\mu_{t+h}$ on $[t, t+h]$.
However, the global approximation property of the proposed method is not clear. This is a challenging problem that requires further study and is beyond the scope of this paper.
\end{remark}

{\color{black}
\begin{theorem}
\label{th1b}
(i)
Representation of the velocity fields: if the density $q_t$ of $\mu_t$ is differentiable, then
\begin{equation}\label{vr}
 \vv_t(\vx) =  -\nabla F^{\prime}(q_t(\vx)) \ \ \mu_t\text{-}a.e. \ \  \vx \in \mathbb{R}^{m}.
 \end{equation}

(ii)
If we let $\Phi$ be time-dependent in  (\ref{leq1})--(\ref{leq2}), i.e., $\Phi_t$,  then the linearized  Monge-Amp\`{e}re equations (\ref{leq1})--(\ref{leq2}) are the same as the continuity equation (\ref{vfp})
by taking  $\Phi_t(\vx)  = -F^{\prime} (q_t(\vx)).$
\end{theorem}
}

Theorem \ref{th1b} and (\ref{Wconverg}) in Proposition \ref{th1} imply that $\{\mu_t\}_t$, the solution of the continuity equation  (\ref{vfp}) with
$\vv_t(\vx) =  -\nabla F^{\prime}(q_t(\vx)),$
converges rapidly  to the target distribution $\nu$.
Furthermore,   the continuity equation has the following representation under mild regularity conditions on the velocity fields.

\begin{theorem}\label{th2}
Assume $\|\vv_t\|_{L^{1}(\mu_t,\mathbb{R}^{m})}\in L^{1}_{\mathrm{loc}}(\mathbb{R}^+)$
and  $\vv_t(\cdot)\in \mathrm{Lip}_{\mathrm{loc}}(\mathbb{R}^{m})$ with upper bound $B_t$ and Lipschitz constant $L_t$ such that $(B_t + L_t) \in  L^{1}_{\mathrm{loc}}(\mathbb{R}^+).$
Then the solution of  the continuity equation  (\ref{vfp}) can be represented as
$\mu_t = (\mathbf{X}_t)_{\#}\mu,
$
 where  $\mathbf{X}_t(\vx): \mathbb{R}^+ \times \mathbb{R}^{m} \rightarrow \mathbb{R}^{m} $ satisfies the  McKean-Vlasov equation
 (\ref{mve}).
\end{theorem}

As shown  in  Lemma \ref{lem2} below, the velocity fields associated with the $f$-divergence (\ref{fdiv}) and the Lebesgue norm  (\ref{lm})
are determined by density ratio and density difference, respectively.
\begin{lemma}\label{lem2} The velocity fields $\vv_t$ satisfy
\begin{equation*}
\vv_{t}(\vx) =
\left\{\begin{array}{ll}
-f^{\prime\prime}(r_t(\vx))\nabla r_t(\vx), \ \  \mathcal{L}[\mu] = \mathbb{D}_f(\mu \Vert \nu), \text{ where } r_t(\vx) = \frac{q_t(\vx)}{p(\vx)}, \\
- 2\nabla d_t(\vx),  \ \ \mathcal{L}[\mu] = \|\mu-\nu\|^2_{L^2(\mathbb{R}^{m})}, \text{ where } d_t(\vx) = q_t(\vx) -  p(\vx).
\end{array}
\right.
\end{equation*}
\end{lemma}

Several methods have been developed to estimate  density ratio and density difference in the literature. Examples include probabilistic classification approaches, moment matching and direct density-ratio (difference) fitting, see \cite{sugiyama2012density2,sugiyama2012density,kanamori2014statistical,mohamed2016learning} and the references therein.

{\color{black}
\begin{proposition}\label{prop2}
For any finite $T > 0$,
suppose that the velocity fields $\vv_{t}$ are Lipschitz continuous with respect to $(\vx,\mu_t)$ for $ t \in [0, T]$, that is, there exists a finite constant $L_{\vv} > 0$ such that
\begin{equation}
\label{Lip}
\|\vv_{t}(\vx) - \vv_{\tilde{t}}(\tilde{\vx})\| \leq L_{\vv} [\|\vx -\tilde{\vx}\| + \mathcal{W}_2(\mu_t, \mu_{\tilde{t}})], t, \tilde{t} \in [0, T]\  \text{ and }\ \vx, \tilde{\vx} \in \real^m.
\end{equation}
Then the bound (\ref{errb}) on the discretization error holds:
$$
\sup_{t \in [0, T]} \mathcal{W}_2(\mu_t, \mu_t^s) = \mathcal{O}(s).$$
\end{proposition}
}
\begin{remark}
If we take $f(x)=(x-1)^2/2$ in Lemma \ref{lem2}, then the velocity fields $\vv_t(\vx) = \nabla \vr_t(\vx)$, where
$\vr_t(\vx) = q_t(\vx)/p(\vx)$. In the proof of Theorem \ref{th1}, part (ii), it is shown that $q_t$ satisfies ${\mathrm{d}q_t}/{\mathrm{d}t} = -\nabla\cdot(q_t \nabla \Phi_t).$ Thus for this simple $f$-divergence function, the verification of  the Lipschitz condition (\ref{Lip}) amounts to
verifying that $\nabla \vr_t(\vx)$ is Lipschitz in the sense of
(\ref{Lip}).
\end{remark}

\section{Deep density-ratio fitting}
\label{dr}
The evaluation of velocity fields depends on the dynamic estimation of a discrepancy between the push-forward distribution $q_t$ and the target distribution $p$.
Density-ratio and density-difference fitting with the Bregman score provides a unified framework for such discrepancy estimation
without estimating each density separately
\citep{gneiting2007strictly,dawid2007geometry,
sugiyama2012density2,sugiyama2012density,kanamori2014statistical}.


Let $r(\vx) = {q(\vx)}/{p(\vx)}$ be the density ratio between a given density  $q(\vx)$ and the target $p(\vx)$.
Let $g: \mathbb{R} \rightarrow \mathbb{R}$ be a differentiable and strictly convex function.
The separable Bregman score with the base probability density $p$ for measuring the discrepancy between $r$ and a measurable function
$R: \real^m \to \real^1$  is
\begin{align*}
&\mathfrak{B}(r, R)
= \mathbb{E}_{X \sim p} [ g^{\prime}(R(X)) R(X) - g(R(X)) ]  - \mathbb{E}_{X\sim q} [g^{\prime}(R(X))].
\end{align*}
Here we focus on the widely used least-squares density-ratio (LSDR) fitting with $g(x) = (x-1)^2$ as a working example, i.e.,
\begin{equation}
\label{LSDRa}
\mathfrak{B}_{\textrm LSDR}(r, R)
= \mathbb{E}_{X \sim p} [ R(X) ^2 ]
- 2 \mathbb{E}_{X \sim q} [R(X)] + 1.
\end{equation}
For other choice of $g$,  such as $g(x) = x\log x - (x+1)\log(x+1)$ corresponding to estimating $r$ via the logistic regression (LR),  and the  scenario of   density difference fitting  will be  presented in  Section \ref{drapp}.


\subsection{Gradient regularizer}
The distributions of real data may have a low-dimensional structure with their support concentrated on a low-dimensional manifold, which   may cause the  $f$-divergence to be  ill-posed due to non-overlapping supports. To exploit such underlying low-dimensional structures and avoid ill-posedness,
we derive a simple weighted gradient regularizer
$\frac12 \mathbb{E}_{p} [g^{\prime\prime}(R) \Vert \nabla R \Vert_2^2 ],$
motivated by recent works on smoothing via noise injection \citep{sonderby2016amortised,arjovsky2017principled}.
This serves as a regularizer for deep density-ratio fitting.
For example, with $g(c) = (c-1)^2$, the resulting gradient regularizer is
\begin{equation}\label{gp}
   \mathbb{E}_{p} [\Vert \nabla R \Vert_2^2],
\end{equation}
which recovers the well-known squared Sobolev semi-norm in nonparametric statistics. Gradient regularization stabilizes and improves the long time performance of EPT. The detailed derivation is presented in Section \ref{drapp}.

\subsection{LSDR estimation with gradient regularizer}
Let   $\{X_i\}_{i=1}^n$ and $\{Y_i\}_{i=1}^n $ be two collections of i.i.d data from densities $p(\vx)$ and $q(\vx)$, respectively.
Let $\mathcal{H} \equiv \mathcal{H}_{\mathcal{D}, \mathcal{W}, \mathcal{S}, \mathcal{B}}$ be the set of ReLU  neural networks $R_{\phi}$ with parameter $\phi$,
depth $\mathcal{D}$, width $\mathcal{W}$, size $ \mathcal{S}$, and $\|R_{\phi}\|_{\infty} \leq \mathcal{B}.$
We combine the least squares loss (\ref{LSDRa}) with the gradient regularizer (\ref{gp}) as our objective function.
The resulting gradient regularized LSDR estimator of $r=p/q$  is given by
\begin{align}\label{sf}
\widehat{R}_{\phi} \in \arg \min_{R_\phi\in \mathcal{H}}
 &\frac{1}{n}\sum_{i=1}^n [R_{\phi}(X_i)^2- 2 R_{\phi}(Y_i)]+\alpha \frac{1}{n}\sum_{i=1}^n \|\nabla R_{\phi}(X_i)\|^2_{2},
\end{align}
where $\alpha \ge 0$ is a regularization parameter.

\subsection{Estimation error bound}
We first show that the density ratio $r$ is identifiable through the objective function by proving that,  at the population level,
we can recover  the density ratio $r$  via minimizing
  $$\mathfrak{B}^{\alpha}_{\textrm LSDR}(R) = \mathfrak{B}_{\textrm LSDR}(r, R) + \alpha \mathbb{E}_{p} [\Vert \nabla R \Vert_2^2]+ \mathcal{C},$$
where $ \mathfrak{B}_{\textrm LSDR}$ is defined in (\ref{LSDRa})
and $ \mathcal{C} =  \mathbb{E}_{X\sim q} [r^2(X)] - 1.$

\begin{lemma}\label{lem3}
For any $\alpha \geq 0$, we have
$r \in \arg\min_{R} \mathfrak{B}^{\alpha}_{\textrm LSDR}(R).$
In addition, $\mathfrak{B}_{\textrm LSDR}^{\alpha}(R)\ge 0$  for any  $R$ with $\mathbb{E}_{X\sim p}R^2(X) < \infty$,  and $\mathfrak{B}_{\textrm LSDR}^{\alpha}(R)=0$ iff $R(\vx) = r(\vx)  = 1  \ \ (q, p)\text{-}a.e. \  \vx \in \mathbb{R}^m.$
\end{lemma}
{\color{black} This identifiabiity result shows that the target density ratio is the unique minimizer of the population version of the empirical criterion in (\ref{sf}). This provides a the basis for establishing the convergence result of deep nonparametric density-ratio estimation.}

Next we bound the nonparametric estimation error  $\|\widehat{R}_{\phi} - r\|_{L^2(\nu)}$  under the assumptions that the support of $\nu$ is concentrated on a compact low-dimensional manifold and $r$ is Lipsichiz continuous.
Let $\mathfrak{M } \subseteq[-c,c]^{m}$ be a
Riemannian manifold \citep{lee2010} with dimension  $\mathfrak{m}$, condition number $1 / \tau$,  volume  $\mathcal{V}$,  geodesic covering regularity $\mathcal{R}$,  and
$\mathfrak{m}\ll  \mathcal{M} =
\mathcal{O}\left(\mathfrak{m} \ln (m \mathcal{V} \mathcal{R}/\tau)\right)
\ll m.$
Denote
$\mathfrak{M}_{\epsilon} =\left\{\vx \in[-c,c]^{m} : \inf \{\|\vx-\vy\|_2 : \vy \in \mathfrak{M}\} \leq \epsilon\right\},$  $\epsilon \in(0,1)$.

\begin{theorem}\label{th3}
Assume  $\mathrm{supp}(r) = \mathfrak{M}_{\epsilon}$ and $r(\vx)$ satisfies $|r(\vx)| \le B$ for a finite constant $B > 0$ and is Lipschitz continuous with Lipschitz constant $L$.
Suppose the topological  parameter of  $\mathcal{H}_{\mathcal{D}, \mathcal{W}, \mathcal{S}, \mathcal{B}}$ in (\ref{sf}) with $\alpha = 0$  satisfies $\mathcal{D} = \mathcal{O}(\log n)$, $ \mathcal{W} = \mathcal{O}(n^{\frac{\mathcal{M}}{2(2+\mathcal{M})}}/\log n)$,  $\mathcal{S} = \mathcal{O}(n^{\frac{\mathcal{M}-2}{\mathcal{M}+2}}/\log^4 n)$, and $\mathcal{B} = 2B$.
Then,
\begin{equation*}
\mathbb{E}_{\{X_i,Y_i\}_{i=1}^n} [\|\widehat{R}_{\phi} - r\|_{L^2(\nu)}^2] \leq C (B^2+ cL m \mathcal{M}) n^{-2/(2+ \mathcal{M})},
\end{equation*}
where $C$ is a universal constant.
\end{theorem}

The error bound established in Theorem \ref{th3} for the nonparametric deep density-ratio fitting is new. This result is of independent interest for nonparametric estimation with deep neural networks. The above derived rate  $\mathcal{O}(n^{-\frac{2}{2+ \mathcal{M}\ln m}})$  is faster than the optimal rate of convergence for nonparametric estimation of a Lipschitz target in $\mathbb{R}^{m},$
where the optimal rate is $\mathcal{O}(n^{-\frac{2}{2+ m}})$ \citep{stone1982optimal,schmidt2017nonparametric} as long as the intrinsic dimension $\mathcal{M}$ of the data is much smaller than the ambient dimension $m$. Therefore, the proposed density-ratio estimators circumvent the ``curse of dimensionality" if data is supported on a lower-dimensional manifold. Low-dimensional structure of complex data is a common phenomenon in image analysis, computer vision and natural language processing.

\section{Implementation} 
\label{implement}
We now described how to implement EPT and train the optimal transport $\cT$ with an i.i.d. sample
$\{{X}_i\}_{i=1}^n \subset \mathbb{R}^{m}$  from an unknown target distribution $\nu$.
The EPT map is trained via the forward Euler iteration  (\ref{eu1})-(\ref{eu3}) with a small step size $s > 0.$
The resulting map is a composition of a sequence of
residual maps, i.e., $ \mathcal{T}_K \circ\mathcal{T}_{K-1} \circ ... \circ\mathcal{T}_0$ for a large $K$.
As implied by Theorem \ref{th3} in Section \ref{dr}, each $\mathcal{T}_k, k = 0,...,K$ can be estimated with high accuracy by
$\widehat{\mathcal{T}}_k = \Id + s \hat{\vv}_k,$
where $ \hat{\vv}_k(\vx)= -f^{\prime\prime}(\widehat{R}_{\phi}(\vx))\nabla \widehat{R}_{\phi}(\vx)$.  Here  $\widehat{R}_{\phi}$ is the density-ratio estimator defined in (\ref{sf}) below based on
 $\{Y_i\}_{i = 1}^n \sim q_{k}$ and the data $\{X_i\}_{i=1}^n \sim p$.
Therefore, according to the EPT map (\ref{ETPa}), the particles
$$\widehat{\cT}(\tilde{Y}_i) \equiv \widehat{\mathcal{T}}_K \circ\widehat{\mathcal{T}}_{K-1} \circ ... \circ\widehat{\mathcal{T}}_0(\tilde{Y}_i), i = 1, \ldots, n
$$
serve as samples drawn from the target distribution $\nu$, where particles $\{\tilde{Y}_i\}_{i=1}^n \subset \mathbb{R}^{m}$ are sampled from a simple reference distribution $\mu$.

In many applications, high-dimensional complex data such as images, texts and natural languages, tend to have low-dimensional latent features. To learn generative models with latent low-dimensional structures, it is beneficial to have the option of first sampling particles $\{Z_i\}_{i=1}^n$ from
a low-dimensional reference distribution $\tilde{\mu} \in \mathcal{P}_2(\mathbb{R}^{\ell})$ with $\ell \ll d$.
Then we apply  $\widehat{\cT}$
to particles $\tilde{Y}_i = G_{\theta}(Z_i), i = 1,...,n$, where we introduce another deep neural network $G_{\theta}:\mathbb{R}^{\ell} \rightarrow \mathbb{R}^{m}$ with parameter $\theta$. We can estimate  $G_{\theta}$ via fitting  the pairs $\{(Z_i, \tilde{Y}_i)\}_{i=1}^n$.   We describe the EPT algorithm below.
\begin{itemize}
\item \textbf{Outer loop for  modeling low dimensional latent structure (optional)}
	\begin{itemize}
	\item  Sample  $\{Z_i\}_{i=1}^n\subset \mathbb{R}^{\ell}$  from a low-dimensional reference distribution $\tilde{\mu}$ and let  $\tilde{Y}_i = G_{\theta}(Z_i), i =1,2,\ldots,n$.
	\item \textbf{Inner loop for finding the push-forward map}
		\begin{itemize}
        \item If there are no outer loops, sample $\tilde{Y}_i \sim \mu, i =1,\ldots,n$.
		\item   Get $\hat{\vv}(\vx) = -f^{\prime\prime}(\widehat{R}_{{\phi}}(\vx))\nabla \widehat{R}_{{\phi}}(\vx)$ via solving (\ref{sf}) below with $Y_i = \tilde{Y}_i$.
       Set $\widehat{\mathcal{T}} = \Id + s \hat{\vv}$ with a small step size $s$.
		\item  Update the particles  $\tilde{Y}_i = \widehat{\mathcal{T}}(\tilde{Y}_i)$,   $i =1,\ldots ,n$.
		\end{itemize}
	\item \textbf{End inner loop}
	\item  If there are outer loops,
update the parameter $\theta$ of $G_\theta(\cdot)$ via solving   $\min_{\theta}\sum_{i=1}^n \|G_{\theta}(Z_i) - \tilde{Y}_i\|_{2}^2/n$.
    \end{itemize}
\item \textbf{End outer loop}
\end{itemize}

\section{Related Work}\label{relate}
We discuss connections between EPT and the existing related works.
The existing generative models, such as VAEs, GANs and flow-based methods, parameterize a transform map with a neural network, say $G$, that solves
\begin{equation}\label{ul}
 \min_{G} \mathfrak{D}(G_{\#} \mu,\nu),
 \end{equation} where $\mathfrak{D}(\cdot, \cdot)$ is an integral probability discrepancy.
The original GAN \citep{goodfellow14}, $f$-GAN \citep{nowozin16}  and WGAN \citep{arjovsky17} solve the dual form of  (\ref{ul}) by parameterizing the dual variable using another neural network with  $\mathfrak{D}$ as the JS-divergence, the $f$-divergence and the $1$-Wasserstein distance, respectively.
Based on the fact that the $1$-Wasserstein distance can be evaluated from  samples via linear programming \citep{srip12},
\cite{liu2018two} and  \cite{genevay2018learning} proposed training the primal form of WGAN  via a two-stage method that solves the linear programm.
SWGAN \citep{deshpande2018generative} and MMDGAN \citep{li17,binkowski18} use the sliced quadratic Wasserstein  distance and the maximum mean discrepancy  (MMD) as $\mathfrak{D}$, respectively.

Vanilla VAE \citep{kingma14} approximately solves the primal form of (\ref{ul}) with the KL-divergence loss  under the framework of  variational inference. Several authors have proposed methods that use optimal transport losses, such as various forms of Wasserstein  distances between the distribution of learned latent codes and the prior distribution as the regularizer in VAE to improve performance. These methods include WAE \citep{tolstikhin2018wasserstein}, Sliced WAE \citep{modelsliced} and Sinkhorn AE \citep{patrinisinkhorn}.

Discrete time  flow-based methods minimize (\ref{ul}) with the KL divergence loss \citep{rezende2015variational,dinh2014nice,dinh2016density,
kingma2016improved,papamakarios2017masked,kingma2018glow}. \cite{grathwohl2019ffjord} proposed an ODE flow approach for fast training in such methods using the adjoint  equation \citep{chen2018neural}.
By introducing the optimal transport tools into maximum likelihood training, \cite{chen2018continuous} and \cite{zhang2018monge}    considered continuous time flow. \cite{chen2018continuous} proposed a gradient flow in measure spaces in the framework of variational inference and  then discretized it with the implicit movement minimizing scheme  \citep{de1993new,jordan1998variational}.
  \cite{zhang2018monge} considered  gradient flows in measure spaces with time invariant velocity  fields.
CFGGAN \citep{johnson18}
derived from the perspective of optimization in the functional space
is a special form of EPT with $\mathcal{L}[\cdot]$ taken as the KL divergence. SW flow \citep{liutkus2019sliced} and MMD flow   \citep{arbel2019maximum} are gradient flows in measure spaces.
MMD flow can be recovered from  EPT  by first choosing $\mathcal{L}[\cdot]$ as the Lebesgue norm  and then projecting the corresponding velocity vector fields onto reproducing kernel Hilbert spaces, please see Appendix \ref{rmmd} for a proof.
However, neither SW flow nor MMD flow can model hidden low-dimensional structure with the particle sampling procedure.

{\color{black}
SVGD in \citep{liu2017} and the proposed EPT are both particle methods based on gradient flow in measure spaces. However, the SVGD
samples from an un-normalized density, while EPT focuses on generative leaning, i.e., learning the distribution from samples.  At the population level, projecting the velocity fields of EPT with KL divergence onto reproducing kernel Hilbert Spaces will recover the velocity fields of SVGD.
{\color{black} The proof is given in Appendix \ref{rsvgd}.}
Score-based methods in \citep{yang2019,yang2020,ho2020}
are also particle methods based on unadjusted Langevin flow and deep score estimators. At the population level, the velocity fields of these score-based methods are random since they have a Brownian motion term, while the velocity fields of EPT are deterministic. At the sample level, these score-based methods  need to learn a vector-valued deep score function.
while in EPT we only need to estimate the density ratios which are scalar functions.
}

\section{Experiments}\label{numerics}

The implementation details on numerical settings, network structures, SGD optimizers and  hyper-parameters are given in the appendix. All experiments are performed using NVIDIA Tesla K80 GPUs. The PyTorch code of EPT is available at \url{https://github.com/anonymous/EPT}.

\subsection{2D simulated data}
\begin{figure}[ht!]
\begin{minipage}[t]{5in}
\centering
\includegraphics[width=0.6in,height=0.6in]{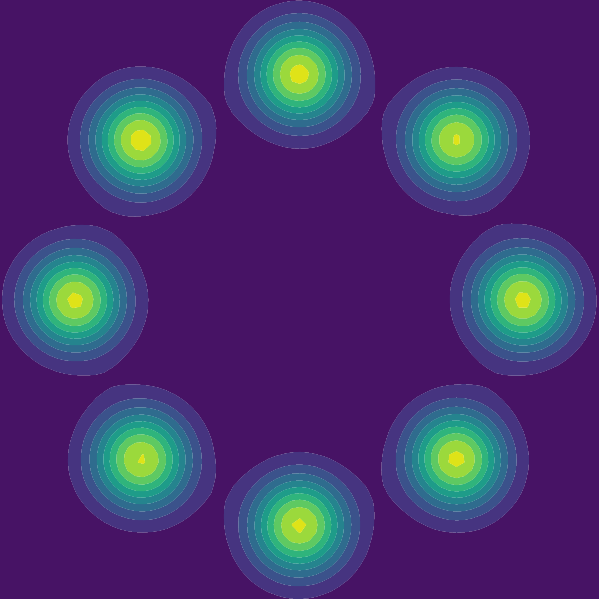}
\includegraphics[width=0.6in,height=0.6in]{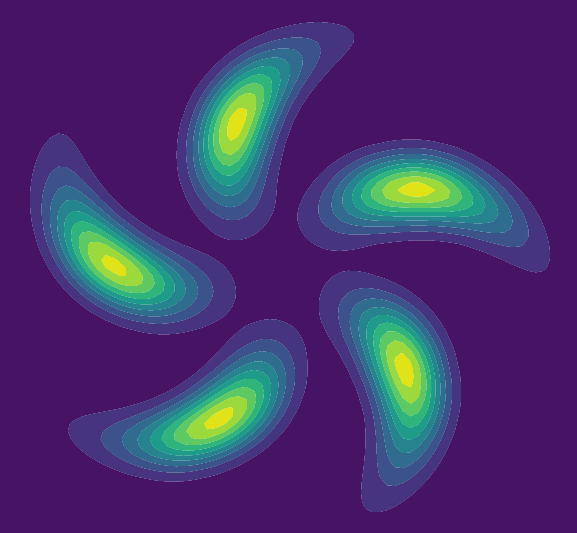}
\includegraphics[width=0.6in,height=0.6in]{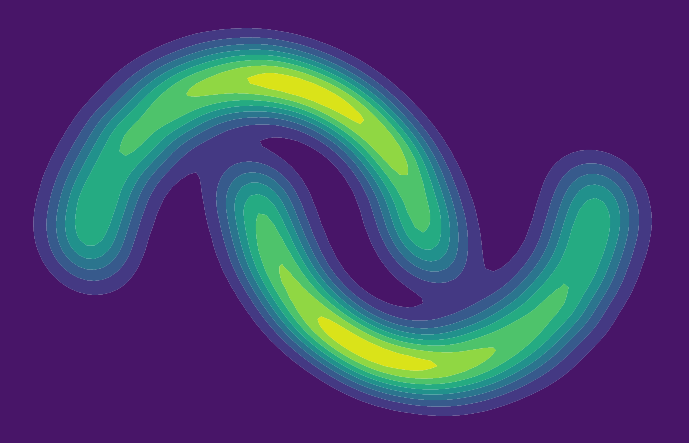}
\includegraphics[width=0.6in,height=0.6in]{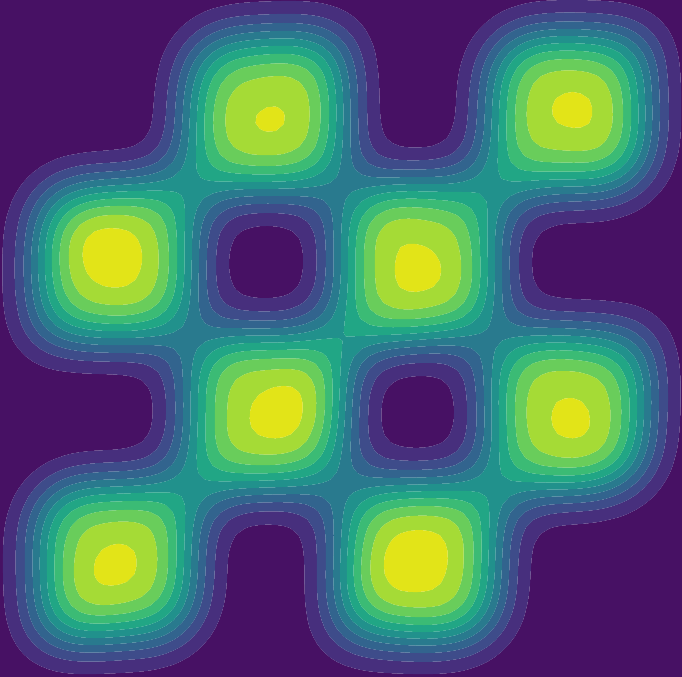}
\includegraphics[width=0.6in,height=0.6in]{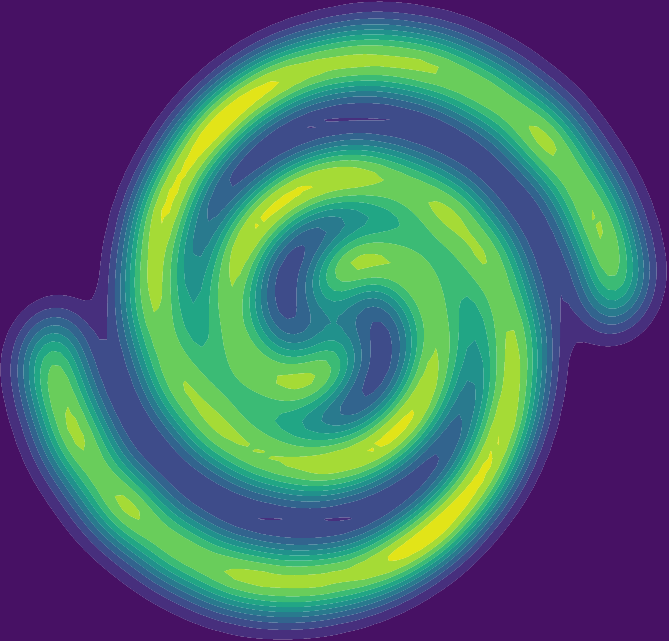}
\includegraphics[width=0.6in,height=0.6in]{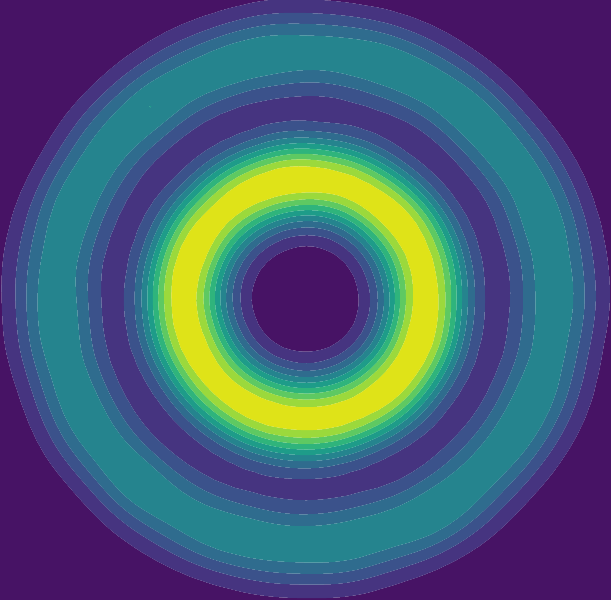}
\end{minipage}
\begin{minipage}[t]{5in}
\centering
\includegraphics[width=0.6in,height=0.6in]{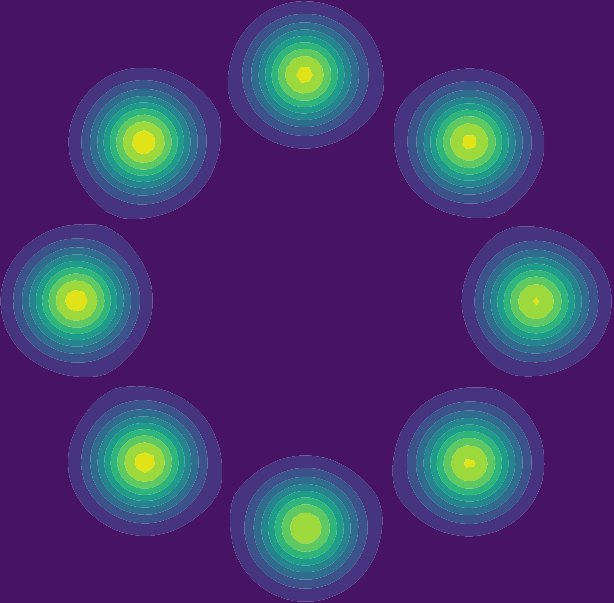}
\includegraphics[width=0.6in,height=0.6in]{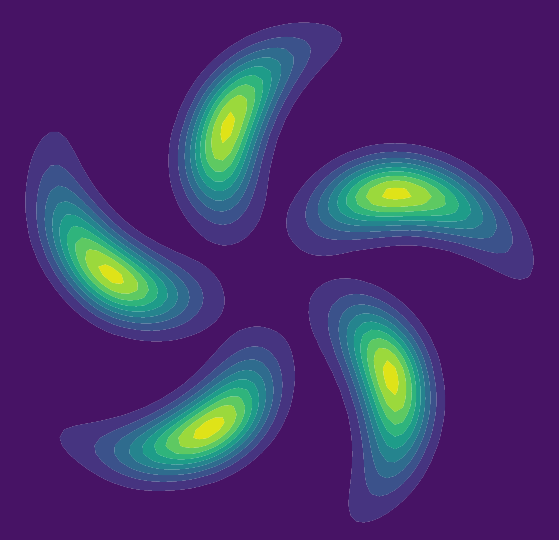}
\includegraphics[width=0.6in,height=0.6in]{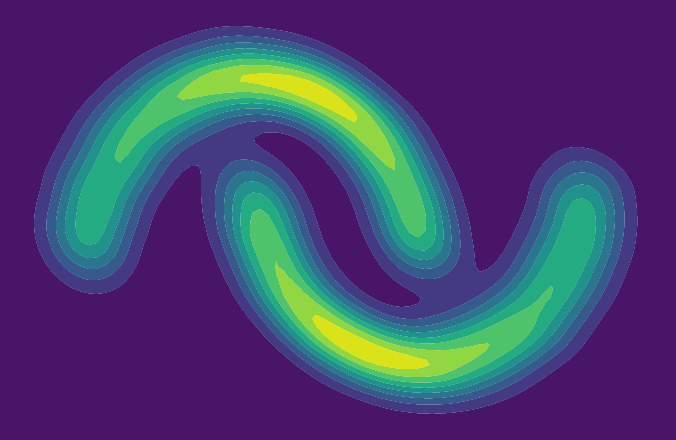}
\includegraphics[width=0.6in,height=0.6in]{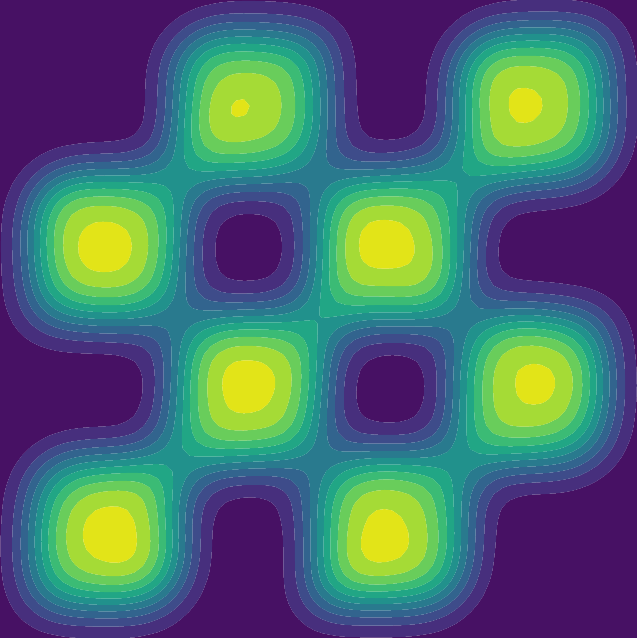}
\includegraphics[width=0.6in,height=0.6in]{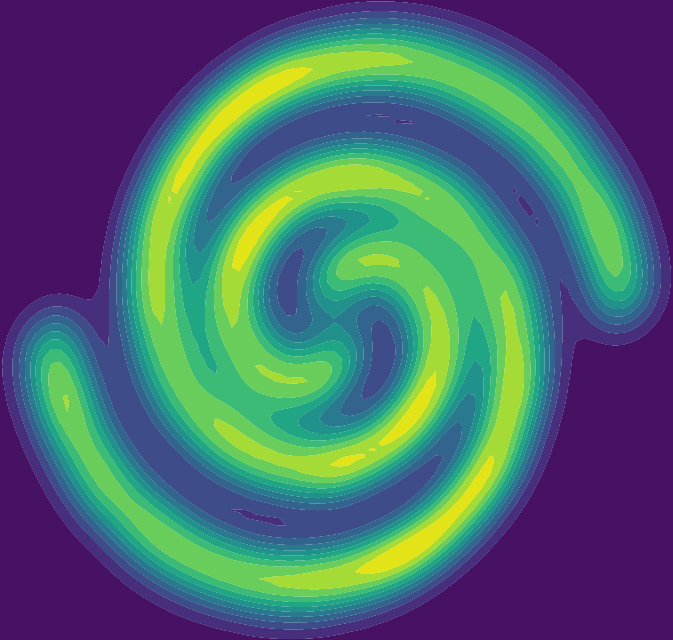}
\includegraphics[width=0.6in,height=0.6in]{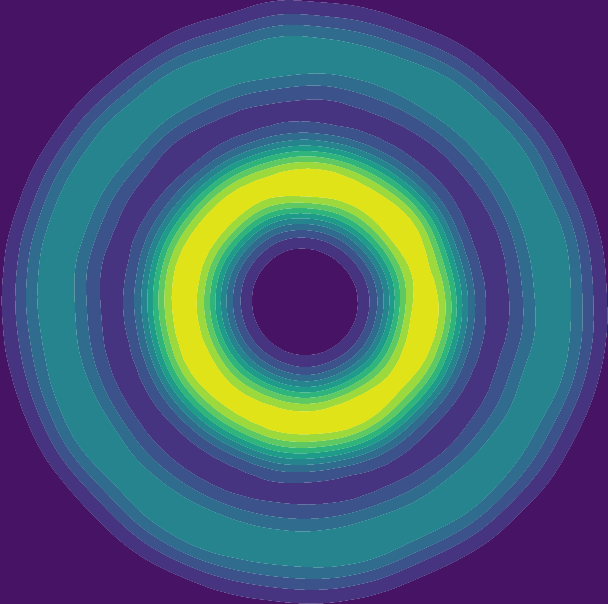}
\end{minipage}
\begin{minipage}[t]{5in}
\centering
\includegraphics[width=0.6in,height=0.6in]{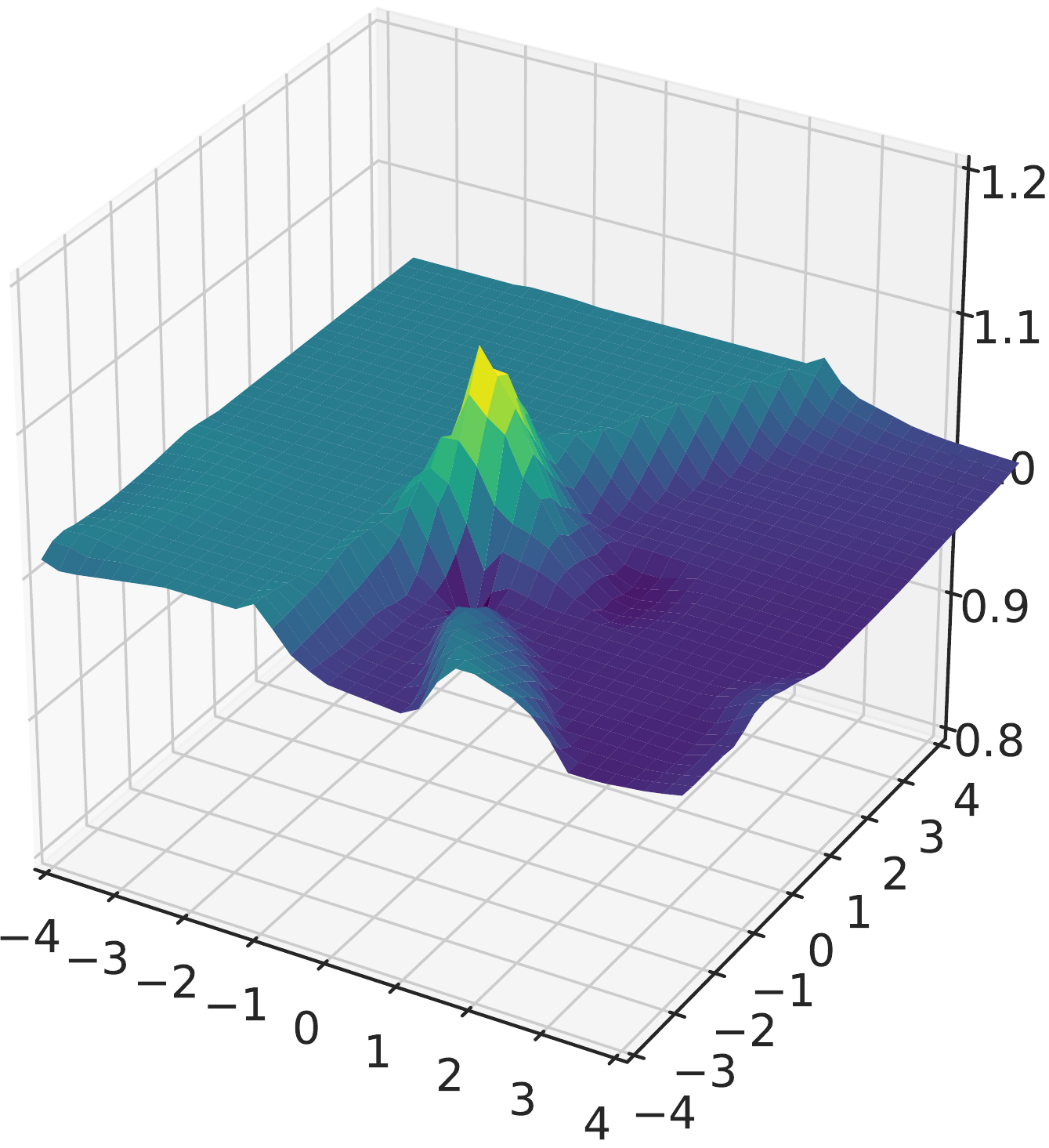}
\includegraphics[width=0.6in,height=0.6in]{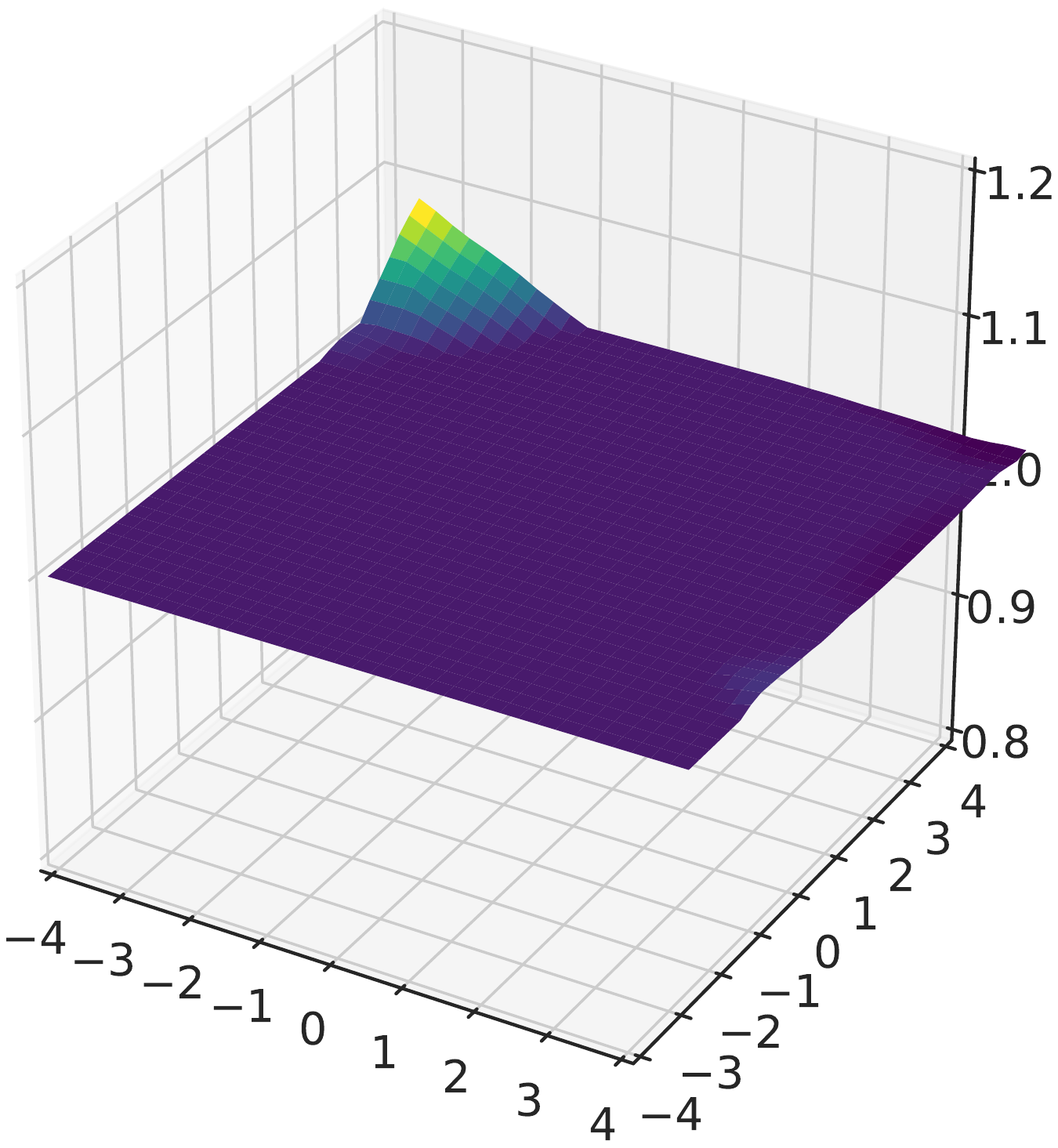}
\includegraphics[width=0.6in,height=0.6in]{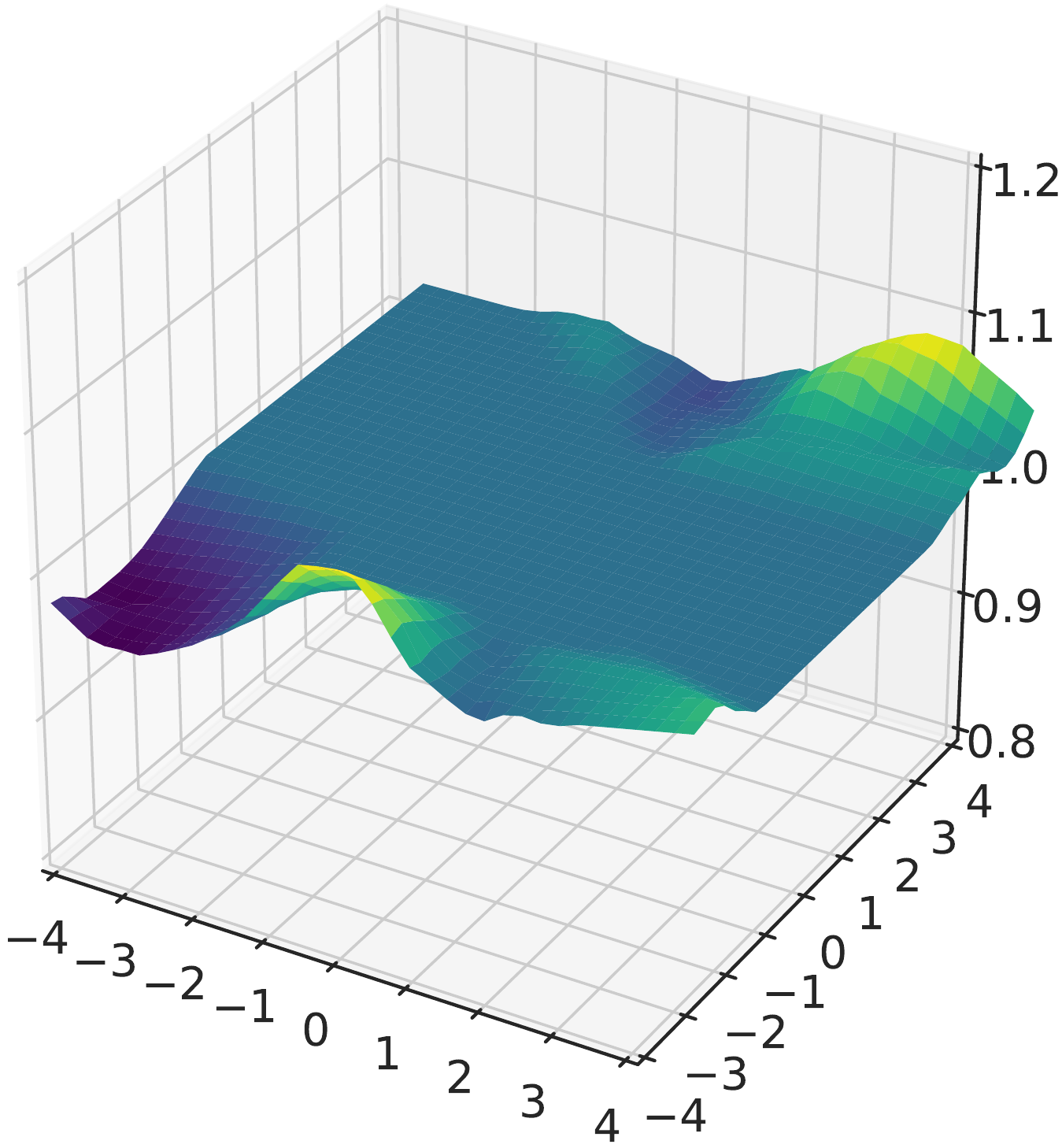}
\includegraphics[width=0.6in,height=0.6in]{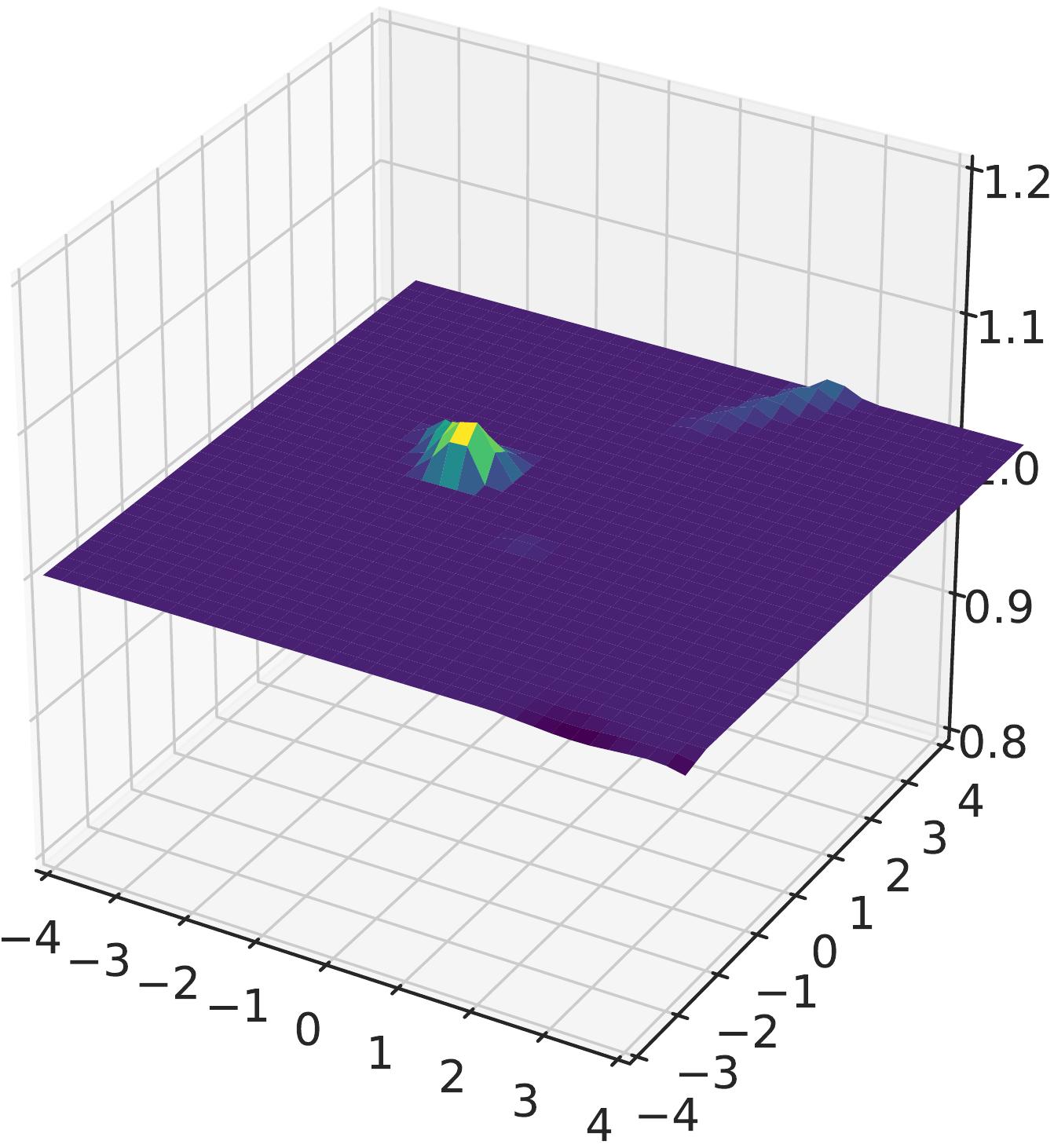}
\includegraphics[width=0.6in,height=0.6in]{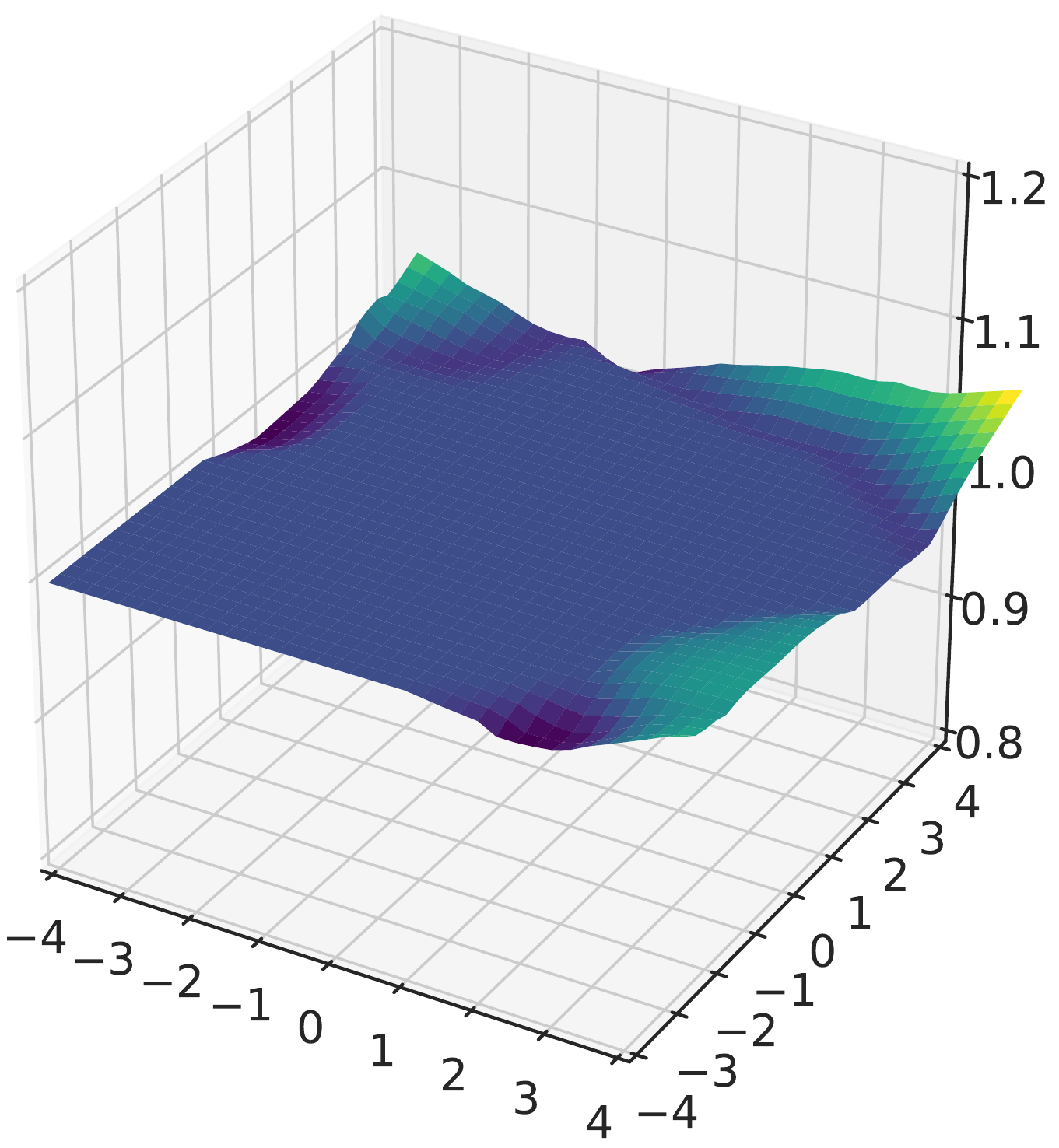}
\includegraphics[width=0.6in,height=0.6in]{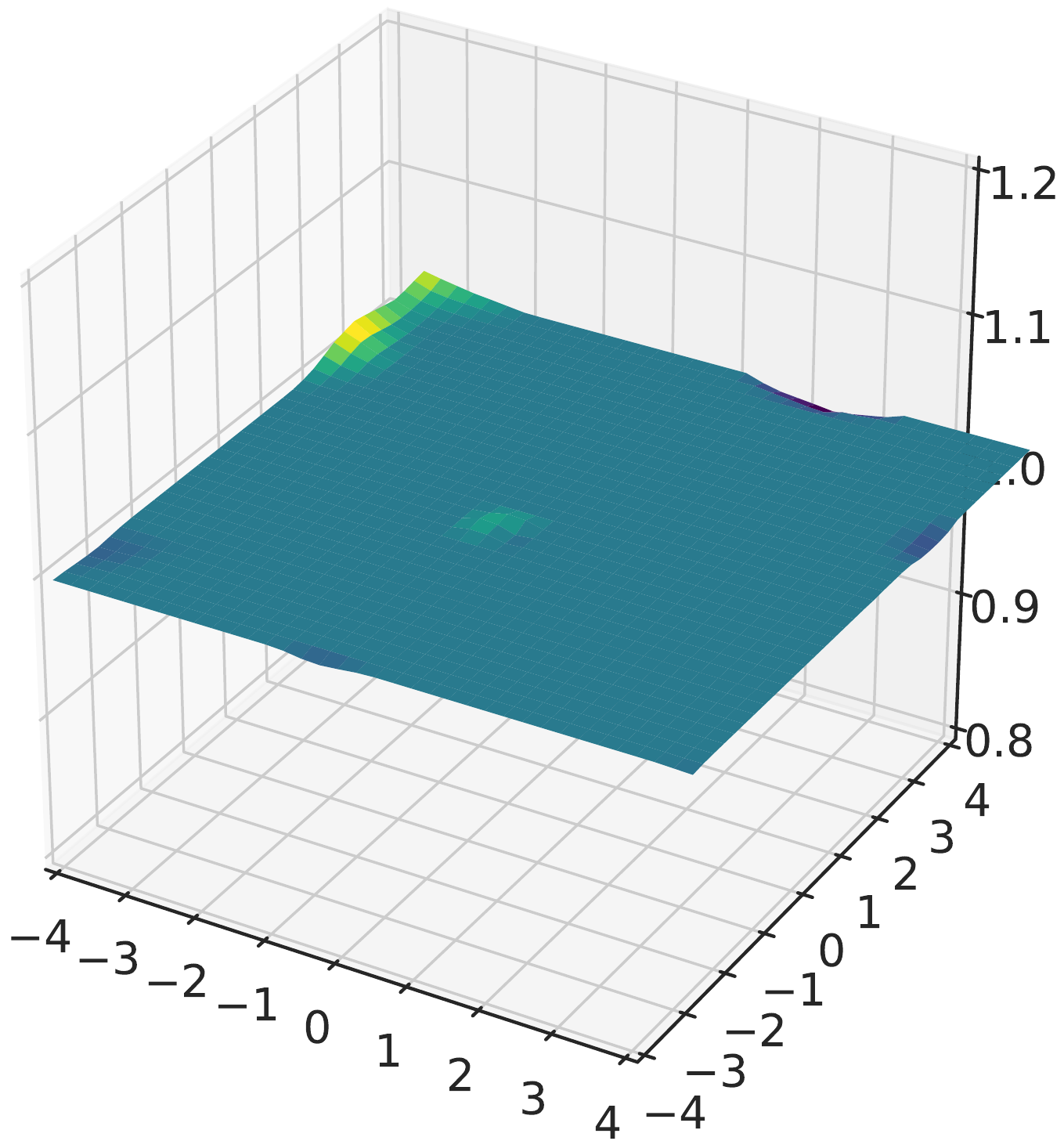}
\end{minipage}
\begin{minipage}[t]{5in}
\centering
\includegraphics[width=0.6in,height=0.6in]{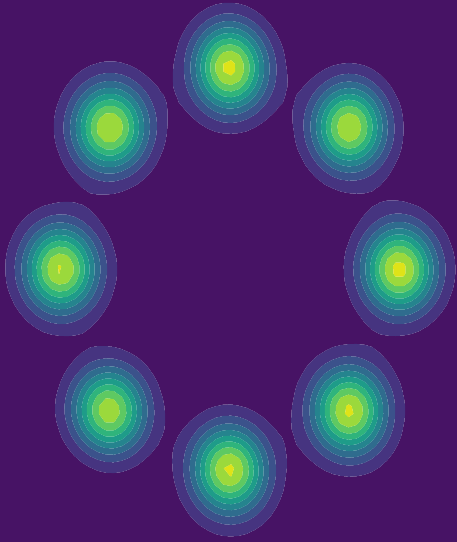}
\includegraphics[width=0.6in,height=0.6in]{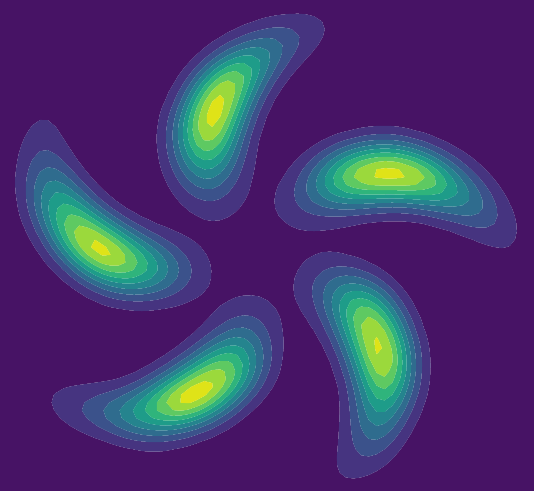}
\includegraphics[width=0.6in,height=0.6in]{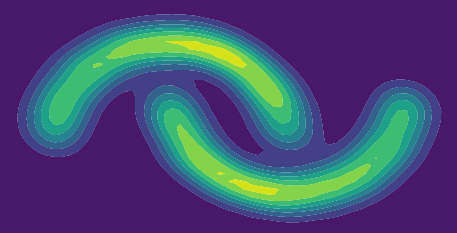}
\includegraphics[width=0.6in,height=0.6in]{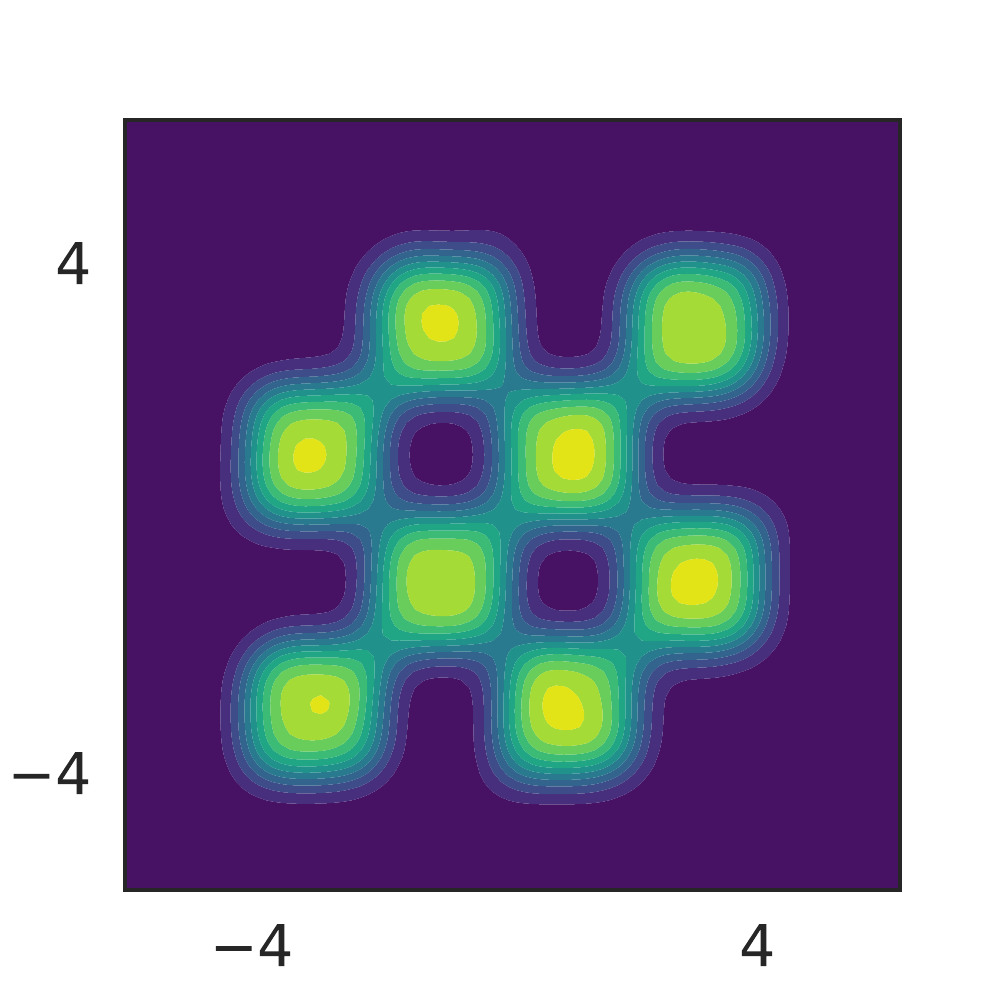}
\includegraphics[width=0.6in,height=0.6in]{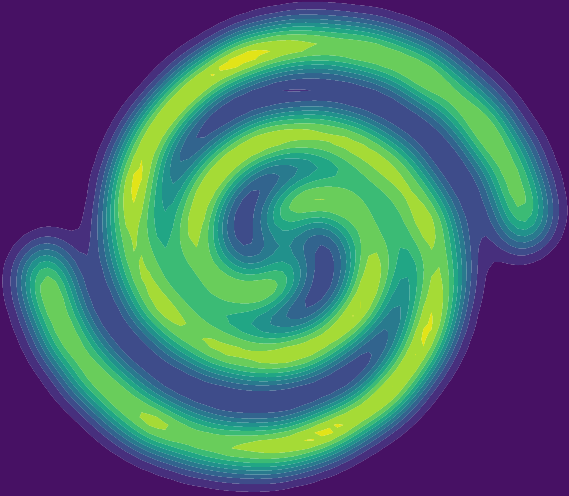}
\includegraphics[width=0.6in,height=0.6in]{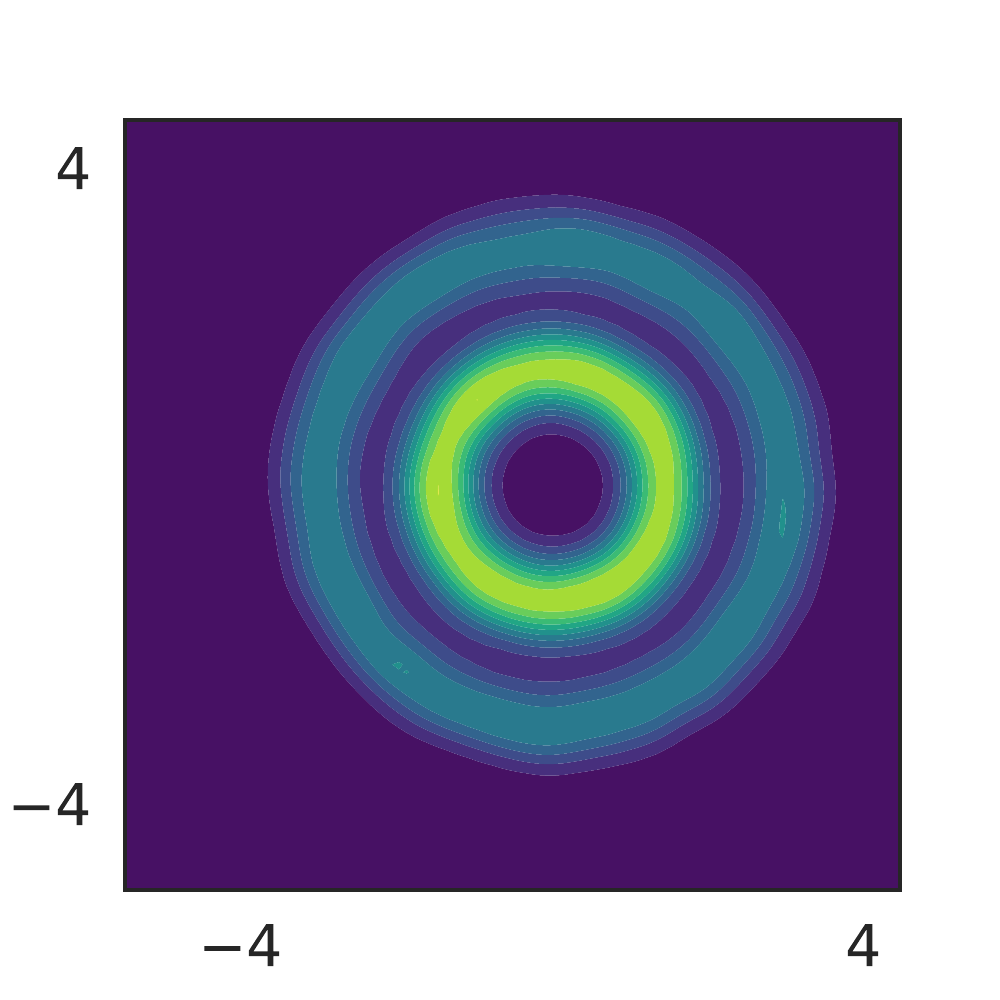}
\end{minipage}
\begin{minipage}[t]{5in}
\centering
\includegraphics[width=0.6in,height=0.6in]{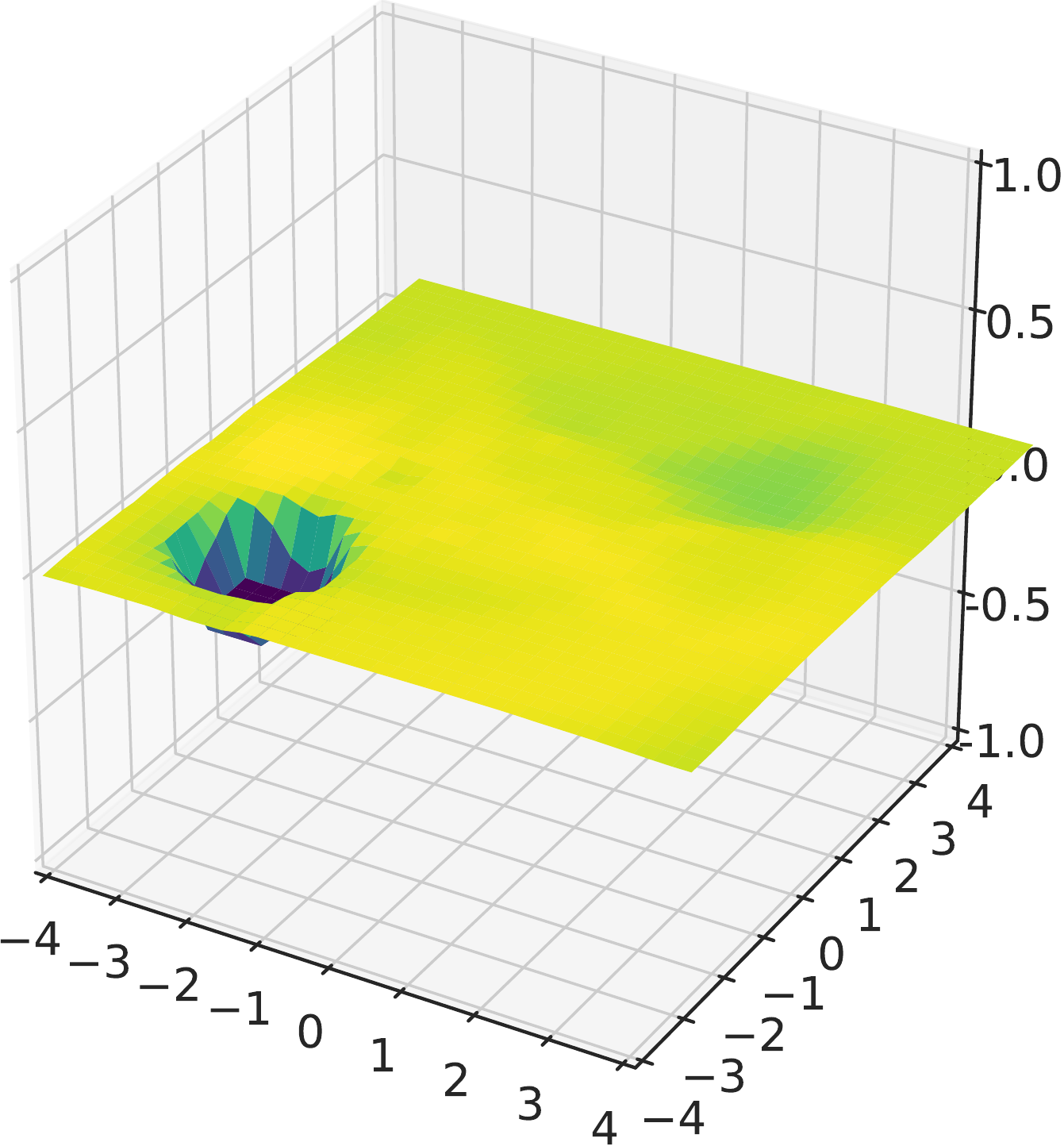}
\includegraphics[width=0.6in,height=0.6in]{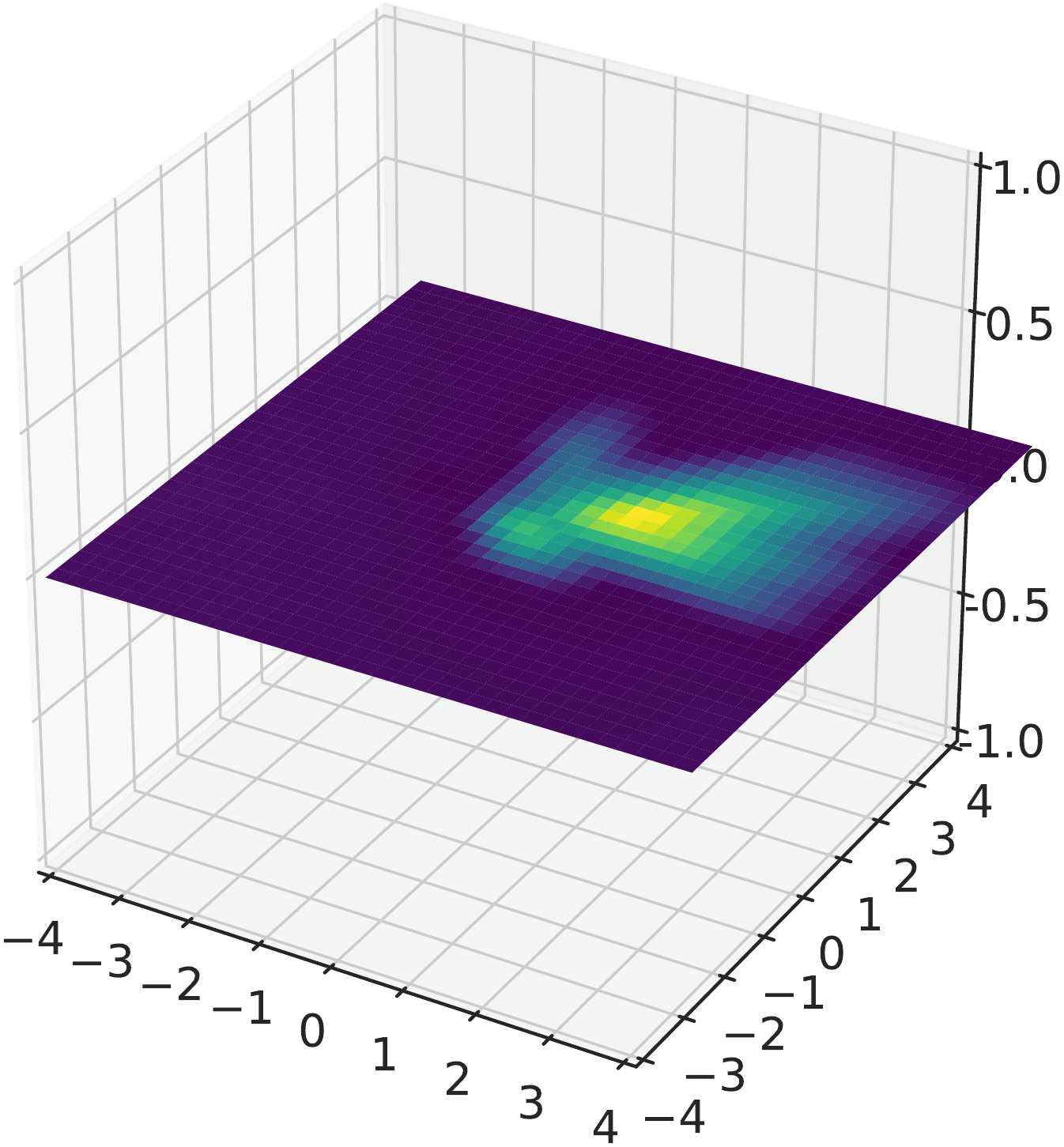}
\includegraphics[width=0.6in,height=0.6in]{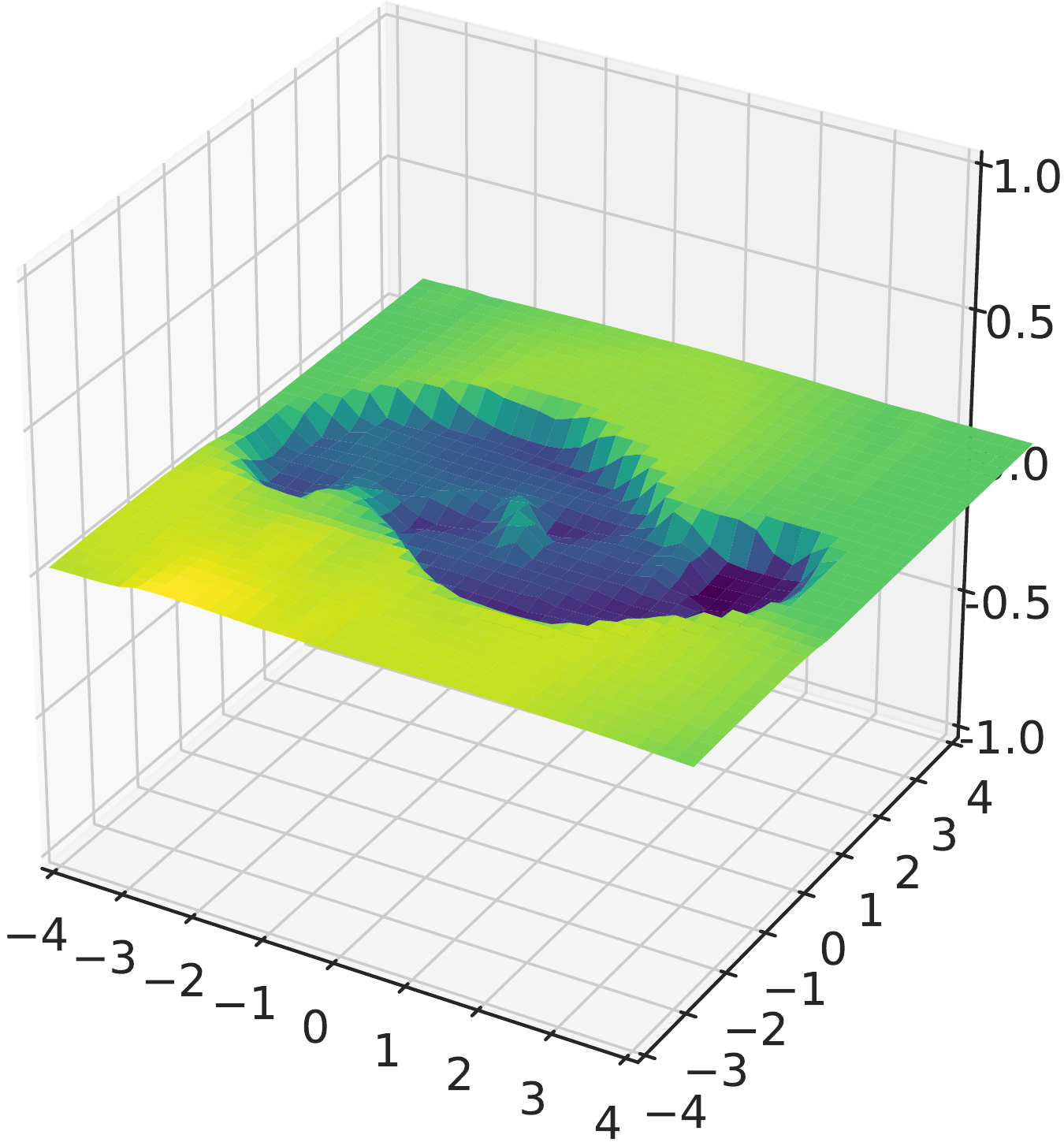}
\includegraphics[width=0.6in,height=0.6in]{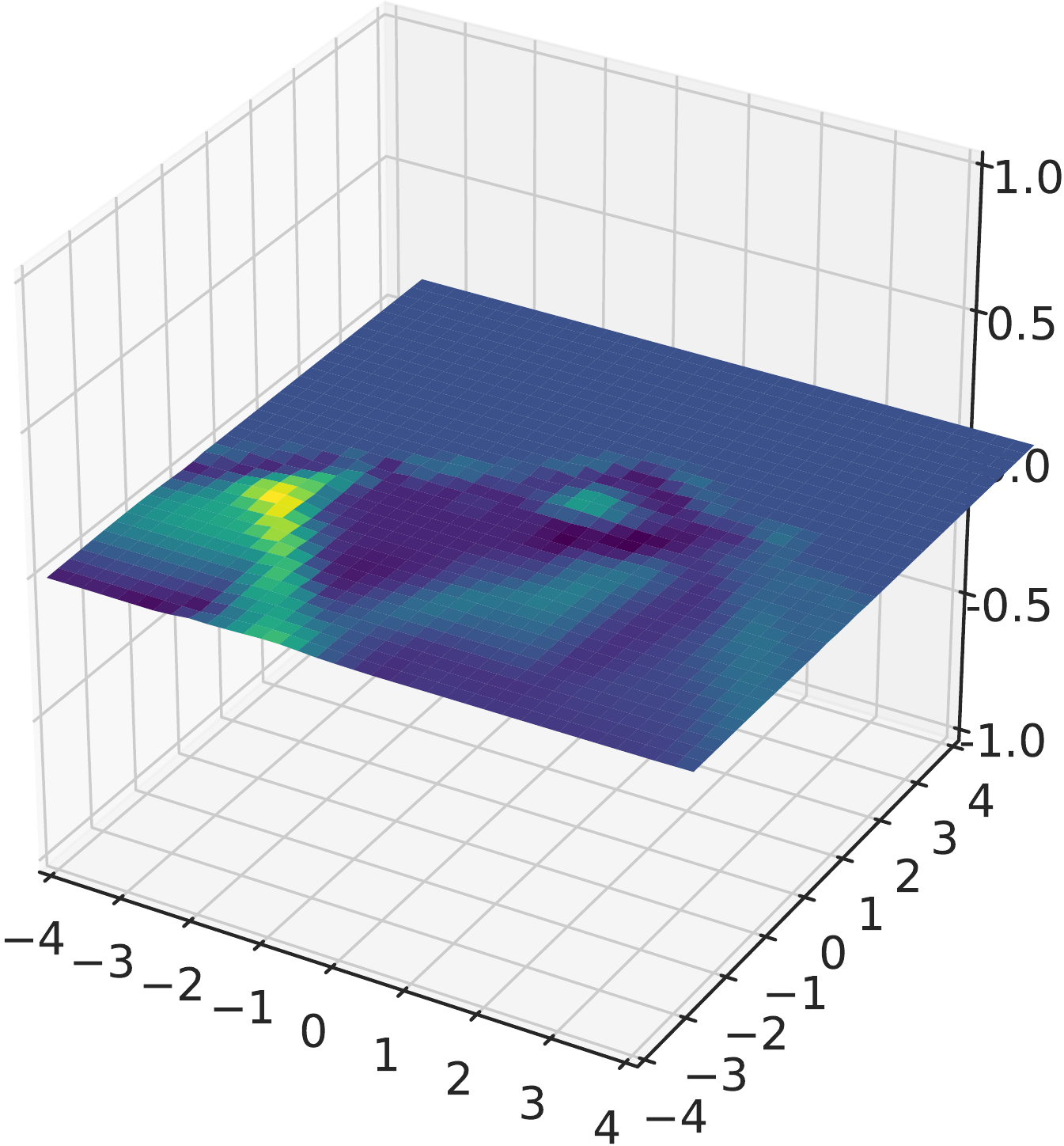}
\includegraphics[width=0.6in,height=0.6in]{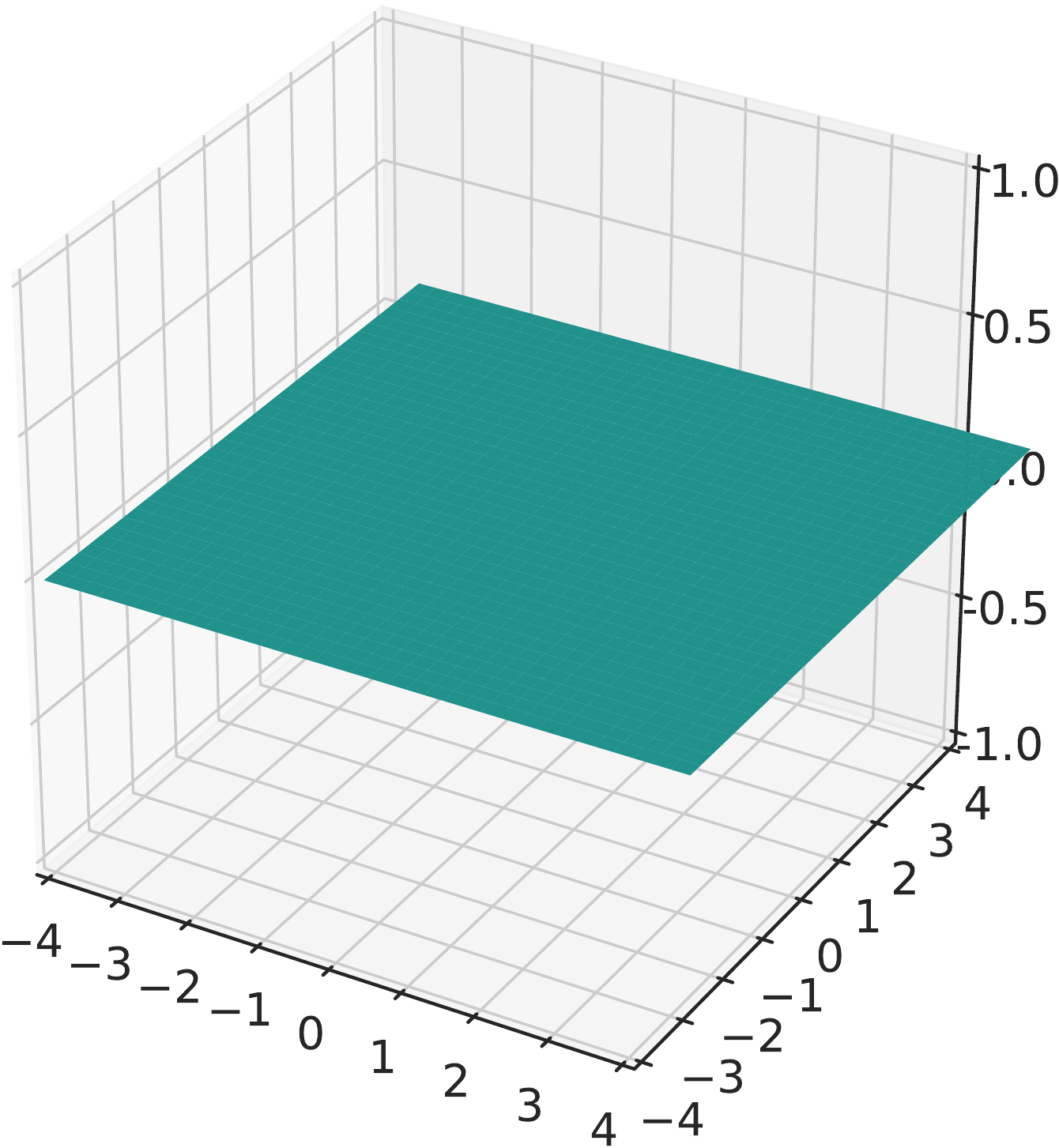}
\includegraphics[width=0.6in,height=0.6in]{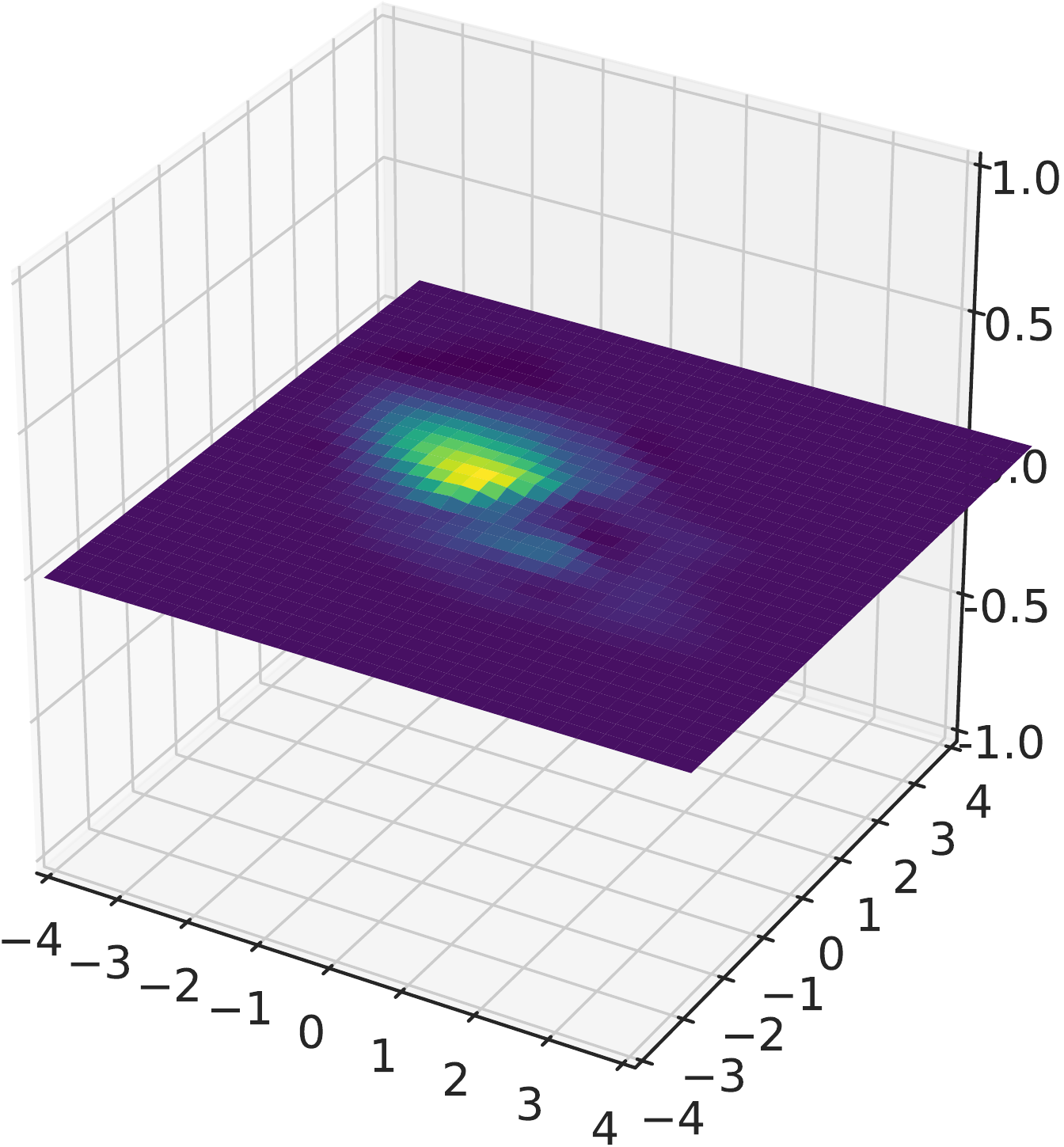}
\end{minipage}
\centering
\caption{KDE plots of the target samples (the first row).
 The second and third rows show
the KDE plots of the learned  samples via  EPT with $f$-divergence
and the surface plots of estimated density ratios after 20k iterations. The fourth and fifth rows show the KDE plots of
the learned sample via EPT with Lebesgue norm of the density difference after 20k iterations. }
\label{kde}
\end{figure}

We use  EPT to learn 2D  distributions  adapted from \cite{grathwohl2019ffjord} with multiple modes and density ridges.
The first row  in Figure \ref{kde} shows kernel density estimation (KDE) plots of 50k samples from target distributions  including (from  left to  right) \emph{8Gaussians, pinwheel, moons, checkerboard, 2spirals,} and \emph{circles}.
The second and third rows show
the KDE plots of the learned  samples via  EPT with $f$-divergence
and the surface plots of estimated density ratios after 20k iterations. The fourth and fifth rows show the KDE plots of
the learned sample via EPT with Lebesgue norm of the density difference.
Clearly, the generated samples via EPT are nearly indistinguishable from those of the target samples and the estimated density-ratio/ difference  functions are approximately equal to 1/0, indicating the learnt distribution  matches the target well.

\begin{figure}[hpt]
\centering{}
\subfigure[\textbf{Left} two figures:  Maps learned without gradient penalty. \textbf{Right} two figures:  Maps learned with  gradient penalty.]{
\begin{minipage}[t]{3.30in}
\includegraphics[width=0.8in,height=0.8in]{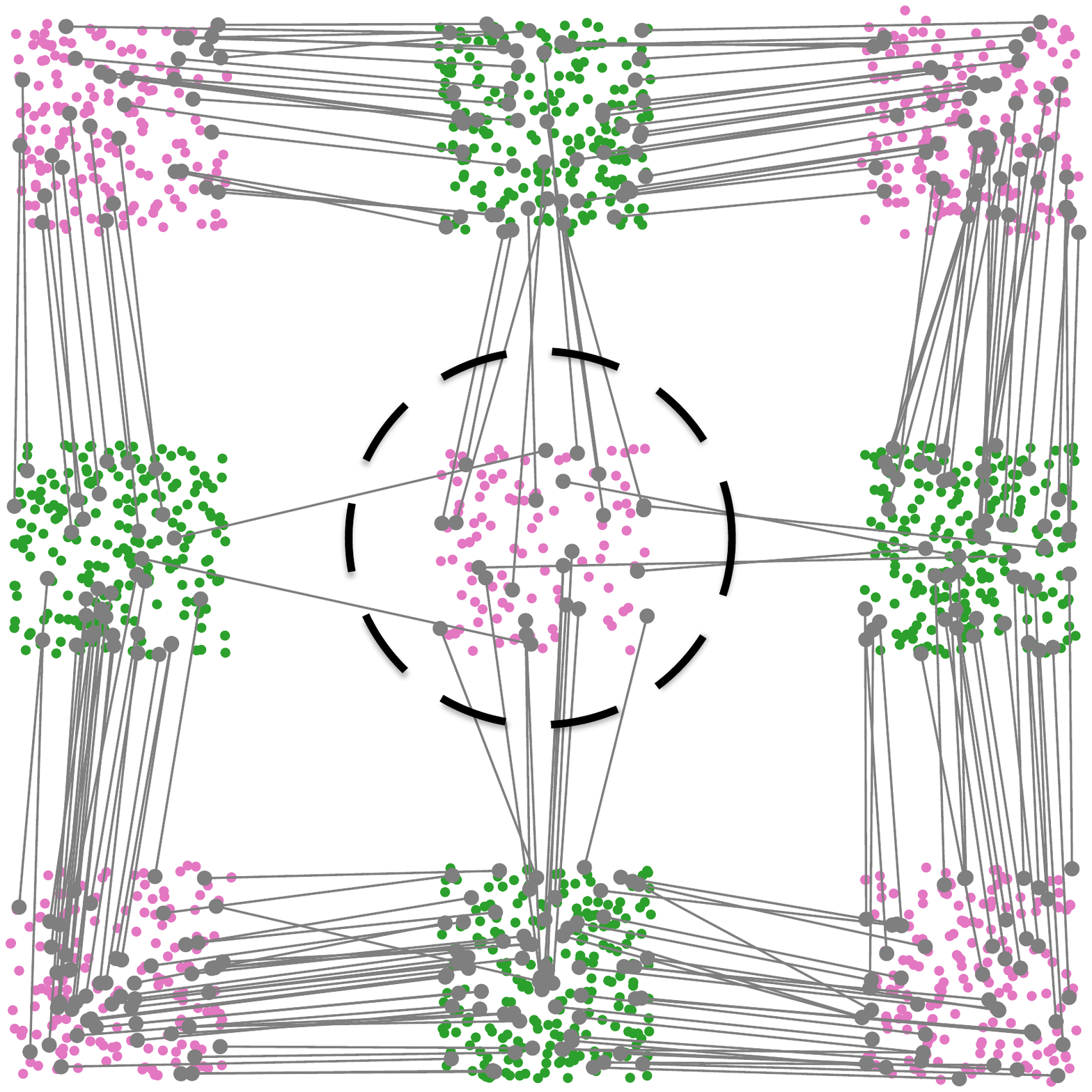}
\includegraphics[width=0.8in,height=0.8in]{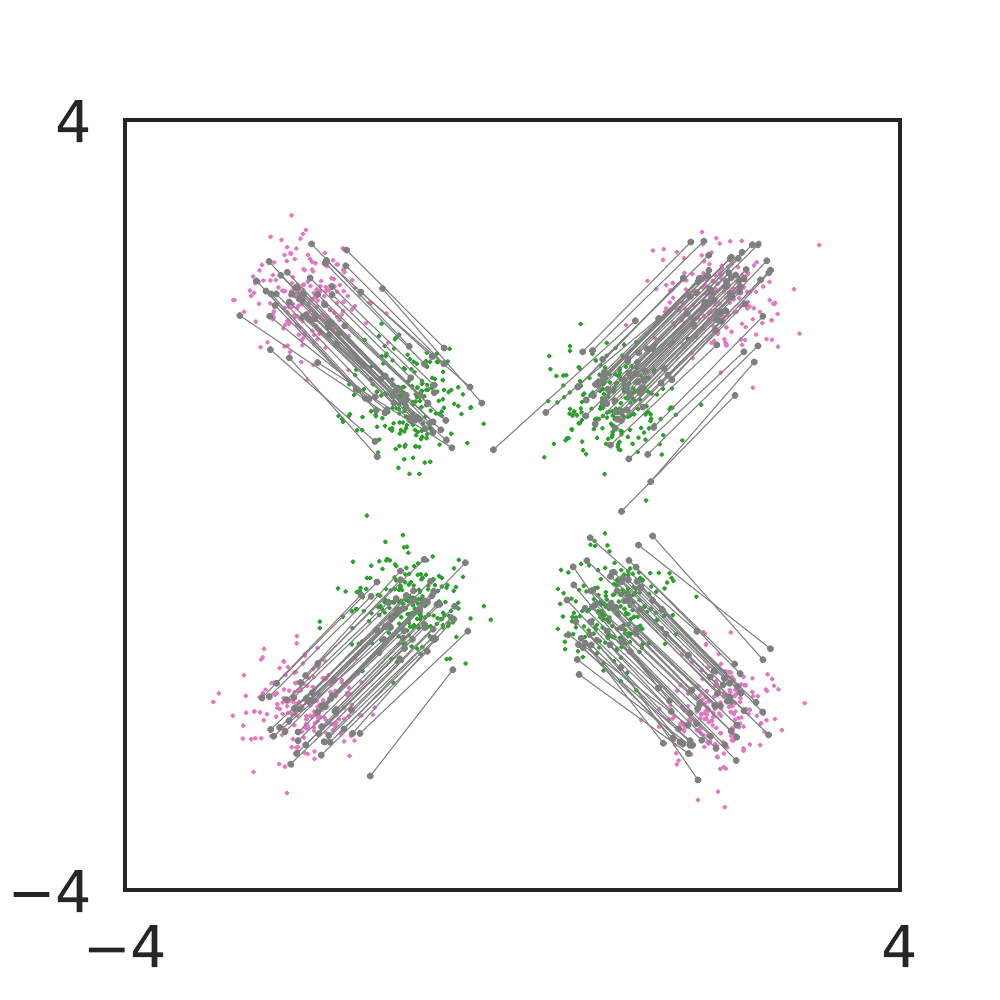}
\includegraphics[width=0.8in,height=0.8in]{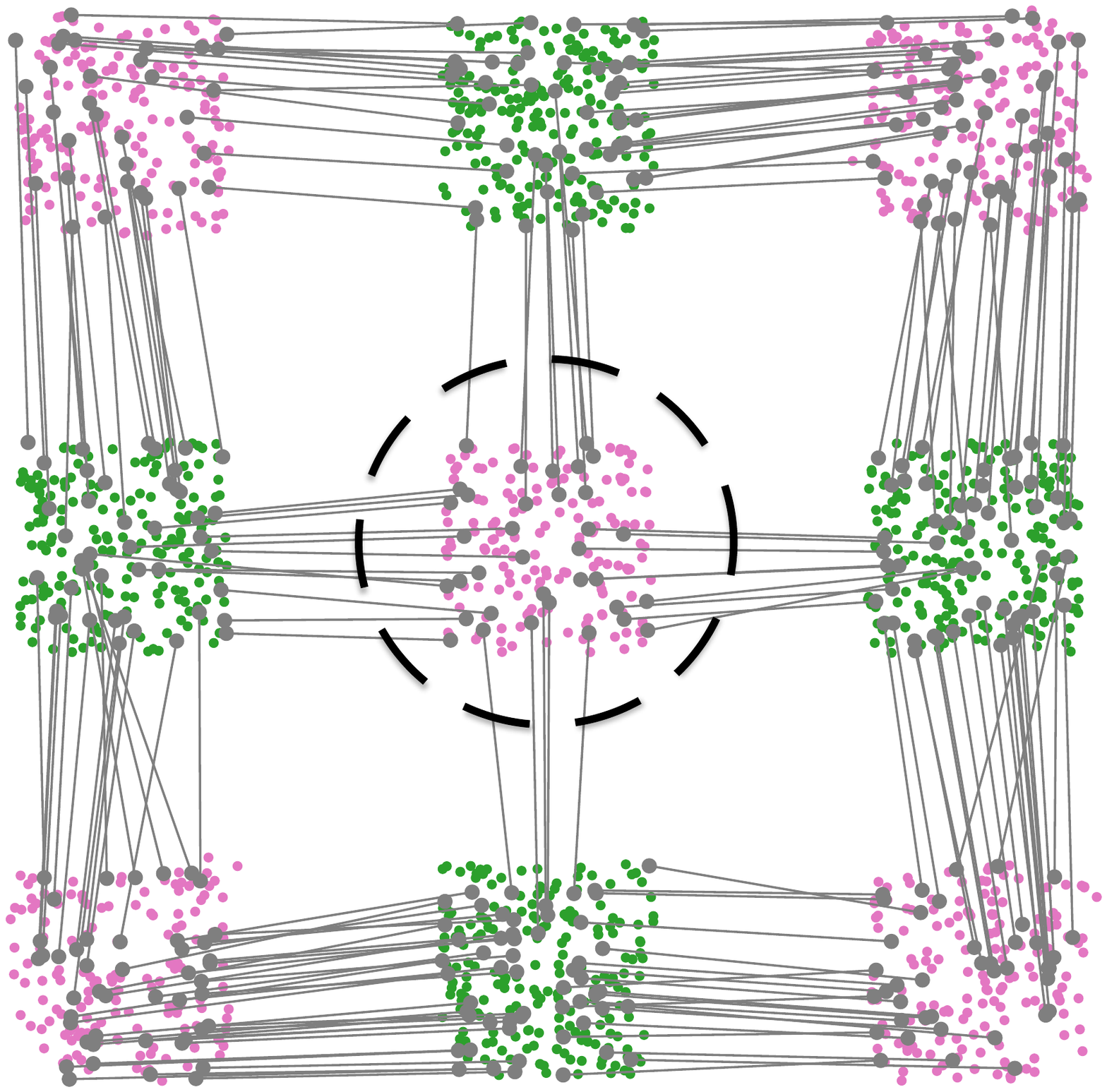}
\includegraphics[width=0.8in,height=0.8in]{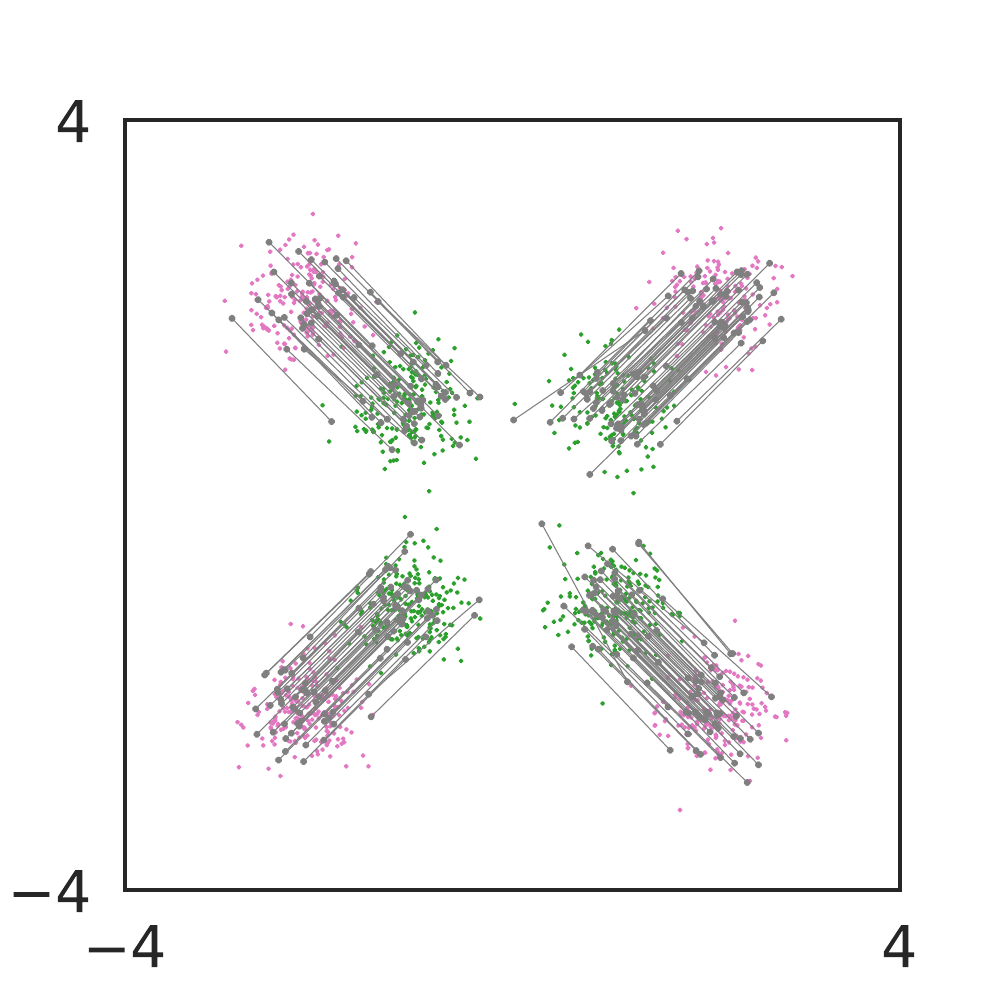}
\end{minipage}
\label{map-map}
}
\subfigure[\textbf{Left} two figures:  Surface plots of estimated density-ratio without gradient penalty. \textbf{Right} two figures:  Surface plots of estimated density-ratio with  gradient penalty.]{
\begin{minipage}[t]{3.30in}
\includegraphics[width=0.8in,height=0.8in]{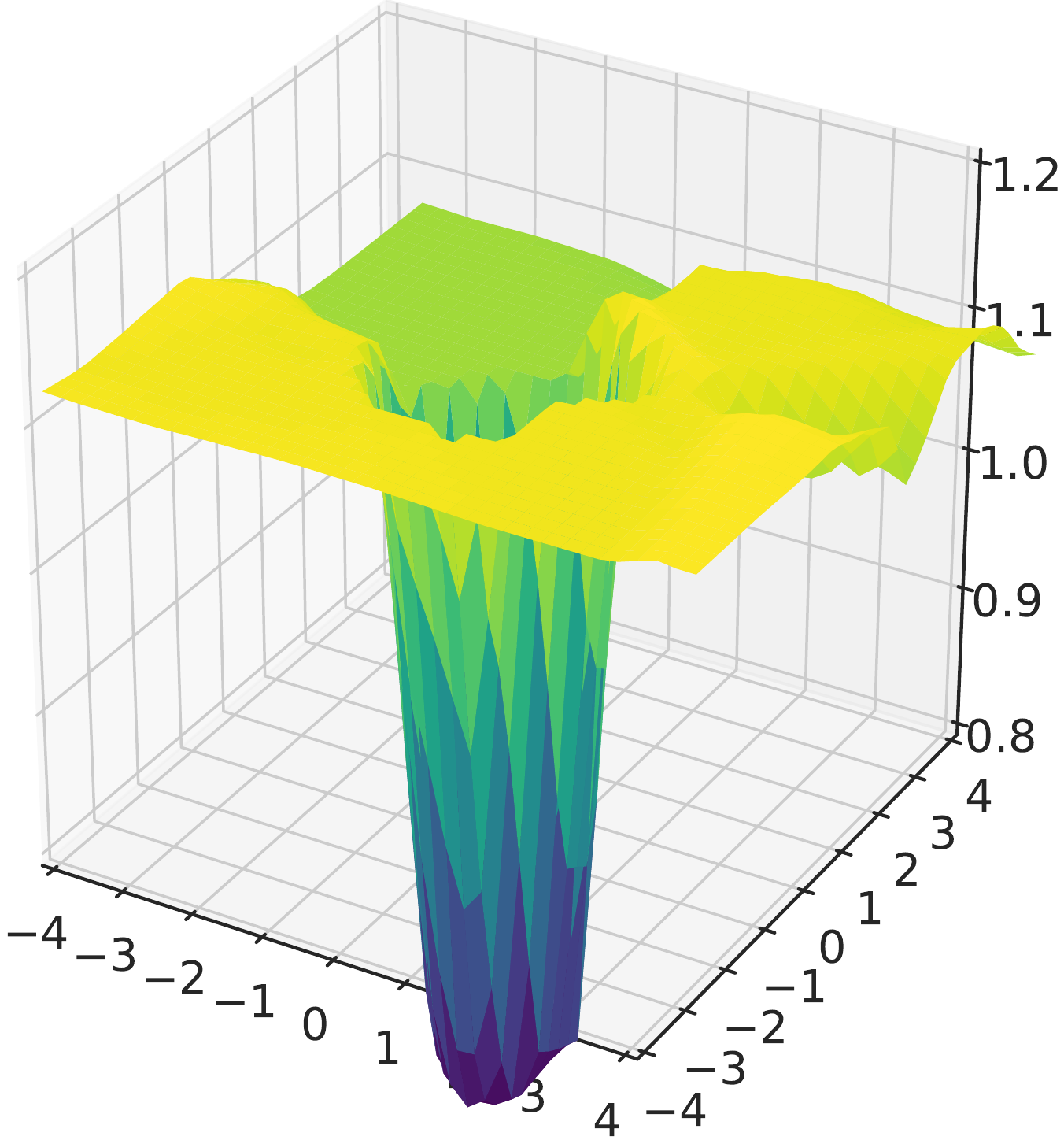}
\includegraphics[width=0.8in,height=0.8in]{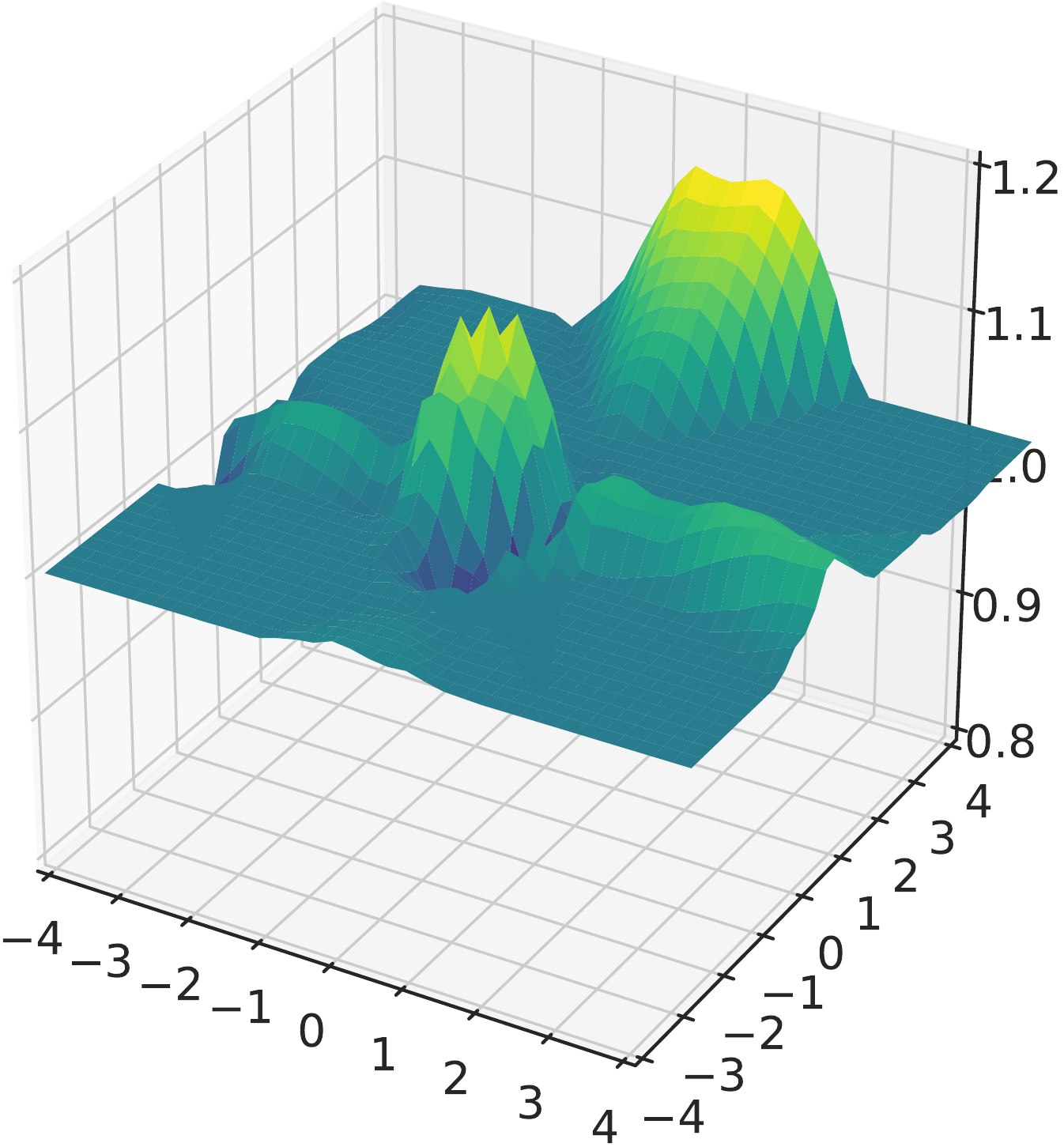}
\includegraphics[width=0.8in,height=0.8in]{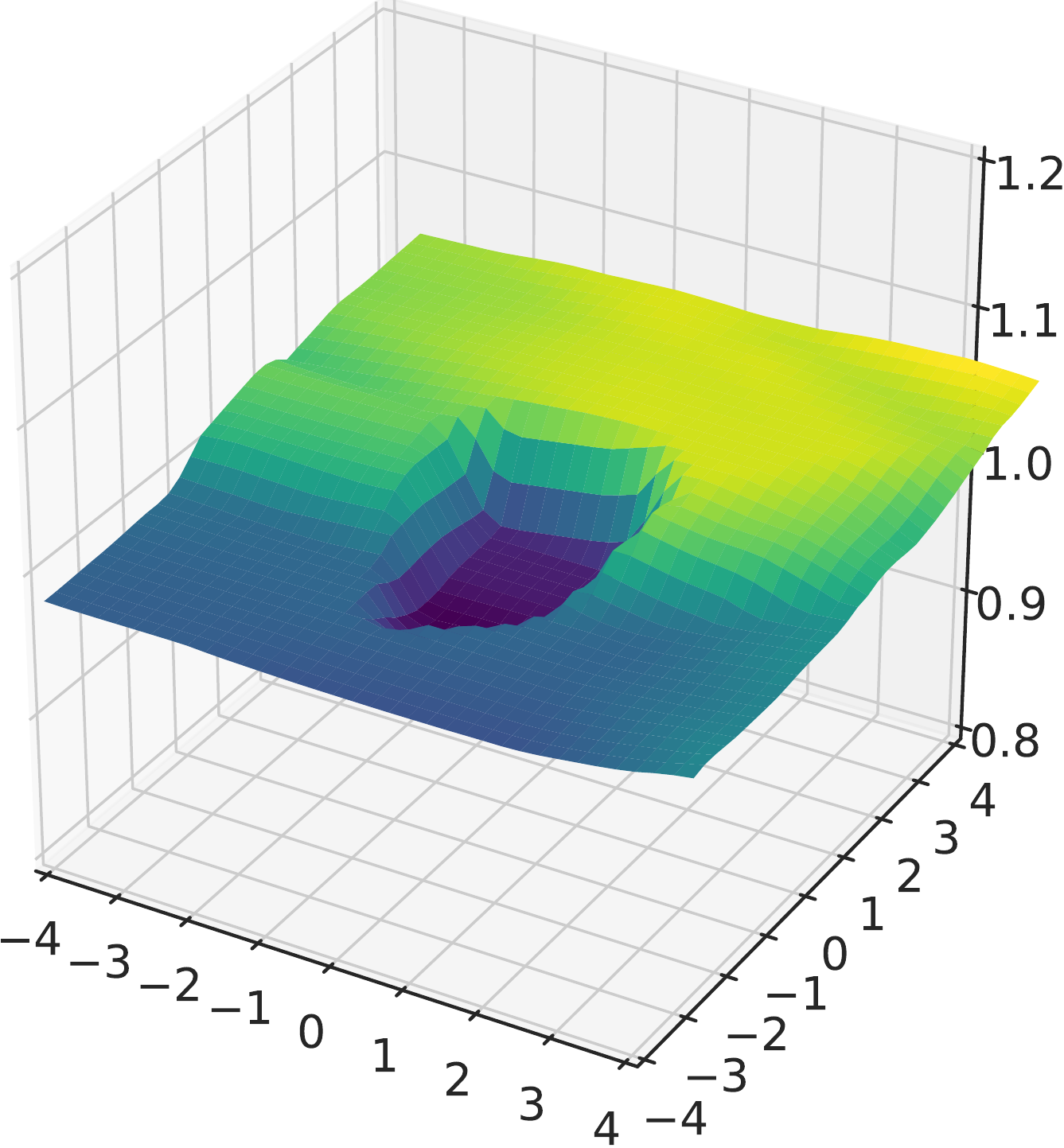}
\includegraphics[width=0.8in,height=0.8in]{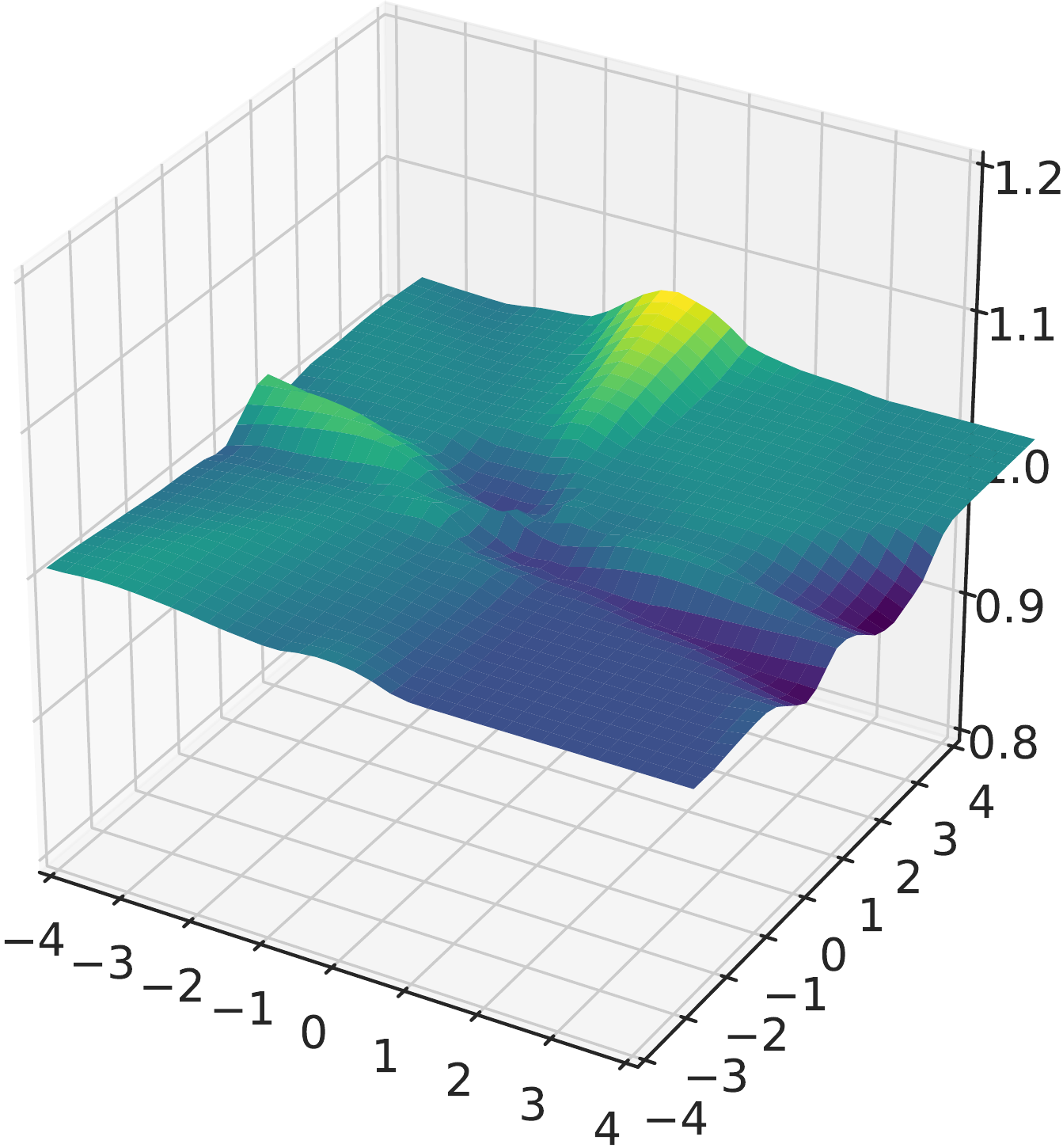}
\end{minipage}
\label{map-dr}
}
\centering
\caption{Learned  transport maps and estimated density-ratio in  learning  $5squares$ from $4squares$,   and learning   $large4gaussians$ from $small4gaussians$.}
\label{map}
\end{figure}

Next, we demonstrate the effectiveness of using  the gradient penalty (\ref{gp}) by
visualizing  the transport maps learned in the generative learning  tasks  with the learning targets  $5squares$ and $large4gaussians$   from $4squares$ and $small4gaussians$, respectively.
We use  200 particles   connected with grey lines to manifest the learned transport maps.
As shown in  Figure \ref{map-map}, the central squares of $5squares$ were learned better with the gradient penalty, which is consistent with the result on the estimated density-ratio  in Figure \ref{map-dr}.
For $large4gaussians$, the learned transport map exhibited some  optimality under quadratic Wasserstein distance due to the obvious correspondence between the samples in Figure \ref{map-map}, and the   gradient penalty also improves  the density-ratio estimation as expected.



\subsection{Benchmark Image Data}
\begin{figure}[ht!]
\centering{}

\begin{tabular}{ll}
\includegraphics[width=2.4in, height=1.2in]{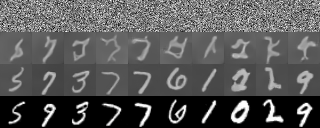}
&
\includegraphics[width=2.4in, height=1.2in]{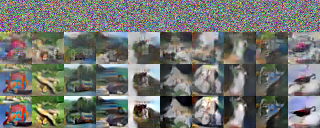}
\end{tabular}

\caption{Particle evolution of EPT on MNIST and  CIFAR10.}
\label{particle_evol}
\end{figure}

We show  the performance of applying  EPT to benchmark data  MNIST \citep{lecun98},  CIFAR10 \citep{krizhevsky09} and CelebA \citep{liu15} using  ReLU ResNets without batch normalization and spectral normalization.
The  particle evolutions  on  MNIST and CIFAR10 without using outer loop are shown in Figure \ref{particle_evol}.
Clearly, EPT  can  transport samples from a multivariate normal distribution into a target distribution.

\begin{figure}[ht!]
\centering{}
\begin{tabular}{ll}
\includegraphics[width=2.4in, height=0.8in]{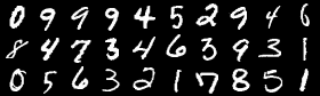}&
\includegraphics[width=2.4in, height=0.8in]{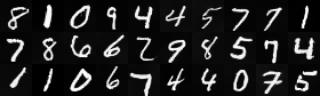}\\
\includegraphics[width=2.4in, height=0.8in]{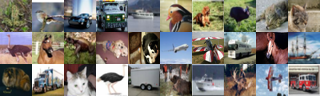} &
\includegraphics[width=2.4in, height=0.8in]{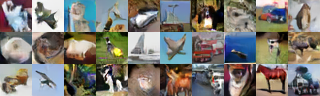}\\
\includegraphics[width=2.4in, height=0.8in]{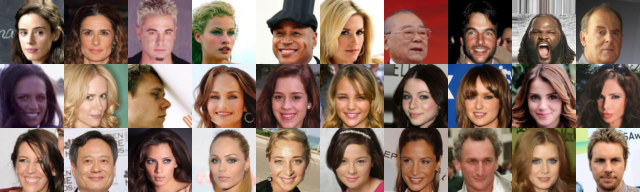} &
\includegraphics[width=2.4in, height=0.8in]{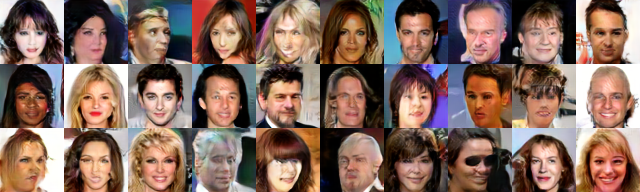}
\end{tabular}
\caption{Visual comparisons between real images (left 3 panels) and generated images (right 3 panels) by {EPT}-LSDR-$\chi^2$  on MNIST, CIFAR10 and CelebA.}
\label{sample_comp}
\end{figure}

We further compare {EPT}  using the outer loop  with  
the generative models including  WGAN, SNGAN and MMDGAN.
We considered different $f$-divergences, including Pearson's $\chi^2$, KL, JS and logD  \citep{gao2019deep} and different deep density-ratio fitting methods (LSDR and LR).
Table \ref{fid-cf10-50k} shows FID  \citep{heusel17} evaluated with  five bootstrap sampling of  {EPT} with four divergences  on CIFAR10.
We can see that {EPT} using ReLU ResNets without batch normalization and spectral normalization attains (usually better) comparable FID  scores with the state-of-the-art generative models.
Comparisons of the real samples and learned samples on MNIST,  CIFAR10 and  CelebA are shown in Figure \ref{sample_comp}, where high-fidelity learned samples  are comparable to real samples  visually.

\begin{table}[ht!]
\caption{{\color{black}Mean (standard deviation) of FID scores on  CIFAR10. The FID score of NSCN is reported in \cite{yang2019} and results in the right table are adapted from \cite{arbel18}}.
 }
\label{fid-cf10-50k}
\vskip 0.15in
\begin{center}
\begin{tabular}{cc}
\begin{small}
\begin{rm}
\begin{tabular}{llccr}
\toprule
Models 			& CIFAR10 (50k) \\
\midrule
{EPT}-LSDR-$\chi^2$ &  \textbf{24.9 (0.1)} \\
{EPT}-LR-KL 		&  25.9 (0.1) \\
{EPT}-LR-JS 		&  25.3 (0.1) \\
{EPT}-LR-logD 		&  \textbf{24.6 (0.1)} \\
NCSN			&  25.3 \\
\midrule
\end{tabular}
\end{rm}
\end{small}
&
\begin{small}
\begin{rm}
\begin{tabular}{lcccr}
\toprule
Models 			& CIFAR10 (50k) \\
\midrule
WGAN-GP			&  31.1 (0.2) \\
MMDGAN-GP-L2		&  31.4 (0.3) \\
SN-GAN		  		&  26.7 (0.2) \\
SN-SMMDGAN			&  \textbf{25.0 (0.3)} \\
\bottomrule
\end{tabular}
\end{rm}
\end{small}
\end{tabular}
\end{center}
\end{table}

\subsection{Numerical convergence}
\begin{figure}[ht!]
\centering
\includegraphics[width=\linewidth, height=2.5in]
{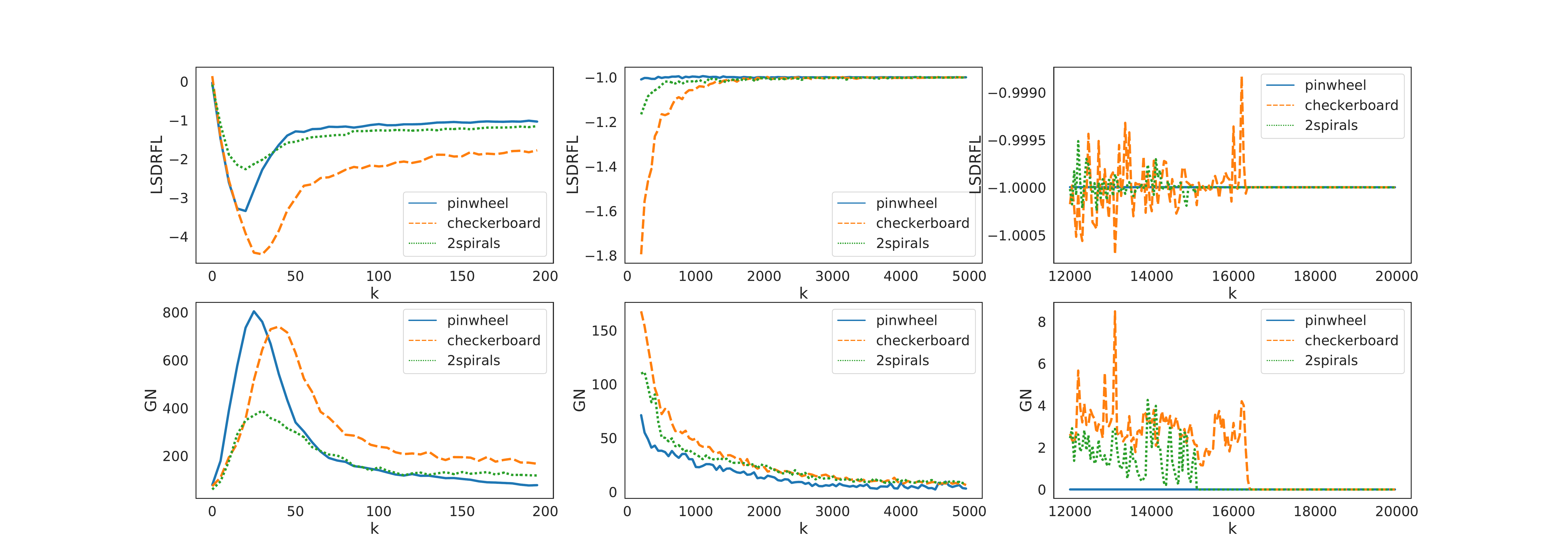}
\caption{The numerical convergence phenomenon of EPT on simulated datasets. First row: LSDR fitting loss  (\ref{sf}) with $\alpha = 0$ v.s. iterations  on \emph{pinwheel, checkerboard} and \emph{2spirals}.
Second row: Estimation of the gradient norm $\mathbb{E}_{X \sim q_k} [\Vert \nabla R_{\phi}(X) \Vert_2]$ v.s. iterations on \emph{pinwheel, checkerboard} and \emph{2spirals}.}
\label{loss_2d}
\end{figure}

Finally, we illustrate the convergence property of  the learning dynamics of  {EPT} on synthetic datasets \emph{pinwheel, checkerboard} and \emph{2spirals}. As shown in Figure \ref{loss_2d}, on the three test datasets, the dynamics of both the estimated LSDR fitting losses  in (\ref{sf}) with $\alpha = 0$ and the
estimated value of the gradient norms $\mathbb{E}_{X \sim q_k} [\Vert \nabla R_{\phi}(X) \Vert_2]$
 demonstrate the estimated LSDR loss converges to the theoretical value $-1$.

\section{Conclusion and future work} \label{con}
EPT is a new approach for generative learning via training a transport map that pushes forward a reference to the target.
This approach uses the forward Euler method for solving the McKean-Vlasov equation, which results from linearizing the Monge-Amp\`{e}re equation that characterizes the
optimal transport map.
The EPT map is a composition of a sequence of simple residual maps.
The key task in training is the estimation of density ratios that completely determine the residual maps.
We estimate density ratios based on the Bregman divergence with gradient penalty using deep density-ratio fitting. We establish bounds on the approximation errors due to linearization, discretization, and density-ratio estimation.
Our results provide strong theoretical guarantees for the proposed method and ensure that the EPT map converges fast to the target.
We also show that the proposed density-ratio (difference) estimators do not suffer from the ``curse of dimensionality'' if data is supported on a lower-dimensional manifold. This is an interesting result in itself since density-ratio estimation is an important basic problem in machine learning and statistics. Because EPT is easy to train, computationally stable, and enjoys strong theoretical guarantees, we expect it to be a useful addition to the methods for generating learning.

There are two important ingredients in EPT:
the velocity field and density-ratio  estimation.
With a suitable choice of the velocity and a density-ratio estimation procedure, EPT can recover several existing generative models such as MMD flow and SVGD. Thus our theoretical results also provide insights into the properties of these methods.
Simulation results on multi-mode synthetic datasets and comparisons  with the existing methods on real benchmark datasets  using simple ReLU ResNets without batch normalization and spectral normalization support our theoretical analysis and demonstrate the effectiveness 

Some aspects and results in this paper are of independent
interest. For example, density-ratio estimation is an important
problem and of general interest in machine learning
and statistics. The estimation error bound established in
Theorem \ref{th3} for the nonparametric deep density-ratio fitting procedure are new. This provides an important example showing that deep nonparametric estimation can circumvent the
curse of dimensionality via exploring the data structure. It is of interest to use the techniques developed here to investigate the generalization errors in setting
of deep nonparametric regressions and classifications.

The proposed EPT method is motivated by the Monge-Amp\`{e}re equation that characterizes the
optimal transport map. However, while  EPT pushes forward a reference distribution to the target, it is not an estimate of the optimal transport map itself. How to consistently estimate the Monge-Amp\'{e}re optimal map remains a challenging and open problem.


\newpage

\bibliographystyle{iclr2021_conference}
\bibliography{EPT_arXiv}

\newpage
\setcounter{equation}{0}  
\renewcommand{\theequation}{B-\arabic{equation}}

\setcounter{table}{0}
\renewcommand{\thetable}{A\arabic{table}}
\appendix
{\center
\textbf{APPENDIX}
}

In the appendix,
we provide the implementation details on numerical settings, network structures, SGD optimizers, and hyper-parameters in the paper.
{\color{black}
We show the numerical convergence of EPT with simulated datasets and compare the learning and inference of EPT with other generative models.
}
We give detailed theoretical background and proofs of the results mentioned in the paper. We also provide proofs MMD flow and SVGD can be derived from EPT by choosing appropriate $f$-divergences.
\section{Appendix: Numerical experiments}

\subsection{Algorithm details}
We provide the details of two versions of the EPT algorithm, EPTv1  in Algorithm \ref{ept} and EPTv2 in
Algorithm \ref{ept_latent} below.
In Algorithm \ref{ept}, we describe the algorithm without outer loops. In Algorithm \ref{ept_latent}, we describe the algorithm with a latent structure and outer loops.

\newcommand\mycommfont[1]{\footnotesize\ttfamily\textcolor{pinegreen}{#1}}
\SetCommentSty{mycommfont}
\IncMargin{1em}
\begin{algorithm}[t]
	\DontPrintSemicolon
	\SetNoFillComment
	\caption{EPTv1: Euler particle transport} \label{ept}
	{\textbf{Input}}:
	$K \in \mathbb{N}^{*}$, $s > 0$, $\alpha > 0$		\tcp*{maximum loop count, step size, regularization coeficient}
	$X_i \sim \nu, \tilde{Y}^{0}_i \sim \mu$, $i =1, 2, \cdots, n$	\tcp*{real samples, initial particles}		
	$k \gets 0$
	
	\While {$k < K$ }
	{
		 $\widehat{R}^k_{{\phi}} \in \arg \min_{R_\phi}  \frac{1}{n} \sum_{i=1}^n [R_{\phi}(X_i)^2 +\alpha \|\nabla R_{\phi}(X_i)\|^2_{2} -2 R_{\phi}(\tilde{Y}^k_i)]$ via SGD \tcp*{determine the density ratio}

		$\hat{\vv}^k (\vx) = -f^{\prime\prime}(\widehat{R}^k_{{\phi}}(\vx))\nabla \widehat{R}^k_{{\phi}}(\vx)$	\tcp*{approximate the velocity field}
		
		$\widehat{\mathcal{T}}^k = \Id + s \hat{\vv}^k$	    \tcp*{define the forward Euler map}
		
		$\tilde{Y}^{k+1}_i = \widehat{\mathcal{T}}^{k} (\tilde{Y}^{k}_i)$, $i =1, 2, \cdots, n$	\tcp*{update particles}
		$ k \gets k + 1$
	}
	{\textbf{Output}}:  $\tilde{Y}^{k}_i \sim \tilde{\mu}_k$, $i =1, 2, \cdots, n$	\tcp*{transported particles}
\end{algorithm}

\SetCommentSty{mycommfont}
\IncMargin{1em}
\begin{algorithm}[t]
	\DontPrintSemicolon
	\SetNoFillComment
	\caption{EPTv2: Euler particle transport with latent structure} \label{ept_latent}
	{\textbf{Input}}:
	$IL, OL\in \mathbb{N}^{*}$, $s > 0$, $\alpha > 0$		\tcp*{maximum inner loop count, maximum outer loop count, step size, regularization coeficient}
	
	$X_i \sim \nu$, $i =1, 2, \cdots, n$	\tcp*{real samples}
	
	$\widehat{G}^0_{\theta} \gets G^{init}_{\theta} $		\tcp*{initialize the transport map}
	
	$j \gets 0$	
	
	\tcc{outer loop}
	
	\While {$j < OL$}
	{
		$Z^{j}_i \sim \tilde{\mu}$, $i =1, 2, \cdots, n$		\tcp*{latent particles}	
		
		$\tilde{Y}^{0}_i = \widehat{G}^j_{\theta}(Z^{j}_i)$, $i =1, 2, \cdots, n$		\tcp*{intermediate particles}	
		
		$k \gets 0$
		
		\tcc{inner loop}
		
		\While {$k < IL$ }
		{
			$\widehat{R}^k_{{\phi}} \in \arg \min_{R_\phi}  \frac{1}{n} \sum_{i=1}^n [R_{\phi}(X_i)^2 +\alpha \|\nabla R_{\phi}(X_i)\|^2_{2} -2 R_{\phi}(\tilde{Y}^k_i)]$ via SGD \tcp*{determine the density ratio}

			$\hat{\vv}^k (\vx) = -f^{\prime\prime}(\widehat{R}^k_{{\phi}}(\vx))\nabla \widehat{R}^k_{{\phi}}(\vx)$	\tcp*{approximate the velocity field}
		
			$\widehat{\mathcal{T}}^k = \Id + s \hat{\vv}^k$	    \tcp*{define the forward Euler map}
		
			$\tilde{Y}^{k+1}_i = \widehat{\mathcal{T}}^{k} (\tilde{Y}^{k}_i)$, $i =1, 2, \cdots, n$	\tcp*{update particles}
			
			$ k \gets k + 1$

		}
		$\widehat{G}^{j+1}_{{\theta}} \in \arg \min_{G_{\theta}} \frac{1}{n} \sum_{i=1}^n \|G_{\theta}(Z^{j}_i) - \tilde{Y}^{IL}_i\|_{2}^2$ via SGD	\tcp*{fit the transport map}
		
		$j \gets j + 1$
	}
	{\textbf{Output}}:  $\widehat{G}^{OL}_{{\theta}}: \mathbb{R}^{\ell} \rightarrow \mathbb{R}^{d}$	\tcp*{transport map with latent structure}
\end{algorithm}

\subsection{Implementation details, network structures, hyper-parameters}
We provide the details of two versions of the EPT algorithm, EPTv1  in Algorithm \ref{ept} and EPTv2 in
Algorithm \ref{ept_latent} below.
In Algorithm \ref{ept}, we describe the algorithm without outer loops. In Algorithm \ref{ept_latent}, we describe the algorithm with a latent structure and outer loops.

\subsubsection{2D examples}

Experiments on 2D examples in our work were performed with deep LSDR fitting and the Pearson $\chi^2$ divergence.
We use the EPTv1 (Algorithm \ref{ept}) without outer loops.
In inner loops, only a multilayer perceptron (MLP) was utilized for dynamic estimation of the density ratio between the model distribution $q_k$ and the target distribution $p$.
The network structure and hyper-parameters in EPT and deep LSDR fitting were shared in all 2D experiments.
We adopt  EPT to push particles from a predrawn pool consisting of 50k i.i.d. Gaussian particles to evolve in 20k steps.
We used RMSProp with the learning rate 0.0005 and the batch size 1k as the SGD optimizer.
The details are given in Table \ref{mlp} and Table \ref{param_2D}. We note that $s$ is the step size, $n$ is the number of particles, $\alpha$ is the penalty coefficient, and $T$ is the mini-batch gradient descent times of deep LSDR fitting or deep logistic regression in each inner loop hereinafter.
\begin{table}[ht]
\caption{MLP for deep LSDR fitting.}
\label{mlp}
\vskip 0.15in
\begin{center}
\begin{small}
\begin{rm}
\begin{tabular}{lcccr}
\toprule
Layer	& Details 			& Output size \\
\midrule
1		& Linear, ReLU 	& 64 \\
\midrule
2		& Linear, ReLU 	& 64 \\
\midrule
3		& Linear, ReLU 	& 64 \\
\midrule
4		& Linear			& 1 \\
\bottomrule
\end{tabular}
\end{rm}
\end{small}
\end{center}
\vskip -0.1in
\end{table}

\begin{table}[h]
\caption{Hyper-parameters in EPT on 2D examples.}
\label{param_2D}
\vskip 0.15in
\begin{center}
\begin{small}
\begin{rm}
\begin{tabular}{lcccr}
\toprule
Parameter		&  $s$ 	& $n$ 	& $\alpha$ 		& $T$ \\
\midrule
Value		& 0.005 	& 50k 	& 0 ${\rm or}$ 0.5 	& 5 \\
\bottomrule
\end{tabular}
\end{rm}
\end{small}
\end{center}
\vskip -0.1in
\end{table}

\subsubsection{Real image data}

\textbf{Datasets.}
We evaluated EPT on three benchmark datasets including two small datasets MNIST, CIFAR10 and one large dataset CelebA from GAN literature. MNIST contains a training set of 60k examples and a test set of 10k examples as $28\times28$ bilevel images which were resized to $32\times32$ resolution. There are a training set of 50k examples and a test set of 10k examples as $32\times32$ color images in CIFAR10. We randomly divided the 200k celebrity images in CelebA into two sets for training and test according to the ratio 9:1. We also pre-processed CelebA images by first taking a $160\times160$ central crop and then resizing to the $64\times64$ resolution. Only the training sets are used to train our models.

\textbf{Evaluation metrics.}
\emph{Fr\'echet Inception Distance} (FID) \citep{heusel17} computes the Wasserstein distance $\mathcal{W}_2$ with summary statistics (mean $\mu$ and variance $\Sigma$) of real samples $\mathbf{x}s$ and generated samples $\mathbf{g}s$ in the feature space of the Inception-v3 model \citep{szegedy16}, i.e., ${\rm FID} = \Vert \mu_{\mathbf{x}} - \mu_{\mathbf{g}} \Vert^2_2 + {\rm Tr}(\Sigma_{\mathbf{x}} + \Sigma_{\mathbf{g}} - 2(\Sigma_{\mathbf{x}} \Sigma_{\mathbf{g}})^{\frac12})$. Here, FID is reported with the TensorFlow implementation and lower FID is better.

\textbf{Network architectures and hyper-parameter settings.}
We employed the ResNet architectures used by \cite{gao2019deep} in our EPT algorithm.
Especially, the batch normalization \citep{ioffe2015batch} and the spectral normalization \citep{miyato18} of networks were omitted for EPT-LSDR-$\chi^2$.
To train neural networks, we set SGD optimizers as RMSProp with the learning rate 0.0001 and the batch size 100.
Inputs $\{ Z_i \}_{i=1}^n$ in EPTv2 (Algorithm \ref{ept_latent}) were vectors generated from a 128-dimensional standard normal distribution on all three datasets. Hyper-parameters are listed in Table \ref{param_ol} where $IL$ expresses the number of inner loops in each outer loop. Even without outer loops, EPTv1 (Algorithm \ref{ept}) can generate images on MNIST and CIFAR10 as well by making use of a large set of particles. Table \ref{param_no_ol} shows the hyper-parameters.

\begin{table}[ht!]
\caption{Hyper-parameters in EPT \textbf{with} outer loops on real image datasets.}
\label{param_ol}
\vskip 0.15in
\begin{center}
\begin{small}
\begin{rm}
\begin{tabular}{lcccccr}
\toprule
Parameter		& $\ell$ 	& $s$ 	& $n$ 	& $\alpha$ 	& $T$ 	& $IL$ \\
\midrule
Value 		& 128  	& 0.5  	& 1k 		& 0 			& 1 		& 20 \\
\bottomrule
\end{tabular}
\end{rm}
\end{small}
\end{center}
\vskip -0.1in
\end{table}

\begin{table}[ht!]
\caption{Hyper-parameters in EPT \textbf{without} outer loops on real image datasets.}
\label{param_no_ol}
\vskip 0.15in
\begin{center}
\begin{small}
\begin{rm}
\begin{tabular}{lcccr}
\toprule
Parameter		&  $s$  	& $n$ 	& $\alpha$ 	& $T$ \\
\midrule
Value		& 0.5  	& 4k 		& 0 			& 5 \\
\bottomrule
\end{tabular}
\end{rm}
\end{small}
\end{center}
\vskip -0.1in
\end{table}

\subsection{Learning and inference}
The learning process of  EPT performs particle evolution via solving the McKean-Vlasov equation using forward Euler iterations. The iterations rely on the estimation of the density ratios (difference) between the pushforward distributions and the target distribution.
To make the inference of EPTv1 more amendable, we propose EPTv2 based on EPTv1. EPTv2 takes advantage of a neural network to fit the pushforward map.
The inference of EPTv2 is fast since the pushforward map is parameterized as a neural network and only forward propagation is involved. These aspects distinguish EPTv2 from score-based generative models \cite{yang2019, yang2020} which simulate Langevin dynamics to generate samples.

\section{Appendix: Proofs}
\label{AppTheory}

\subsection{Proof for Section \ref{theory}}
\noindent
\textbf{Proof of Proposition \ref{th1}.}
(i) The continuity equation (\ref{vfp}) follows from the definition  of the gradient flow directly, see, page 281 in \citep{ambrosio2008gradient}.
(ii) The first equality  follows from the chain rule and integration by part, see, Theorem 24.2 of \cite{villani2008optimal}.
The second one on linear convergence  follows from Theorem 24.7 of \cite{villani2008optimal},  where the assumption on $\lambda$  in equation (24.6) is equivalent to the $\lambda$-geodetically convex assumption here.
(iii) Similar to (i) see, page 281 in \cite{ambrosio2008gradient}.
$\hfill$ $\Box$

\medskip\noindent
\textbf{Proof of Theorem \ref{th1b}.}
(i) Recall  $\mathcal{L}[\mu]$ is a functional  on  $\mathcal{P}_2^{a}(\mathbb{R}^{m})$. By the classical results in calculus  of variation \citep{gelfand2000calculus}, $$\frac{\partial \mathcal{L}[q]}{\partial q}(\vx) = \frac{\mathrm{d}}{\mathrm{d}t} \mathcal{L}[q+tg]\mid_{t = 0} = F^{\prime}(q(\vx)),$$
where $\frac{\partial \mathcal{L}[q]}{\partial q} $ denotes the first order of variation of $\mathcal{L}[\cdot]$ at $q$, and $q, g $ are the densities  of $\mu$ and an arbitrary  $ \xi \in \mathcal{P}_2^{a}(\mathbb{R}^{m})$, respectively. Let
$$L_{F}(z) = z F^{\prime}(z) -F(z): \mathbb{R}^{1} \rightarrow \mathbb{R}^{1}.$$
Some algebra shows, $$\nabla L_{F}(q(\vx)) = q(\vx) \nabla F^{\prime} (q(\vx)).$$ Then, it follows from Theorem 10.4.6 in \citep{ambrosio2008gradient} that $$ \nabla F^{\prime} (q(\vx)) = \partial^{o}{L}(\mu),$$ where, $\partial^{o}{L}(\mu)$ denotes the one in $\partial{L}(\mu)$ with minimum length.  The above display  and  the definition of  gradient flow  implies  the representation of the velocity fields $\vv_t$.\\

(ii)
The time dependent form of (\ref{leq1})-(\ref{leq2}) reads
\begin{align*}
\frac{\mathrm{d} \vx_t}{\mathrm{d} t} &= \nabla \Phi_t(\vx_t), \ \  \mathrm{with} \ \  \vx_0  \sim q,\\
\frac{\mathrm{d} \ln  q_t(\vx_t)}{\mathrm{d} t} &= - \Delta
\Phi_t (\vx_t), \ \  \mathrm{with} \ \  q_0 = q.
\end{align*}
By chain rule and substituting the first equation into the second one, we have
 \begin{align*}
 \frac{1}{q_t}(\frac{\mathrm{d}q_t}{\mathrm{d}t}+\frac{\mathrm{d}q_t}{\mathrm{d}\vx_t}\frac{\mathrm{d}\vx_t}{\mathrm{d}t}) &= \frac{1}{q_t}(\frac{\mathrm{d}q_t}{\mathrm{d}t}+\nabla q_t\nabla \Phi_t(\vx_t))\\
 &=- \Delta\Phi_t (\vx_t),
\end{align*}
which implies,
$$\frac{\mathrm{d}q_t}{\mathrm{d}t} = - q_t\Delta\Phi_t (\vx_t) -\nabla q_t\nabla \Phi_t(\vx_t) = -\nabla\cdot(q_t \nabla \Phi_t).$$
By (\ref{vr}), the above display coincides with the continuity equation  (\ref{vfp}) with $\vv_t = \nabla \Phi_t = -\nabla F^{\prime} (q_t(\vx))$.
$\hfill$ $\Box$
\textbf{Proof of Theorem \ref{th2}.}
The Lipschitz assumption of $\vv_t$ implies the existence and uniqueness of the   McKean-Vlasov equation (\ref{mve}) according to the classical results in ODE \citep{arnold2012geometrical}.
By the uniqueness of the continuity equation, see Proposition 8.1.7 in \cite{ambrosio2008gradient},  it is sufficient to show that $\mu_t = (\bX_t)_{\#}\mu$
satisfies the continuity equation  (\ref{vfp}) in a weak sense. This can be done by the standard test function and smoothing approximation arguments, see, Theorem 4.4 in \cite{santambrogio2015optimal} for details.
$\hfill$ $\Box$

\textbf{Proof of Lemma \ref{lem2}.}
By  definition,
\begin{equation*}
F(q_t(\vx))  =
\left\{\begin{array}{ll}
p(\vx) f(\frac{q_t(\vx)}{p(\vx)}), \ \  \mathcal{L}[\mu] = \mathbb{D}_f(\mu \Vert \nu), \\
(q_t(\vx)-p(\vx))^2,  \ \ \mathcal{L}[\mu] = \|\mu-\nu\|^2_{L^2(\mathbb{R}^{m})}.
\end{array}
\right.
\end{equation*}
Direct calculation shows
\begin{equation*}
F^{\prime}(q_t(\vx))  =
\left\{\begin{array}{ll}
f^{\prime}(\frac{q_t(\vx)}{p(\vx)}), \ \  \mathcal{L}[\mu] = \mathbb{D}_f(\mu \Vert \nu), \\
2(q_t(\vx)-p(\vx)),  \ \ \mathcal{L}[\mu] = \|\mu-\nu\|^2_{L^2(\mathbb{R}^{m})}.
\end{array}
\right.
\end{equation*}
Then, the desired  result follows from the above display and  (\ref{vr}).
$\hfill$ $\Box$

{\color{black}
\medskip\noindent
\textbf{Proof of Proposition \ref{prop2}.}
Without loss of generality let  $K = \frac{T}{s} >1$ be an integer.
Recall $\{\mu_t^{s} \ \  t\in [ks,(k+1)s)$ is  the piecewise constant  interpolation between $\mu_{k}$ and  $\mu_{k+1}$  defined as
$$\mu_t^{s}  = (\mathcal{T}_t^{k,s})_{\#}\mu_k,$$
where, $$\mathcal{T}_t^{k,s}= \Id + (t-ks) \vv_{k},$$ $\mu_k$ is defined in (16)-(18) with  $\vv_{k}= \vv_{ks}$, i.e., the continuous velocity in (\ref{vr})  at time $ks$,   $k = 0,.., K-1$, $\mu_0 = \mu.$
Under assumption (\ref{Lip})
we can first show in a way similar to the proof of Lemma 10 in \cite{arbel2019maximum} that
  \begin{equation}
\label{(A1)}
\mathcal{W}_2(\mu_{ks}, \mu_k) = \mathcal{O}(s).
\end{equation}
  Let $\Gamma$ be the optimal coupling between $\mu_k$ and $\mu_{ks}$, and $(X,Y) \sim \Gamma$.
  Let $X_t = \mathcal{T}_t^{k,s}(X)$ and  $Y_t$ be the solution of  (\ref{mve})  with $\mathbf{X}_0 = Y$ and $t\in [ks,(k+1)s)$.
   Then $$X_t \sim \mu_t^s, \ \  Y_t \sim \mu_t$$ and  $$Y_t = Y + \int_{ks}^{t} \vv_{\tilde{t}}(Y_{\tilde{t}}) \mathrm{d} \tilde{t}.$$
It follows that
\begin{eqnarray}
   \mathcal{W}_2^{2}(\mu_t, \mu_{ks}) &\leq& \mathbb{E} [\|Y_t -Y\|_2^2] \\ \nonumber
   &= &\mathbb{E} [\|\int_{ks}^{t} \vv_{\tilde{t}}(Y_{\tilde{t}}) \mathrm{d} \tilde{t}\|_2^2] \\ \nonumber
   &\leq & \mathbb{E}[(\int_{ks}^{t} \|\vv_{\tilde{t}}(Y_{\tilde{t}})\|_2\mathrm{d} {\tilde{t}})^2]  \\ \nonumber
   &\leq &  \mathcal{O}(s^2).   \label{(A2)}
\end{eqnarray}
   where, the first inequality follows from the definition of $\mathcal{W}_2$, and the last equality follows from the  the uniform bounded assumption of $\vv_t$.
   Similarly,
   \begin{eqnarray}
   \mathcal{W}_2^{2}(\mu_k, \mu_{t}^{s})
   &\leq & \mathbb{E} [\|X -X_t\|_2^2] \nonumber \\
   &=& \mathbb{E} [\|(t-ks)\vv_{k}(X)\|_2^2] \nonumber \\
   &\leq &  \mathcal{O}(s^2).   \label{(A3)}
   \end{eqnarray}
   Then,
 \begin{eqnarray*}
  \mathcal{W}_2(\mu_t, \mu_t^s)
  & \leq & \mathcal{W}_2(\mu_t, \mu_{ks}) + \mathcal{W}_2(\mu_{ks}, \mu_k)+ \mathcal{W}_2(\mu_{k}, \mu_t^s)\\
  &\leq& \mathcal{O}(s),
  \end{eqnarray*}
where the first inequality follows from the triangle inequality, see for example Lemma 5.3 in  \cite{santambrogio2015optimal}, and the second  one follows from (\ref{(A1)})-(\ref{(A3)}).
$\hfill$ $\Box$
}

\subsection{Derivation and Proofs of the results in Section \ref{dr}.}
\label{drapp}
\subsubsection{Bregman score for Density ratio/Difference}\label{dde}

The separable Bregman score with the base probability measure $p$ to measure the discrepancy between a measurable function
$R: \mathbb{R}^{m}\rightarrow \mathbb{R}^1$ and
the density ratio $r$ is
\begin{eqnarray*}
\mathfrak{B}_{\rm ratio}(r, R) & =& \mathbb{E}_{X \sim p} [ g^{\prime}(R(X)) (R(X) - r(X)) - g(R(X)) ] \\
&=& \mathbb{E}_{X \sim p} [ g^{\prime}(R(X)) R(X) - g(R(X)) ]  - \mathbb{E}_{X\sim q} [g^{\prime}(R(X))].
\end{eqnarray*}
It can be verified that  $\mathfrak{B}_{\rm ratio}(r, R) \ge \mathfrak{B}_{\rm ratio}(r, r)$, where the equality holds iff $R = r$.

For deep density-difference fitting, a neural network $D: \mathbb{R}^{m}\rightarrow \mathbb{R}^1$
is utilized to estimate the density-difference
$d(\vx)= q(\vx) - p(\vx)$ between a given density $q$ and the target $p$. The separable Bregman score with the base probability measure $w$ to measure the discrepancy between $D$ and $d$ can be derived  similarly,
\begin{align*}
 \mathfrak{B}_{\rm diff}(d, D)
&= \mathbb{E}_{X \sim p} [w(X) g^{\prime}(D(X))] - \mathbb{E}_{X \sim q} [w(X) g^{\prime}(D(X))] \\
 & + \mathbb{E}_{X\sim w} [ g^{\prime}(D(X)) D(X) - g(D(X)) ].
\end{align*}
Here, we focus on the widely used least-squares density-ratio (LSDR) fitting with $g(c) = (c-1)^2$ as a working example for estimating the density ratio $r$. The LSDR loss function is
\[
\mathfrak{B}_{\rm LSDR}(r, R)
= \mathbb{E}_{X \sim p} [ R(X) ^2 ]
- 2 \mathbb{E}_{X \sim q} [R(X)] + 1.
\]

\subsubsection{Gradient Penalty }\label{gpd}
We consider a noise convolution form of $\mathfrak{B}_{\rm ratio}(r, R)$ with Gaussian noise  $\bm{\epsilon} \sim \mathcal{N}(\mathbf{0}, \alpha  \mathbf{I})$,
\begin{align*}
&\mathfrak{B}_{\rm ratio}^{\alpha}(r, R) = \mathbb{E}_{X\sim p} \mathbb{E}_{\bm{\epsilon} } [ g^{\prime}(R(X + \bm{\epsilon})) R(X + \bm{\epsilon}) - g(R(X + \bm{\epsilon})) ]  - \mathbb{E}_{X \sim q} \mathbb{E}_{\bm{\epsilon} } [g^{\prime}(R(X + \bm{\epsilon}))].
\end{align*}
Taylor expansion applied to $R$ gives
\[
\mathbb{E}_{\bm{\epsilon}} [R(\vx + \bm{\epsilon})] = R(\vx) + \frac{\alpha}{2} \Delta{R(\vx)} + \mathcal{O}(\alpha^2).
\]
Using  equations (13)-(17) in \cite{roth2017stabilizing}, we get
\[\mathfrak{B}_{\rm ratio}^{\alpha}(r, R)
\approx \mathfrak{B}_{\rm ratio}(r, R) + \frac{\alpha}{2} \mathbb{E}_{p} [g''(R) \Vert \nabla R \Vert_2^2 ],
\]
i.e., $\frac12 \mathbb{E}_{p} [g^{\prime\prime}(R) \Vert \nabla R \Vert_2^2 ]$ serves as a regularizer for deep
density-ratio fitting when $g$ is twice differentiable.

\subsubsection{Proofs  in Section \ref{dr}}\label{prf}

Below we prove Lemma \ref{lem3} and Theorem \ref{th3} in Section \ref{dr}.

\medskip\noindent
\textbf{Proof Lemma \ref{lem3}.}
By definition, it is easy to check
$$\mathfrak{B}^{0}_{\rm LSDR}(R) = \mathfrak{B}_{\rm ratio}(r, R) - \mathfrak{B}_{\rm ratio}(r, r),$$
where $\mathfrak{B}_{\rm ratio}(r, R)$ is
the Bregman score with the base probability measure $p$ between $R$ and $r$.
Then $r \in \arg\min_{\text{measureable}\, R } \mathfrak{B}^{0}_{\rm LSDR}(R)$ follow from the fact
$\mathfrak{B}_{\rm ratio}(r, R) \ge \mathfrak{B}_{\rm ratio}(r, r)$ and  the equality holds iff $R = r$.
Since $$\mathfrak{B}^{\alpha}(R) = \mathfrak{B}^{0}_{\rm LSDR}(R) +  \alpha \mathbb{E}_{p} [\Vert \nabla R \Vert_2^2]\geq 0,$$
Then, $$\mathfrak{B}^{\alpha}(R) = 0$$ iff $$\mathfrak{B}^{0}_{\rm LSDR}(R) = 0 \ \ \mathrm{and}  \ \  \mathbb{E}_{p} [\Vert \nabla R \Vert_2^2] = 0,$$ which is further equivalent to
$$R = r = \mathrm{constant}  \ \ (q, p)\text{-}a.e. \, , $$  and the $\mathrm{constant}  = 1$  since  $r$ is a density ratio.
$\hfill$ $\Box$

\medskip\noindent
\textbf{Proof of Theorem \ref{th3}.}
We use  $\mathfrak{B}(R)$ to denote $\mathfrak{B}_{\rm LSDR}^{0}-C$ for simplicity, i.e.,
\begin{equation}
\label{A4}
\mathfrak{B}(R) = \mathbb{E}_{X \sim p} [ R(X) ^2 ]
- 2 \mathbb{E}_{X \sim q} [R(X)].
\end{equation}
Rewrite  (20) with $\alpha = 0$ as
\begin{equation}
\label{A5}
\widehat{R}_{\phi} \in \arg \min_{R_\phi\in \mathcal{H}_{\mathcal{D}, \mathcal{W}, \mathcal{S}, \mathcal{B}}} \widehat{\mathfrak{B}}(R_{\phi}) =  \sum_{i=1}^n   \frac{1}{n}(R_{\phi}(X_i)^2
  -2R_{\phi}(Y_i)).
\end{equation}
  By Lemma \ref{lem3} and Fermat's rule \citep{clarke1990optimization}, we know $ \mathbf{0} \in \partial  \mathfrak{B}(r).$
  Then,  $\forall R $      direct calculation yields,
  \begin{equation}
  \label{A6}
  \|R - r\|_{L^2(\nu)}^2  = \mathfrak{B}(R) - \mathfrak{B}(r) - \langle \partial  \mathfrak{B}(r),  R-r \rangle
  = \mathfrak{B}(R)- \mathfrak{B}(r).
  \end{equation}
   $ \forall \bar{R}_{\phi} \in \mathcal{H}_{\mathcal{D}, \mathcal{W}, \mathcal{S}, \mathcal{B}} $ we have,
  \begin{eqnarray}
\label{A7}
\|\widehat{R}_{\phi} - r\|_{L^2(\nu)}^2 & = & \mathfrak{B}(\widehat{R}_{\phi})- \mathfrak{B}(r) \\
 & = & \mathfrak{B}(\widehat{R}_{\phi}) - \widehat{\mathfrak{B}}(\widehat{R}_{\phi}) +    \widehat{\mathfrak{B}}(\widehat{R}_{\phi})-  \widehat{\mathfrak{B}}(\bar{R}_{\phi}) \nonumber  \\
  & + &  \widehat{\mathfrak{B}}(\bar{R}_{\phi}) -  \mathfrak{B}(\bar{R}_{\phi}) + \mathfrak{B}(\bar{R}_{\phi})  -  \mathfrak{B}(r) \nonumber \\
  &\leq & 2 \sup_{R \in \mathcal{H}_{\mathcal{D}, \mathcal{W}, \mathcal{S}, \mathcal{B}}} |\mathfrak{B}(R) - \widehat{\mathfrak{B}}(R) |+ \|\bar{R}_{\phi} - r\|_{L^2(\nu)}^2,\nonumber
  \end{eqnarray}
  where the  inequality uses the definition of  $\widehat{R}_{\phi}$, $\bar{R}_{\phi}$ and  (\ref{A6}).
  We prove the theorem by upper bounding the expected value of the right hand side term in (\ref{A7}). To this end, we need the following
  auxiliary results (\ref{A8})-(\ref{A10}).
 \begin{equation}
 \label{A8}
 \mathbb{E}_{\{Z_i\}_{i}^n} [\sup_{R } |\mathfrak{B}(R) - \widehat{\mathfrak{B}}(R) | ] \leq 4C_1(2\mathcal{B}+1) \mathfrak{G}(\mathcal{H}),
 \end{equation}
 where $$\mathfrak{G}(\mathcal{H}) =  \mathbb{E}_{\{Z_i, \epsilon_i \}_{i}^n}\left[\sup_{R\in \mathcal{H}_{\mathcal{D}, \mathcal{W}, \mathcal{S}, \mathcal{B}}}|\frac{1}{n}\sum_{i=1}^n\epsilon_i R(Z_i)|\right]$$ is the Gaussian complexity of $\mathcal{H}_{\mathcal{D}, \mathcal{W}, \mathcal{S}, \mathcal{B}}$ \citep{bartlett2002rademacher}.\\

  \textbf{Proof of (\ref{A8}).}
  Let $g(c) = c^2 - c$,  $\vz = (\vx,\vy) \in \mathbb{R}^m \times \mathbb{R}^m$,  $$\widetilde{R}(\vz) = (g\circ R)(\vz) = R^2(\vx) -R(\vy).$$
  Denote  $Z = (X,Y)$, $Z_i = (X_i,Y_i), i = 1,...,n$ with $X, X_i$ i.i.d.  $\sim p$, $Y, Y_i$ i.i.d. $\sim q$.
  Let $\widetilde{Z}_i$ be an i.i.d. copy of $Z_i,$ and $\sigma_i (\epsilon_i) $ be  i.i.d. Rademacher random (standard  normal) variables that are independent of
  $Z_i$ and $\widetilde{Z}_i$.
   Then,
   $$\mathfrak{B}(R) = \mathbb{E}_{Z} [\widetilde{R}(Z)] = \frac{1}{n}\mathbb{E}_{\widetilde{Z}_i} [\widetilde{R}(\widetilde{Z}_i)],$$ and
   $$ \widehat{\mathfrak{B}}(R) = \frac{1}{n}\sum_{i=1}^n \widetilde{R}(Z_i).$$
     Denote $$\mathfrak{R}(\mathcal{H}) = \frac{1}{n} \mathbb{E}_{\{Z_i, \sigma_i \}_{i}^n}[\sup_{R\in \mathcal{H}_{\mathcal{D}, \mathcal{W}, \mathcal{S}, \mathcal{B}}}|\sum_{i=1}^n\sigma_i R(Z_i)|]$$ as the 	
Rademacher complexity of $\mathcal{H}_{\mathcal{D}, \mathcal{W}, \mathcal{S}, \mathcal{B}}$ \citep{bartlett2002rademacher}.
    Then,
  \begin{eqnarray*}
  \mathbb{E}_{\{Z_i\}_{i}^n} [\sup_{R } |\mathfrak{B}(R) - \widehat{\mathfrak{B}}(R) | ]
  &= &\frac{1}{n} \mathbb{E}_{\{Z_i\}_{i}^n} [\sup_{R } |\sum_{i=1}^n (\mathbb{E}_{\widetilde{Z}_i} [\widetilde{R}(\widetilde{Z}_i)] - \widetilde{R}(Z_i))|]\\
  & \leq & \frac{1}{n} \mathbb{E}_{\{Z_i, \widetilde{Z}_i\}_{i}^n} [\sup_{R } |\widetilde{R}(\widetilde{Z}_i) - \widetilde{R}(Z_i)|]\\
  & = & \frac{1}{n} \mathbb{E}_{\{Z_i, \widetilde{Z}_i,\sigma_i \}_{i}^n} [\sup_{R } |\sum_{i=1}^n\sigma_i(\widetilde{R}(\widetilde{Z}_i) - \widetilde{R}(Z_i))|]\\
  & \leq & \frac{1}{n}  \mathbb{E}_{\{Z_i, \sigma_i \}_{i}^n} [\sup_{R } |\sum_{i=1}^n\sigma_i \widetilde{R}(Z_i)| ]
  + \frac{1}{n}  \mathbb{E}_{\{\widetilde{Z}_i, \sigma_i \}_{i}^n} [\sup_{R } |\sum_{i=1}^n\sigma_i \widetilde{R}(\widetilde{Z}_i)| ] \\
  &=& 2\mathfrak{R}(g\circ\mathcal{H})\\
  &\leq& 4(2\mathcal{B}+1)\mathfrak{R}(\mathcal{H})\\
  & \leq & 4C_1(2\mathcal{B}+1) \mathfrak{G}(\mathcal{H}),
  \end{eqnarray*}
where, the first inequality follows from the Jensen's inequality, and the second equality holds since the distribution of $\sigma_i(\widetilde{R}(\widetilde{Z}_i) - \widetilde{R}(Z_i))$ and $\widetilde{R}(\widetilde{Z}_i) - \widetilde{R}(Z_i)$ are the same, and the last equality holds since the distribution of the two terms are the same, and last two inequality follows from the  Lipschitz contraction property where the Lipschitz constant of $g$ on $\mathcal{H}_{\mathcal{D}, \mathcal{W}, \mathcal{S}, \mathcal{B}}$ is bounded by $2\mathcal{B}+1$ and the relationship between the Gaussian complexity and  the Rademacher complexity, see for Theorem 12 and Lemma 4 in \cite{bartlett2002rademacher}, respectively.
  \begin{equation}
  \label{(A9)}
  \mathfrak{G}(\mathcal{H}) \leq
  C_2\mathcal{B} \sqrt{\frac{n}{\mathcal{D}\mathcal{S}\log \mathcal{S}}}\log \frac{n}{\mathcal{D}\mathcal{S}\log \mathcal{S}} \exp(-\log^2 \frac{n}{\mathcal{D}\mathcal{S}\log \mathcal{S}}).
  \end{equation}

\noindent
  \textbf{Proof of (\ref{(A9)}).}
  Since $\mathcal{H}$ is negation closed,
\begin{eqnarray*}
\mathfrak{G}(\mathcal{H}) &=&  \mathbb{E}_{\{Z_i, \epsilon_i \}_{i}^n}[\sup_{R\in \mathcal{H}_{\mathcal{D}, \mathcal{W}, \mathcal{S}, \mathcal{B}}}\frac{1}{n}\sum_{i=1}^n\epsilon_i R(Z_i)]\\
& = & \mathbb{E}_{Z_i }[ \mathbb{E}_{\epsilon_i}[\sup_{R\in \mathcal{H}_{\mathcal{D}, \mathcal{W}, \mathcal{S}, \mathcal{B}}}\frac{1}{n}\sum_{i=1}^n\epsilon_i R(Z_i)]|\{Z_i\}_{i=1}^n].
\end{eqnarray*}
Conditioning on $\{Z_i\}_{i =1}^n$,
$\forall R, \widetilde{R} \in \mathcal{H}_{\mathcal{D}, \mathcal{W}, \mathcal{S}, \mathcal{B}}$ it easy to check $$\mathbb{V}_{\epsilon_i} [\frac{1}{n}\sum_{i=1}^n\epsilon_i (R(Z_i) - \widetilde{R}(Z_i))] = \frac{d^{\mathcal{H}}_{2}(R,\tilde{R})}{\sqrt{n}},$$
where, $d^{\mathcal{H}}_2(R,\tilde{R}) = \frac{1}{\sqrt{n}} \sqrt{\sum_{i =1}^n (R(Z_i)-\tilde{R}(Z_i))^2}$.
Observing the diameter of $\mathcal{H}_{\mathcal{D}, \mathcal{W}, \mathcal{S}, \mathcal{B}}$ under $d^{\mathcal{H}}_2 $ is at most $\mathcal{B}$,  we have
  \begin{align*}\mathfrak{G}(\mathcal{H}) &\leq \frac{C_3}{\sqrt{n}} \mathbb{E}_{\{Z_i\}_{i=1}^n}[\int_{0}^{B} \sqrt{\log \mathcal{N}(\mathcal{H}, d^{\mathcal{H}}_{2}, \delta)} \mathrm{d} \delta]\\
  &\leq\frac{C_3}{\sqrt{n}} \mathbb{E}_{\{Z_i\}_{i=1}^n}[\int_{0}^{\mathcal{B}} \sqrt{\log \mathcal{N}(\mathcal{H}, d^{\mathcal{H}}_{\infty}, \delta)} \mathrm{d} \delta]\\
  &\leq \frac{C_3}{\sqrt{n}}\int_{0}^{\mathcal{B}} \sqrt{\mathrm{VC}_{\mathcal{H}} \log \frac{6\mathcal{B}n}{\delta \mathrm{VC}_{\mathcal{H}} }} \mathrm{d} \delta, \\
  & \leq C_4 \mathcal{B}(\frac{n}{\mathrm{VC}_{\mathcal{H}}})^{1/2}\log (\frac{n}{\mathrm{VC}_{\mathcal{H}}}) \exp( -\log^2(\frac{n}{\mathrm{VC}_{\mathcal{H}}}))\\
  & \leq C_2\mathcal{B} \sqrt{\frac{n}{\mathcal{D}\mathcal{S}\log \mathcal{S}}}\log \frac{n}{\mathcal{D}\mathcal{S}\log \mathcal{S}} \exp(-\log^2 \frac{n}{\mathcal{D}\mathcal{S}\log \mathcal{S}})
  \end{align*}
  where, the first inequality follows from the chaining  Theorem 8.1.3 in \cite{vershynin2018high}, and the second inequality holds due to
  $d^{\mathcal{H}}_{2}\leq d^{\mathcal{H}}_{\infty}$, and in the third inequality we used
  the relationship between the matric entropy and the VC-dimension of the ReLU networks  $\mathcal{H}_{\mathcal{D}, \mathcal{W}, \mathcal{S}, \mathcal{B}}$ \citep{anthony2009neural}, i.e.,
  $$\log \mathcal{N}(\mathcal{H}, d^{\mathcal{H}}_{\infty}, \delta) \leq \mathrm{VC}_{\mathcal{H}} \log \frac{6\mathcal{B}n}{\delta\mathrm{VC}_{\mathcal{H}}},$$
  and the fourth inequality follows by  some calculation,
  and the last inequality  holds due to the  upper bound of VC-dimension for the ReLU network $\mathcal{H}_{\mathcal{D}, \mathcal{W}, \mathcal{S}, \mathcal{B}}$ satisfying  $$\mathrm{VC}_{\mathcal{H}} \leq C_5 \mathcal{D}\mathcal{S}\log \mathcal{S},$$ see \cite{bartlett2019}.\\
  For any two integer $M,N$, there exists  a  $\bar{R}_{\phi} \in \mathcal{H}_{\mathcal{D}, \mathcal{W}, \mathcal{S}, \mathcal{B}}$  with width
 $ \mathcal{W} = \max\{8 \mathcal{M} N^{1 /\mathcal{M}} +4 \mathcal{M}, 12 N+14\}$ and
 depth  $\mathcal{D} = 9M+12$, and $\mathcal{B} = 2B,$
 such that
  \begin{equation}
  \label{A10}
  \|r-\bar{R}_{\phi}\|^2_{L^{2}(\nu)}  \leq C_6    c L m \mathcal{M} (NM)^{-4/\mathcal{M}}.
  \end{equation}

    \textbf{Proof of (\ref{A10}).}
    We use Lemma 4.1, Theorem 4.3, 4.4   and following  the proof of Theorem 1.3 in  \cite{shen2019deep}.
    Let $\mathbf{A}$ be the random orthoprojector in Theorem 4.4, then
  it is to check $\mathbf{A}(\mathfrak{M}_{\epsilon}) \subset \mathbf{A}([-c,c]^{m}) \subset [-c\sqrt{m},\sqrt{m}c]^{\mathcal{M}}.$
  Let $\tilde{r}$ be an extension of the restriction of  $r$  on  $\mathfrak{M}_{\epsilon}$, which is defined  similarly  as $\tilde{g}$ on page 30  in  \cite{shen2019deep}.
   Since we assume the target $r$ is Lipschitz continuous with the bound $B$ and the Lipschitz constant $L$, let $\epsilon$ small enough,  then by Theorem 4.3, there exist a ReLU network  $\tilde{R}_{\phi} \in \mathcal{H}_{\mathcal{D}, \mathcal{W}, \mathcal{S}, \mathcal{B}}$  with width
 $$\mathcal{W} = \max\{8 \mathcal{M} N^{1 /\mathcal{M}} +4 \mathcal{M}, 12 N+14\},$$ and
 depth  $$\mathcal{D} = 9M+12,$$ and $\mathcal{B} = 2B,$
 such that
 $$\|\tilde{r} - \tilde{R}_{\phi}\|_{L^{\infty}(\mathfrak{M}_{\epsilon} \setminus \mathcal{N})} \leq 80c L \sqrt{m \mathcal{M}} (NM)^{-2/m},$$
 and
 $$\|\tilde{R}_{\phi}\|_{L^{\infty}(\mathfrak{M}_{\epsilon})} \leq B+3Lc \sqrt{m\mathcal{M}},$$
 where, $\mathcal{N}$ is a $\nu-$ negligible set with $\nu(\mathcal{N})$ can be arbitrary small.
 Define $\bar{R}_{\phi} = \tilde{R}_{\phi}\circ \mathbf{A}$. Then, following
 the proof after equation (4.8) in  Theorem 1.3 of \cite{shen2019deep}, we get our  (\ref{A10}) and $$\|\bar{R}_{\phi}\|_{L^{\infty}(\mathfrak{M}_{\epsilon} \setminus \mathcal{N})} \leq 2B, \|\bar{R}_{\phi}\|_{L^{\infty}(\mathcal{N})} \leq 2B+3cL \sqrt{m\mathcal{M}}.$$
Let $\mathcal{D}\mathcal{S}\log \mathcal{S}< n$, combing the results  (\ref{A7}) - (\ref{A10}),
 we have
 \begin{align*}
&\mathbb{E}_{\{X_i,Y_i\}_{1}^n} [\|\widehat{R}_{\phi} - r\|_{L^2(\nu)}^2] \\
&\leq 8C_1(2B+1) \mathfrak{G}(\mathcal{H}) +  C_6   c L m \mathcal{M} (NM)^{-4/\mathcal{M}}\\
&\leq  8C_1(2B+1)C_2 B\sqrt{\frac{\mathcal{D}\mathcal{S}\log \mathcal{S}}{n}}\log \frac{n}{\mathcal{D}\mathcal{S}\log \mathcal{S}}\\
& + C_6    cL m \mathcal{M} (NM)^{-4/\mathcal{M}}\\
& \leq  C(B^2+ cL m \mathcal{M}) n^{-2/(2+ \mathcal{M})},
 \end{align*}
 where, last inequality holds since  we  choose $M = \log n$, $ N = n^{\frac{\mathcal{M}}{2(2+\mathcal{M})}}/\log n$,  $\mathcal{S} = n^{\frac{\mathcal{M}-2}{\mathcal{M}+2}}/\log^4 n$, i.e.,
 $\mathcal{D} = 9 \log n + 12$, $\mathcal{W} = 12n^{\frac{\mathcal{M}}{2(2+\mathcal{M})}}/\log n+14.$
$\hfill$ $\Box$

\subsection{The relationship between EPT and MMD flow}\label{rmmd}
\label{pstein}
Here we show that MMD flow can be considered a special case of EPT.

\begin{proof}
Let $\mathcal{H}$ be a reproducing kernel Hilbert space with  characteristic kernel
 $K(\vx,\vz)$.
 Recall in MMD flow, $$\mathcal{L}[\mu] = \frac{1}{2}\|\mu-\nu\|_{\mathrm{mmd}}^2,$$ and
$$\frac{\partial \mathcal{L}[\mu]}{\partial \mu} (\vx) = \int K(\vx,\vz) \mathrm{d}\mu(\vz) -\int K(\vx,\vz) \mathrm{d}\nu(\vz),$$
and the vector fields
\begin{align*}
&\vv_{t}^{\mathrm{mmd}} = -\nabla \frac{\partial \mathcal{L}[\mu]}{\partial \mu_t}\\
 &= \int \nabla_{\vx} K(\vx,\vz) \mathrm{d}\nu(\vz)-\int \nabla_{\vx} K(\vx,\vz) \mathrm{d}\mu_t(\vz)\\
 & =  \int \nabla_{\vx} K(\vx,\vz)  p(\vz) \mathrm{d} \vz-\int \nabla_{\vx} K(\vx,\vz)q_t(\vz) \mathrm{d}\vz
\end{align*}
By Lemma \ref{lem2}, the vector fields corresponding the Lebesgue norm
$\frac{1}{2}\|\mu-\nu\|^2_{L^2(\mathbb{R}^{m})} =  \frac{1}{2}\int_{\mathbb{R}^{m}} |q(\vx)- p(\vx)|^2  {\mathrm{d}} \vx$  are defined as
$$\vv_{t} =  \nabla p(\vx) - \nabla q_t(\vx).$$
Next, we will show the vector fields $\vv_{t}^{\mathrm{mmd}}$ is exactly by projecting  the vector fields $\vv_{t}$ on to
the reproducing kernel Hilbert space $\mathcal{H}^{m} = \mathcal{H}^{\otimes m} $.
By the definition of reproducing kernel we have,
$$p(\vx) = \left \langle p(\cdot), K(\vx,\cdot) \right \rangle_{\mathcal{H}} = \int K(\vx,\vz) p(\vz)\mathrm{d}\vz,$$ and $$q_t(\vx) = \left \langle q_t(\cdot), K(\vx,\cdot) \right \rangle_{\mathcal{H}} = \int K(\vx,\vz) q_t(\vz)\mathrm{d}\vz.$$
Hence,
\begin{align*}
&\vv_{t}(\vx ) = \nabla p(\vx) - \nabla q_t(\vx) \\
=&\int \nabla_{\vx} K(\vx,\vz)( p(\vz) -  q_t(\vz))\mathrm{d}\vz\\
=& \vv_{t}^{\mathrm{mmd}}(\vx).
\end{align*}
This completes the proof.
\end{proof}

\subsection{Proof of the relation between EPT and SVGD} \label{rsvgd}
Here we show that SVGD can be derived from EPT.

\begin{proof}
Let
$f(u) = u\log u$ in (\ref{fdiv}). With this $f$ the velocity fields $\vv_t = -f''(r_t)\nabla r_t = -\frac{\nabla r_t(\mathbf{x})}{r_t(\mathbf{x})}$ Let $\mathbf{g}$ in a Stein class associated with $q_t$.
\begin{align*}
&\left \langle\vv_t, \mathbf{g} \right \rangle_{\mathcal{H}(q_t)} \\
=&-\int \mathbf{g}(\mathbf{x})^T\frac{\nabla r_t(\mathbf{x})}{r_t(\mathbf{x})}q_t(\mathbf{x})\mathrm{d}\mathbf{x}\\
=&-\int \mathbf{g}(\mathbf{x})^T\nabla \log r_t(\mathbf{x})q_t(\mathbf{x})\mathrm{d}\mathbf{x}\\
=&-\mathbb{E}_{\mathbf{X}\sim q_t(\mathbf{x})}[ \mathbf{g}(\mathbf{x})^T\nabla \log q_t(\mathbf{X}) +  \mathbf{g}(\mathbf{x})^T\nabla \log p(\mathbf{X})]\\
=&-\mathbb{E}_{\mathbf{X}\sim q_t(\mathbf{x})}[\mathbf{g}(\mathbf{x})^T\nabla \log q_t(\mathbf{X}) + \nabla\cdot \mathbf{g}(\mathbf{x})]\\
 &+ \mathbb{E}_{\mathbf{X}\sim q_t(\mathbf{x})}[\mathbf{g}(\mathbf{x})^T\nabla \log p(\mathbf{X}) + \nabla\cdot \mathbf{g}(\mathbf{x})]\\
=&-\mathbb{E}_{\mathbf{X}\sim q_t(\mathbf{x})}[\mathcal{T}_{q_t} \mathbf{g}]+ \mathbb{E}_{\mathbf{X}\sim q_t(\mathbf{x})}[\mathcal{T}_{p} \mathbf{g}]\\
=&  \mathbb{E}_{\mathbf{X}\sim q_t(\mathbf{x})}[\mathcal{T}_{p} \mathbf{g}],
\end{align*}
where the last equality is obtained by restricting $\mathbf{g}$ in a Stein class associated with $q_t$, i.e., $\mathbb{E}_{\mathbf{X}\sim q_t(\mathbf{x})}\mathcal{T}_{q_t} \mathbf{g} = 0$.
This is the velocity fields of SVGD \citep{liu2017}.
\end{proof}

\end{document}

%% file: math_commands.tex
\usepackage{amsmath,amsfonts,bm}

\def\eqref#1{equation~\ref{#1}}

\def\1{\bm{1}}

\def\vr{{\bm{r}}}

\def\vv{{\bm{v}}}

\def\vx{{\bm{x}}}
\def\vy{{\bm{y}}}
\def\vz{{\bm{z}}}

\def\m0{\bm{0}}

\DeclareMathAlphabet{\mathsfit}{\encodingdefault}{\sfdefault}{m}{sl}
\SetMathAlphabet{\mathsfit}{bold}{\encodingdefault}{\sfdefault}{bx}{n}

